\documentclass[10pt, letterpaper]{article}
\usepackage{hyperref}
\usepackage[utf8]{inputenc}
\usepackage[letterpaper, margin=1in]{geometry}
\usepackage{authblk}
\usepackage[english]{babel}
\usepackage{array,booktabs,multirow,threeparttable}
\usepackage{indentfirst}
\usepackage{url}

\usepackage{graphicx,subcaption}
\graphicspath{{figures/}, {figures/elapsed_time/}, {figures/datasets/}, {figures/experiments/}}

\usepackage{amssymb,bm,mathtools}
\DeclareMathOperator*{\argmax}{arg\,max}
\DeclarePairedDelimiter\norm{\lVert}{\rVert}

\usepackage[square,numbers,sort]{natbib}
\usepackage[ruled,lined,linesnumbered]{algorithm2e}
\SetKwInOut{Parameter}{parameter}
\makeatletter
\newcommand{\algrule}[1][.01pt]{\par\vskip.5\baselineskip\hrule height #1\par\vskip.5\baselineskip}
\makeatother

\begin{document}
\title{iCVI-ARTMAP: Accelerating and improving clustering using \\
adaptive resonance theory predictive mapping and \\
incremental cluster validity indices}
\author[1]{Leonardo Enzo Brito da Silva}
\author[1]{Nagasharath Rayapati}
\author[1,2]{Donald C. Wunsch II}
\affil[1]{Guise AI, Inc., USA}
\affil[2]{Applied Computational Intelligence Laboratory, Missouri University of Science and Technology, USA}
\date{\today}
\setcounter{Maxaffil}{0}
\renewcommand\Affilfont{\itshape\small}

\maketitle

\begin{abstract}
This paper presents an adaptive resonance theory (ART) model for unsupervised learning, namely iCVI-ARTMAP, which uses incremental cluster validity indices (iCVIs) to drive the clustering process within an ART predictive mapping (ARTMAP) model. Incorporating iCVIs to the decision-making and many-to-one mapping capabilities of ARTMAP can improve the choices of clusters to which samples are incrementally assigned. These improvements are accomplished by intelligently performing the operations of swapping sample assignments between clusters, splitting and merging clusters, and caching the values of variables when iCVI values need to be recomputed. Using recursive formulations enables iCVI-ARTMAP to considerably reduce the computational burden associated with the cluster validity index (CVI)-based offline incremental multi-prototype-based clustering task. Depending on the iCVI and the data set, it can achieve running times up to two orders of magnitude shorter than when performing the same clustering task using batch CVI computations. The aforementioned merging operation is performed because it is known that, when guided by a CVI, it has the potential to improve the data partition; splitting, however, is used to enforce the desired number of clusters. In this work, the incremental versions of Calinski-Harabasz (iCH), WB-index (iWB), Xie-Beni (iXB), Davies-Bouldin (iDB), Pakhira-Bandyopadhyay-Maulik (iPBM), and negentropy increment (iNI) were integrated into fuzzy ARTMAP. In extensive comparative experiments on synthetic benchmark data sets, iNI-ARTMAP yielded the best performance among these iCVI-ARTMAP variants, followed by iCH-, iWB- and iXB-ARTMAPs. Moreover, \mbox{iNI-ARTMAP} outperformed fuzzy ART, dual vigilance fuzzy ART, kmeans, spectral clustering, Gaussian mixture models and hierarchical agglomerative clustering algorithms in the vast majority of the synthetic benchmark data sets. It also performed competitively when clustering on projections and on the latent space generated by deep neural clustering models for real world image benchmark data sets. Naturally, the performance of iCVI-ARTMAP is subject to the selected iCVI and its suitability to the data at hand; fortunately, it is a general model wherein other iCVIs can be easily embedded.
\end{abstract}

\section{Introduction} \label{sec:intro}

Clustering is an unsupervised learning task performed in the machine learning pipelines of many applications. Briefly, it consists of partitioning a data set (in its raw or some transformed version~\cite{Bengio.2013a}) into different groups, wherein samples within a group are similar and between groups are dissimilar according to some predetermined criteria. A myriad of clustering algorithms have been designed in the corpus of computational intelligence, see~\cite{xu2005,xu2009} for a roadmap of traditional methods. In particular, \textit{adaptive resonance theory} (ART) networks~\cite{Carpenter1987} have been widely used for clustering and have addressed different problems, such as arbitrarily shaped clusters~\cite{Tscherepanow2010,leonardo.2018b,Elnabarawy.2019b,leonardo.2020b}, online normalization~\cite{Park.2019a}, heterogeneous data~\cite{Meng2014} and sparse data~\cite{Meng.2019b}. Moreover, ART networks have been used in diverse domains such as social media data~\cite{Meng.2019a} and biomedical data~\cite{xu2011}, as well as applications such as recommendation systems~\cite{Elnabarawy2016}, robotics~\cite{Park.2018a} and games~\cite{Wang.2009a, Wang.2015a, Silva.2018a}.

In parallel with clustering a data set, the practitioner usually needs to evaluate the quality of the obtained partitions, which is accomplished via \textit{cluster validity indices} (CVIs)~\cite[Chapter~10]{xu2009}. Succinctly, the latter map partitions to scalar values that represent quantitative assessments of such partitions, thus guiding the practitioner in selecting solutions yielded by clustering algorithms. Like clustering algorithms, many CVIs have been designed in the literature, and we refer the reader to~\cite{vendramin2010} for a more detailed treatment. To complement the traditional batch (or offline) CVIs, incremental (or online) CVIs (iCVIs)~\cite{Moshtaghi2018b} were recently developed to assess the performance of clustering algorithms applied to data streams, a use-case in which recursive formulations are mandatory: samples are assigned to clusters and discarded afterwards. In this context, by formulating a recursive computation for fuzzy compactness (a common quantity in the computations of sum-of-squared-based CVIs), iCVI versions were developed with the aim of evaluating \textit{cluster footprints} (term coined in~\cite{Moshtaghi2018b} to designate the stored statistics, such as prototypes) of data streams, where an incremental update was developed for the operation of adding one sample to the current partition. In the case of offline incremental clustering, more operations are required. For this purpose, in this work, those recursive computations of crisp compactness were extended to the operation of removing one sample from a cluster (in order to swap a sample between two clusters, along with the previously developed incremental update for adding one sample to a cluster, it is also necessary to perform an incremental update to remove the same sample from its original cluster), as well as to the operations of splitting and merging clusters.  

\begin{figure}[!t]
\centering
\includegraphics[width=\textwidth]{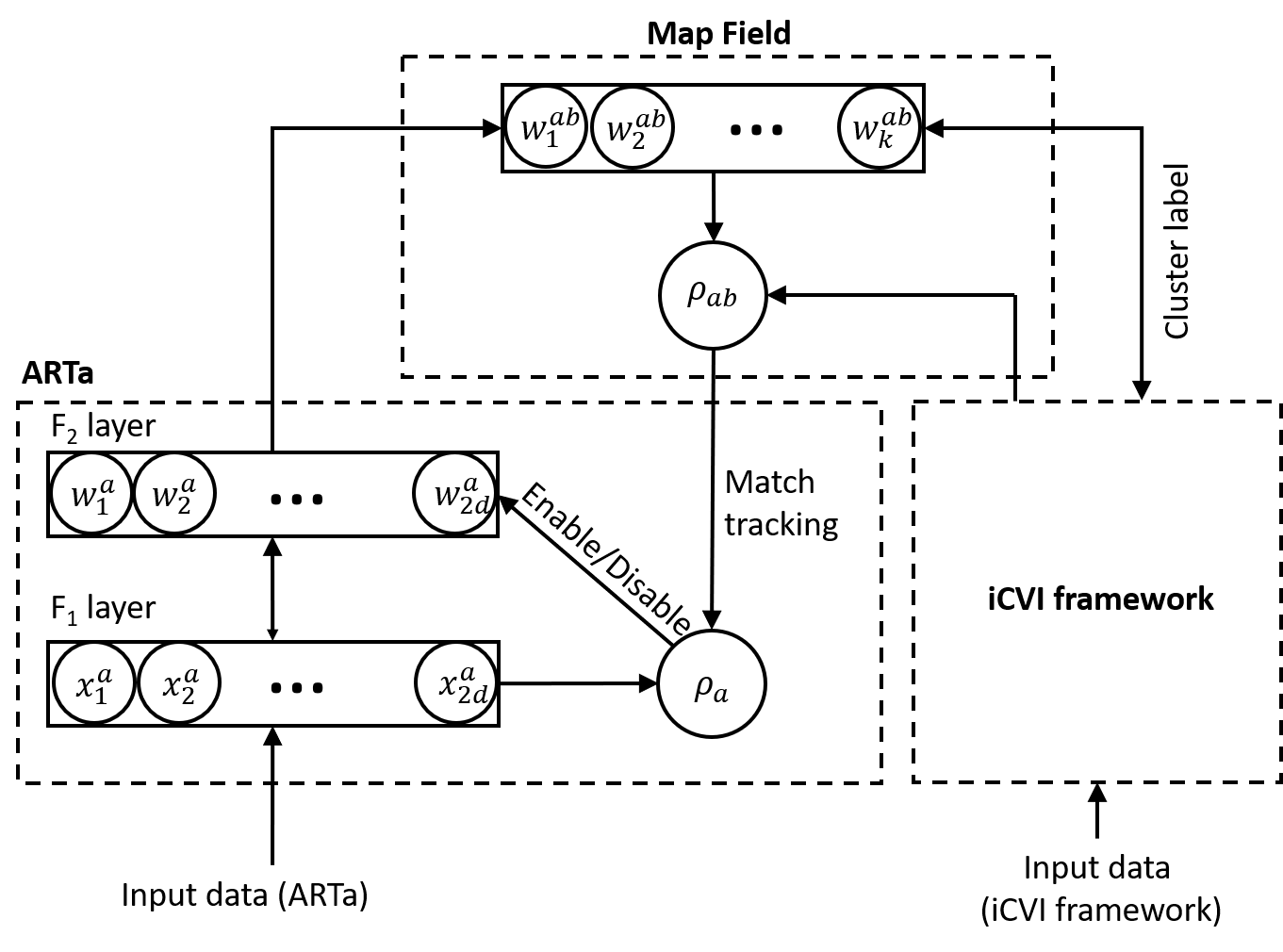}
\caption{iCVI-ARTMAP neural network. The inputs to ARTa and the iCVI-framework are the complement coded $\bm{x}^a \in \mathbb{R}^{2d}$ and the standardized $\bm{x}^b \in \mathbb{R}^d$ versions of the input data $\bm{x} \in \mathbb{R}^d$, respectively. Following the network initialization, the iCVI framework module generates a one-hot encoded cluster label when a sample is presented. This label is based on the iCVI values corresponding to the assignment of the sample to each cluster of the current data partition, where such values are computed incrementally. The cluster label is then used to carry out the dynamics of ARTa and map field modules. Next, the data partition and iCVI are updated using the map field prediction. If there is a change in the category assignment of the presented sample then pruning or shrinkage procedures take place in ARTa. Finally, at the end of each epoch, cluster merging and/or splitting can occur to improve the data partition and/or enforce the user-specified number of clusters, respectively.}
\label{fig:icviartmap}
\end{figure}

Batch CVIs have been used as a secondary vigilance parameter in a fuzzy ART~\cite{Carpenter1991} variant in~\cite{leonardo2017} and to optimize the vigilance parameter of fuzzy ART via particle swarm optimization in~\cite{smith2015}. In the online learning context, the partition separation index~\cite{Yang2001} has been embedded in an ART-like model in~\cite{lughofer2008}. Note that this CVI is computed online using the means and frequencies from a clustering algorithm; neither compactness nor covariance matrices are considered. Still in the ART domain, \textit{ART predictive mapping} (ARTMAP) networks~\cite{carpenter1992} traditionally have been employed to perform supervised learning tasks. Note that in~\cite{leonardo.2020a}, the ARTMAP variant used to realize the graph-based iCVI therein was set to supervised mode to emulate a highly performing clustering algorithm. Nonetheless, ARTMAP has also been transformed to perform unsupervised learning in models such as biclustering ARTMAP (BARTMAP)~\cite{xu2011}, its topological learning (multi-prototype) variant~\cite{Raghu.2020a} and hierarchical variant~\cite{Kim2016} (the latter employs a batch CVI in a typical use-case, i.e., to assess the resulting bicluster hierarchy levels); moreover, the ARTMAP-like model in~\cite{seiffertt2010} has been used to perform mixed-modality learning. However, the latter do not use iCVIs, completely integrated into the ARTMAP mechanisms, to drive the clustering process. The iCVI-ARTMAP introduced here (Fig.~\ref{fig:icviartmap}) repurposes ARTMAP to perform unsupervised learning via the combination with iCVIs as well as inherits its multi-prototype representation nature via its mapping mechanisms, thereby yielding improved performance. In addition, iCVI-ARTMAP inherently takes advantage of the incremental computation of iCVIs along with caching variables across the previously mentioned operations in order to achieve much shorter execution times than when performing the same offline (but incremental) clustering using batch CVI computations. Therefore, the contributions of this paper are three-fold:
\begin{enumerate}
\item Presents a novel ART-based clustering method consisting of an ARTMAP model combined with iCVIs.
\item In the context of iCVIs, offers recursive computations for hard cluster compactness when removing a sample from a cluster, as well as when merging and splitting clusters.
\item Provides a comparison of iCVIs used for offline incremental clustering purposes. 
\end{enumerate}

To the best of our knowledge, this is the first integration of iCVIs to an ARTMAP model. The remaining sections of this paper are organized as follows: Section~\ref{sec:rel} briefly discusses the previous work most relevant to this paper, Section~\ref{sec:related_work} provides an overview of ART and iCVIs; Section~\ref{sec:icvi-artmap} formally presents the iCVI-ARTMAP model (the main contribution of this work); Section~\ref{sec:experiments} describes the numerical experiments, and then reports on and discusses the results obtained; finally, Section~\ref{sec:conclusion} summarizes the findings of this paper.

\section{Related work} \label{sec:rel}

\textit{Incremental cluster validity indices} (iCVIs) were introduced by Moshtaghi et al.~\cite{Moshtaghi2018b} by developing a recursive computation for the fuzzy compactness of clusters (see Section~\ref{sec:iCVIs}) to incrementalize (term coined by Chenaghlou~\cite{Milad.2019a} to designate the derivation of incremental versions) the Xie-Beni (XB)~\cite{Xie1991} and Davies-Bouldin (DB)~\cite{db} batch CVIs, namely the iXB and iDB, respectively. Variants of iXB and iDB with exponential forgetting factors were also developed to take into account the time component by making recent samples more relevant. In lifelong learning use-cases, samples are presented ad infinitum; thus, as the number of processed samples increases, the influence of individual samples on the compactness value decreases~\cite{Moshtaghi2018b,Milad.2019a}. The explainability power of these iCVIs was investigated in the context of sequential k-means~\cite{kmeans} and online ellipsoidal clustering~\cite{Moshtaghi.2016a} algorithms.

Ibrahim et al.~\cite{Ibrahim2018, Keller2018} investigated the behavior of the iDB for the particular case of clustering data streams using the extended robust online streaming clustering (EROLSC) algorithm~\cite{Ibrahim.2016a,Ibrahim.2018a}. Specifically, in~\cite{Ibrahim2018}, the iDB dynamics were analyzed when creating new clusters, dealing with unbalanced clusters and varying the value of the membership exponent parameter of the fuzzy compactness (see Section~\ref{sec:iCVIs}), whereas in~\cite{Keller2018}, the dynamics of this iCVI were analyzed when processing large data and high dimensional data (in their original as well as in lower dimensional projected spaces). Also, in the context of the EROLSC online learner, incremental versions of the partition coefficient and exponential separation~\cite{Wu.2005a} (iPCAES) and the generalized Dunn's indices 43 and 53~\cite{Bezdek.1998a} (iGD\textsubscript{43} and iGD\textsubscript{53}) CVIs were presented and their behavior studied in~\cite{Ibrahim2018b} and~\cite{Ibrahim.2019a}, respectively. Additionally, in~\cite{Ibrahim2018}, the compactness was used to decide if buffered samples previously classified as anomalies by the EROLSC algorithm should be considered as a new cluster. 

Recently, Brito da Silva et al.~\cite{leonardo.2020a} presented incremental versions of the following CVIs: Calinski-Harabasz~\cite{vrc} (iCH), Pakhira-Bandyopadhyay-Maulik~\cite{pbm} (iPBM), WB-index~\cite{Zhao.2014a} (iWB), centroid-based silhouette~\cite{Rawashdeh2012} (iSIL), representative cross information potential and representative cross-entropy~\cite{araujo20132} (irCIP and irH, respectively), conn\_index~\cite{tasdemir2011} (iconn\_index), and negentropy increment~\cite{fernandez2010} (iNI). A comparison study including 13 iCVIs was also conducted to investigate, in a (for the most part) clustering algorithm agnostic setting, their behavior in cases of correct, under- and over-partitioning of data sets associated with different challenges such as imbalance, varied overlap degrees and cardinalities.

In addition to their post-processing and monitoring roles, CVIs have been used extensively in the clustering literature as fitness functions in optimization approaches. For instance, in the context of offline learning, Xu et al.~\cite{xu2012} used batch CVIs as fitness functions in a differential evolution and particle swarm optimization hybrid approach to clustering, whereas Smith et al.~\cite{smith2015} used them to optimize the vigilance parameter of fuzzy ART via particle swarm optimization. Brito da Silva et al.~\cite{leonardo2017} employed batch CVIs as a second vigilance criteria for fuzzy ART to improve the performance of the original model, as well as to provide robustness to the standard vigilance parameter selection and mitigate ordering effects. 

In the context of online learning, Lughofer~\cite{lughofer2008} presented evolving vector quantization (eVQ), an ART-like clustering method that includes several enhancements to the class of incremental vector quantization algorithms, such as the removal of satellite clusters and a split-and-merge heuristic guided by a crisp version of the PS CVI~\cite{Yang2001}. The latter can be readily computed when clustering algorithms use frequencies and means for prototype representation. In 2019, Chenaghlou~\cite[Chapter~6]{Milad.2019a} presented a framework to aid the decision-making ability of online clustering algorithms using iXB with a forgetting factor. This framework consisted of an online clustering algorithm and a controller, the latter of which determines the creation of a new prototype or merging of two prototypes when certain conditions based on iXB are met. In particular, it was shown to improve the performance of sequential kmeans~\cite{kmeans} and Online Clustering and Anomaly Detection (onCAD)~\cite[Chapter~4]{Milad.2019a}. Although it merges prototypes and frequencies, this framework does not address the merging or splitting of clusters~\cite[Chapter~7]{Milad.2019a}. Moreover, other controllers may need to be designed for different iCVIs and the performance of the same controller investigated for other iCVIs~\cite[Chapter~7]{Milad.2019a}. 

As opposed to the methods previously discussed, our approach presented here is an offline incremental learner that makes use of full-fledged sum-of-squares-based and information-theoretic-based iCVIs to completely drive the clustering process, including assigning individual samples to clusters and splitting and merging clusters. Moreover, the system presented in our work belongs to the class of multi-prototype-based clustering and can seamlessly use any of the sum-of-squares-based and information-theoretic-based iCVIs without any change to the general framework. Finally, to allow this incremental learner to process data, incremental formulations for hard compactness when merging and splitting clusters, as well as when removing a sample from a cluster (so as to perform a sample swap between clusters), are presented. 

\section{Preliminaries} \label{sec:related_work}

The following subsections provide background to contextualize the contributions of this paper and make it self-contained. For a more comprehensive treatment of adaptive resonance theory and incremental clustering validity indices, refer to~\cite{leonardo.2019b} and~\cite{leonardo.2020a} (and the references within), respectively.

\subsection{Adaptive resonance theory}\label{sec:ART}

Adaptive resonance theory (ART) neural networks~\cite{Carpenter1987} are incremental match-based models able to learn in both online and offline settings. Many have been designed with inherent plasticity and stability properties. Different networks have been developed to address the three canonical machine learning paradigms, i.e., unsupervised, supervised and reinforcement learning. These networks often share common design principles; in particular, their building blocks are usually elementary ART networks devised for unsupervised learning. This work makes use of fuzzy ARTMAP~\cite{carpenter1992}, and the following exposition briefly describes its building blocks: fuzzy ART~\cite{Carpenter1991} and the map field module~\cite{carpenter1992}.

Consider an input pair $(\bm{x}^a, \bm{x}^b)$ presented to fuzzy ARTMAP. The inputs $\bm{x}^a \in \mathbb{R}^{d_a \times 1}$ and $\bm{x}^b \in \mathbb{R}^{d_b \times 1}$ are presented to two distinct fuzzy ARTs, namely modules A (ARTa) and B (ARTb), respectively. These modules are interconnected by a map field network, which is responsible for the mapping across both domains. In the supervised learning scenario (classification), $\bm{x}^a$ represents a sample $\bm{x}$, whereas $\bm{x}^b$ represents a class label $\bm{y}$. In such a case, fuzzy ARTMAP is usually simplified by replacing the ARTb module with a stream of class labels~\cite{kasuba1993}. 

We start here by describing the dynamics of a fuzzy ART network. When presenting a normalized and complement-coded~\cite{carpenter1992} sample $\bm{x} \in \mathbb{R}^{2d \times 1}$ to fuzzy ART, the activation values ($T$) across the $C$ network nodes ($\bm{w} \in \mathbb{R}^{2d \times 1}$) are computed as:
\begin{equation}
T_j = \dfrac{\| \bm{x} \wedge \bm{w}_j \|_1}{\alpha + \| \bm{w}_j \|_1}, \quad \alpha > 0,
\label{Eq:FA_T}
\end{equation}
\noindent where $\wedge$ represents an element-wise minimum operation, and $\| \cdot \|_1$ represents the $\ell_1$ norm. A winner-take-all competition is then promoted to find the node $J$ that  maximizes $T$ (i.e., $J = \argmax\limits_{j \in \{1,...,C\}} \left( T_j \right)$). If such a node is found, then its match value ($M$) is computed as:
\begin{equation}
M_J = \dfrac{\| \bm{x} \wedge \bm{w}_J \|_1}{\| \bm{x} \|_1},
\label{Eq:FA_M}
\end{equation}
\noindent where $M_J$ is used to perform a resonance test. The latter consists of comparing $M_J$ with a vigilance parameter $\rho \in [0,1]$: 
\begin{equation}
M_J \ge \rho.
\label{Eq:FA_vig}
\end{equation}
If the condition expressed by Eq.~(\ref{Eq:FA_vig}) is satisfied, then learning ensues:
\begin{equation}
\bm{w}_J(t+1) = (1 - \beta)\bm{w}_J(t) + \beta\left[ \bm{x} \wedge \bm{w}_J(t) \right], \quad \beta \in (0,1],
\label{Eq:FA_learn}
\end{equation}
\noindent where $\beta$ is the learning rate. Otherwise, node $J$ is inhibited and another winner-take-all competition takes place, thus restarting the entire process. A node that has underwent learning is referred to as committed whereas a node that has not is referred to as uncommitted and is equal to $\Vec{\bm{1}}$. If an uncommitted node is selected for learning then another one is created. Thus, a standard fuzzy ART network maintains one uncommitted node at all times.

In fuzzy ARTMAP, the dynamics of ARTa follow the description above with the additional constraint that in order to reach a resonant state and learn, in addition to satisfying the unsupervised learning criteria imposed by Eq.~(\ref{Eq:FA_vig}), node $J$ must also satisfy the supervised learning resonance test performed by a map field module. Hereafter, the subscripts or superscripts ``a'' and ``ab'' will indicate ARTa and map field variables and parameters, respectively. Assuming ARTa has $C_a$ nodes and there are $k$ classes in the data set, once a resonant category $J$ is found, the match function $M^{ab}$ of the map field is computed as:
\begin{equation}
M_J^{ab} =  \dfrac{||\bm{y} \wedge \bm{w}^{ab}_J||_1}{|| \bm{y} ||_1},
\end{equation}
\noindent where $\bm{w}^{ab}_J \in \mathbb{R}^{1 \times k}$ is the $J^{th}$ row of the map field's mapping matrix $\bm{W}^{ab} \in \mathbb{R}^{C_a \times k}$, and $\bm{y} \in \mathbb{R}^{1 \times k}$ corresponds to the one-hot encoding of the class to which $\bm{x}$ belongs. If a mismatch occurs, i.e., if the resonance criteria
\begin{equation}
M_J^{ab} \geq \rho_{ab}
\label{Eq:MF_vig}
\end{equation}
\noindent is not satisfied, then the map field inhibits this category~$J$ of ARTa, and a new search takes place. On the other hand, if ARTa's category $J$ satisfies both $\rho_a$ (Eq.~(\ref{Eq:FA_vig})) and $\rho_{ab}$ (Eq.~(\ref{Eq:MF_vig})), then learning ensues in ARTa (Eq.~(\ref{Eq:FA_learn})) and in the map field as~\cite{Carpenter1995c}:
\begin{equation}
\bm{w}_J^{ab}(t+1) = (1 - \beta_{ab})\bm{w}_J^{ab}(t) + \beta_{ab}\left[ \bm{y} \wedge \bm{w}^{ab}_J(t) \right], \quad \beta_{ab} \in (0, 1], 
\end{equation}
\noindent where $\beta_{ab}$ is the learning rate of the map field. If ARTa created a new category $\bm{w}_{new}^{a}$, then the map field also creates a corresponding weight vector $\bm{w}_{new}^{ab}$.

\subsection{Incremental cluster validity indices}\label{sec:iCVIs}

\textit{Cluster validity indices} (CVI)~\cite[Chapter~10]{xu2009} map a partition to a scalar  that represents an assessment of such partition, i.e., CVIs are designed to quantitatively evaluate the quality of clustering solutions. CVIs are essential when performing cluster analysis because of the absence of class labels in unsupervised learning problems, as well as the myriad of clustering algorithms available in the literature, and the fact that different parameter settings or ordering effects~\cite{xu2012,leonardo2018} may affect their outputs. Traditional CVIs either perform a comparison to a reference partition (external CVIs) or use the data and the partition itself to make an assessment (internal CVIs). Both cases rely on batch (offline) computations. Substantial research efforts have been employed in designing and comparing the relative merits/performances of CVIs, and we refer interested readers to~\cite{milligan1985, Bezdek1997, Halkidi2002a, Halkidi2002b, vendramin2010, Arbelaitz2013, Hamalainen.2017a}. 

Recently, research efforts have shifted towards devising incremental (and online) versions of CVIs to evaluate partitions in the context of data streams~\cite{Moshtaghi2018b, Ibrahim2018, Keller2018, Ibrahim2018b, leonardo.2020a}, namely, \textit{incremental cluster validity indices} (iCVI). In such applications, samples are discarded after their presentation, thus becoming unavailable for iteration. This was first introduced in~\cite{Moshtaghi2018b} for fuzzy versions of sum-of-squares-based CVIs by devising a recursive formulation for compactness, which is a recurring quantity in such types of CVIs (most of these comprise some sort of trade-off between compactness (or scatter, dispersion) and separation (or isolation)~\cite{xu2012}). Fuzzy compactness is defined as~\cite{Moshtaghi2018b}:
\begin{equation}
CP_i = \sum \limits_{j=1}^{N} \gamma_{j,i}^m \| \bm{x}_j - \bm{\mu}_i \|^2_2,
\label{Eq:fuzzy_CP}
\end{equation}
\noindent where $N$ is the cardinality of the dataset, $\gamma_{i,j}$ is the membership function of sample $\bm{x}_j$ to cluster $\Omega_i$, $m \geq 1$~\cite{Ibrahim2018} is the membership exponent and $CP_i$ and $\bm{\mu}$ are the compactness and the mean (or centroid, prototype) of cluster $\Omega_i$, respectively. In particular, a crisp version of the compactness in Eq.~(\ref{Eq:fuzzy_CP}) can be obtained by straightforwardly replacing the fuzzy membership values in Eq.~(\ref{Eq:fuzzy_CP}) with indicator functions. Thus, the fuzzy incremental compactness of cluster $i$~\cite{Moshtaghi2018b} has a crisp version that can be computed as~\cite{leonardo.2020a}:
\begin{equation}
CP_i(t+1)=CP_i(t) + \| \bm{x} - \bm{\mu}_i(t+1) \|^2_2 + n_i(t) \| \bm{\mu}_i(t) - \bm{\mu}_i(t+1) \|^2_2 + 2\left[\bm{\mu}_i(t) - \bm{\mu}_i(t+1)\right]^T\bm{g}_i(t),
\label{Eq:iCP_1}
\end{equation}
\noindent where
\begin{equation}
CP_i(t) = \sum \limits_{j=1}^{n_i(t)} \| \bm{x}_j - \bm{\mu}_i(t) \|^2_2, \quad \bm{x}_j \in \Omega_i,
\label{Eq:iCP_2}
\end{equation}
\begin{equation}
\bm{\mu}_i(t+1) = \bm{\mu}_i(t) + \dfrac{1}{n_i(t+1)}\left[\bm{x} - \bm{\mu}_i(t)\right],
\label{Eq:iCP_3}
\end{equation}
\begin{equation}
n_i(t+1) = n_i(t) + 1,
\label{Eq:iCP_4}
\end{equation}
\begin{equation}
\bm{g}_i(t) = \sum\limits_{j=1}^{n_i(t)} \left[ \bm{x}_j - \bm{\mu}_i(t) \right],
\label{Eq:iCP_5}
\end{equation}
\noindent vector $\bm{g}$ is incrementally updated as
\begin{equation}
\bm{g}_i(t+1) = \bm{g}_i(t) + \left[ \bm{x} - \bm{\mu}_i(t+1) \right] + n_i(t) \left[\bm{\mu}_i(t) - \bm{\mu}_i(t+1)\right],
\label{Eq:iCP_6}
\end{equation}
\noindent and
\begin{equation}
N(t+1) = N(t) + 1.
\label{Eq:iCP_7}
\end{equation}

For iCVIs whose computations require the estimation of covariance matrices, the latter may be calculated incrementally using the classic recursive computation of covariance matrices~\cite{duda2000}: 
\begin{equation}
\bm{\Sigma}_i(t+1) = \frac{n(t+1)-2}{n_i(t+1)-1}\bm{\Sigma}_i(t) + \frac{1}{n_i(t+1)}\left[\bm{x} - \bm{\mu}_i(t) \right]\left[\bm{x} - \bm{\mu}_i(t) \right]^T,
\label{Eq:iSigma}
\end{equation}
\noindent where
\begin{equation}
\bm{\Sigma}_i(t) = \frac{1}{n_i(t) - 1} \left[\sum\limits_{i=1}^{n_i(t)} \bm{x}_i \bm{x}_i^T -  n_i(t) \bm{\mu}_i(t) \bm{\mu}_i(t)^T \right]. 
\label{Eq:Sigma}
\end{equation}

Incremental versions have been developed for sum-of-squares-based, information-theoretic-based and graph-based CVIs, including: Xie-Beni (XB)~\cite{Xie1991} and Davies-Bouldin (DB)~\cite{db} in \cite{Moshtaghi2018b}; generalized Dunn 43 and~53 (GD\textsubscript{43} and GD\textsubscript{53})~\cite{Bezdek.1998a} in~\cite{Ibrahim2018b}; and Partition Coefficient and Exponential Separation (PCAES)~\cite{Wu.2005a} in~\cite{Ibrahim.2019a}; as well as Calinski-Harabasz (CH)~\cite{vrc}, Pakhira-Bandyopadhyay-Maulik (PBM)~\cite{pbm}, WB-index (WB)~\cite{Zhao.2014a}, centroid-based Silhouette (SIL)~\cite{Rawashdeh2012}, representative cross-information potential and representative cross-entropy (rCIP and rH)~\cite{araujo20132}, conn\_index~\cite{tasdemir2011} and negentropy increment (NI)~\cite{fernandez2009a,fernandez2010} in~\cite{leonardo.2020a}. Table~\ref{Tab:CVI_summary} summarizes the formulae of the iCVIs used in this work.

\begin{table}[!b]
\centering
\caption{iCVIs used in this study (adapted from~\cite{leonardo.2020a}).}
\begin{tabular}{llll}
\toprule
iCVI & Definition & Optimality & Reference(s) \\
\midrule
\midrule
iCH
& $\displaystyle \dfrac{\sum \limits_{i=1}^{k} n_i \| \bm{\mu}_i - \bm{\mu}_{data} \|^2_2}{\sum \limits_{i=1}^{k} CP_i} \times \dfrac{N-k}{k-1}$  
& max-optimal
& \cite{vrc, leonardo.2020a} \\
iWB   
& $\displaystyle \dfrac{\sum \limits_{i=1}^{k} CP_i}{\sum \limits_{i=1}^{k} n_i \| \bm{\mu}_i - \bm{\mu}_{data} \|^2_2} \times k$  
& min-optimal
&  \cite{Zhao.2014a, leonardo.2020a} \\
iDB    
& $\displaystyle \dfrac{1}{k} \sum\limits_{i=1}^k \max\limits_{i \neq j}\left( 
\dfrac{ 
\dfrac{CP_i}{n_i} +  \dfrac{CP_j}{n_j} 
}{\| \bm{\mu}_i - \bm{\mu}_j \|_2^2} \right)$   
& min-optimal
& \cite{db, Moshtaghi2018b} \\
iXB  
&  $\displaystyle \dfrac{\sum \limits_{i=1}^{k} CP_i}{\min\limits_{i \neq j} \| \bm{\mu}_i - \bm{\mu}_j \|^2_2 } \times \dfrac{1}{N}$
& min-optimal
& \cite{Xie1991, Moshtaghi2018b} \\
iPBM 
& $\displaystyle \left[\dfrac{
\sum\limits_{i=1}^N \norm{\bm{x}_i - \bm{\mu}_{data}}_2^2
}
{\sum\limits_{i=1}^k CP_i} \times 
\max\limits_{i \neq j}\norm{\bm{\mu}_i - \bm{\mu}_j}_2^2 \times \dfrac{1}{k}  \right]^2$       
& max-optimal
& \cite{pbm, leonardo.2020a}\\
iNI
& $\displaystyle \dfrac{1}{2}\sum\limits_{i=1}^{k} p_i\ln | \bm{\Sigma}_i | -  \sum\limits_{i=1}^{k} p_i\ln p_i - \dfrac{1}{2} \ln |\bm{\Sigma}_{data}|$  
& min-optimal
& \cite{fernandez2009a, fernandez2010, leonardo.2020a}\\
\bottomrule
\end{tabular}
\label{Tab:CVI_summary}
\end{table} 

\section{The iCVI-ARTMAP model} \label{sec:icvi-artmap}

The iCVI-ARTMAP model presented in this paper is based on fuzzy ARTMAP and is illustrated in Figure~\ref{fig:icviartmap}. The dynamics of each component of iCVI-ARTMAP, as well as its training procedure, are described in the following subsections. 

\subsection{ARTa and map field}	\label{sec:arta}

The ARTa and map field modules of iCVI-ARTMAP follow the dynamics described in Section~\ref{sec:ART}. However, there are no uncommitted categories. In particular, if no category in ARTa satisfies the resonance tests, then a new one is created as $\bm{w}_{new}^{a} = \bm{x}^a$ (fast commit). In addition, the map field creates the following corresponding weight vector: $\bm{w}_{new}^{ab} = \Vec{\bm{1}}$.

\subsection{iCVI-framework}	\label{sec:artb}

iCVI-ARTMAP follows a simplified design, and thus, the ARTb module is replaced with the iCVI-framework which is responsible for computing the vector $\bm{y}$ that encodes the cluster assignment. When a sample $\bm{x}^b$ is presented at time $t$, the cluster assignment is defined as the cluster label that optimizes the iCVI given the current partition:
\begin{equation}
y_l = 
\begin{cases}
1 & \text{, if} \quad l = \argmax\limits_i(T^b_i)\\
0 & \text{, otherwise}
\end{cases}, 
\label{Eq:ARTb_yF2}
\end{equation}
\noindent where $T^b_i$ represents the iCVI value when swapping the assignment of the presented sample $\bm{x}^b$ from its current cluster $j$ to cluster $i$. If all $T^b_i$s are equal, then $\bm{y} = \Vec{\bm{1}}$. Therefore, $\bm{y} \in \mathbb{R}^{1 \times k}$ is a binary vector (one-hot encoding) of crisp membership function of sample $\bm{x}$ to current clusters under the assumptions of the selected iCVI. Note that Eq.~(\ref{Eq:ARTb_yF2}) corresponds to a max-optimal iCVI; naturally, $-T^b_i$ is used for the min-optimal case. The computation of $T^b_i$ within the iCVI-framework is discussed in Section~\ref{sec:icvi}.

\subsection{Merge and split heuristic}	\label{Sec:split_and_merge}

Splitting and merging strategies are common approaches used in clustering algorithms~\cite{Krishnapuram.1992a, Rhee.2003a, Beringer.2006a, lughofer2008, leonardo.2020b}. Here, they are performed only once per epoch due to their computational costs. Specifically, if there are at least three clusters at the end of a training epoch, pairs of clusters are merged in an iterative manner until one of the following conditions are satisfied:
\begin{enumerate}
\item The iCVI value for merging any two clusters is worse than the iCVI of the partition prior to merging.
\item There are only two clusters left.
\end{enumerate}

When any two clusters $i$ and $j$ are merged, in addition to updating the iCVI variables (Section~\ref{sec:icvi}), the assignment of the categories in the map field module also change. The ARTa weight vectors, however, remain the same. Assuming that there are $k'$ clusters at the end of an epoch, then a column vector $\bm{v} \in \mathbb{R}^{C_a \times 1}$ is created such that its component $l$ is given by 
\begin{equation}
v_l = 
\begin{cases}
\max\limits_{m \in \{i,j\}}(w_{l, m}^{ab})
& \text{, if} \quad \argmax\limits_m(w_{l, m}^{ab}) \in \{i, j\} \\
\min\limits_{m \in \{i,j\}}(w_{l, m}^{ab}) 
& \text{, otherwise}
\end{cases}, 
\label{Eq:merge_1}
\end{equation}
\noindent where a $\bm{w}_l^{ab} \in \mathbb{R}^{1 \times k'}$ is the $l^{th}$ row vector of the map field matrix $\bm{W}^{ab} \in \mathbb{R}^{C_a \times k'}$. The column vector $\bm{v}$ is then used to extend $\bm{W}^{ab}$ (via concatenation) to
\begin{equation}
\bm{W}^{ab} \leftarrow [\bm{W}^{ab} ~|~ \bm{v}],
\label{Eq:merge_2}
\end{equation}
\noindent such that $\bm{W}^{ab} \in \mathbb{R}^{C_a \times (k'+1)}$. Finally, the columns $i$ and $j$ of $\bm{W}^{ab}$ are deleted such that $\bm{W}^{ab} \in \mathbb{R}^{C_a \times (k'-1)}$ (i.e., $\bm{w}_{i}^{ab} \in \mathbb{R}^{C_a \times 1}$ and $\bm{w}_{j}^{ab} \in \mathbb{R}^{C_a \times 1}$ are removed from $\bm{W}^{ab}$). 
 
Next, in order to enforce a partition with $k$ clusters, clusters represented by multiple ARTa prototypes according to the map field prediction (Eq.~(\ref{Eq:mf_pred})) and that have samples currently assigned to more than one ARTa category become candidates for  splitting. Specifically, from these potential clusters (if any), the ARTa category that yields the best iCVI value when representing a cluster on its own is defined as a new cluster (note that this best value may still be worse than the previous partition with a smaller value of $k$). When a cluster is represented by two categories, creating a new cluster from either category yields the same iCVI value. In such a scenario, we observe the differences between the largest and second largest map field entries (i.e. cluster assignments) associated with each of these two categories; this difference has been used, for instance, in~\cite{Mrabah.2020a} as one of the criteria to flag samples with uncertain cluster assignment. For consistency, the category selected to create a new cluster and have its map field matrix entries modified is the one with the smallest value for the ratio defined by the aforementioned difference divided by its largest map field entry. Considering that there are $k''$ clusters after the merging procedure, when a cluster $i$ is split and a category $q$ is regarded as a new cluster, a column vector $\bm{v} \in \mathbb{R}^{C_a \times 1}$ is computed as
\begin{equation}
v_l = 
\begin{cases}
\max\limits_{j}(w_{q, j}^{ab})
& \text{, if} \quad l=q \\
0 
& \text{, otherwise}
\end{cases},
\label{Eq:split_1}
\end{equation}
\noindent where $\bm{w}_q^{ab} \in \mathbb{R}^{1 \times k''}$ is the $q^{th}$ row vector of the map field matrix $\bm{W}^{ab} \in \mathbb{R}^{C_a \times k''}$. The former has the following entry updated:
\begin{equation}
w_{q, h} \leftarrow \min\limits_{j}(w_{q, j}^{ab}),
\label{Eq:split_2}
\end{equation}
\noindent where $h = \argmax\limits_j(w_{q,j}^{ab})$ is the current assignment of category $q$ according to the map field prediction (Eq.~(\ref{Eq:mf_pred})). For the edge cases in which $\bm{w}_q^{ab} = c\Vec{\bm{1}}$, $c \in [0,1]$, instead of Eq.~(\ref{Eq:split_2}), all components of $\bm{w}_q^{ab}$ are updated using:
\begin{equation}
\bm{w}_q \leftarrow (c - \delta)\Vec{\bm{1}},
\label{Eq:split_3}
\end{equation} 
\noindent where $\delta$ is set to a very small value, such as $10^{-6}$. Finally, the column vector $\bm{v}$ is used to grow $\bm{W}^{ab}$ (which had its $q^{th}$ row modified according to Eq.~(\ref{Eq:split_2}) or~(\ref{Eq:split_3})):
\begin{equation}
\bm{W}^{ab} \leftarrow [\bm{W}^{ab} ~|~ \bm{v}],
\label{Eq:concat_s}
\end{equation}
\noindent such that $\bm{W}^{ab} \in \mathbb{R}^{C_a \times (k''+1)}$. In addition to the map field matrix update, the iCVI variables also undergo splitting (Section~\ref{sec:icvi}). Again, the ARTa weight vectors remain the same. 

\subsection{iCVIs and ARTMAP design} \label{sec:icvi}
 
At each iteration (or time step), a maximum of one cluster label swap of the presented sample $\bm{x}^b$ will take place, so recomputing all variables (e.g., all clusters' frequencies, means, pairwise (dis)similarities, etc.) used for the CVI calculation is unnecessary. Moreover, the variables that must be recomputed can be recomputed in an incremental manner. In this context, note that in the case of data streams (domain for which the iCVIs were originally devised), an incremental computation of compactness was presented for the case of adding one sample to a cluster. Moreover, the number of data set samples $N$ is always increasing. In our case, however, $N$ is a known and fixed quantity. Besides, in addition to the case of adding one sample to a cluster, the following incremental updates for three additional cases are required:
\begin{enumerate}
\item Removing one sample from a cluster. 
\item Merging cluster $i$ with cluster $j$ (these clusters possibly have multiple samples).
\item Creating a new cluster $j$ with multiple samples from cluster $i$ (in the latter case, the variables of the new cluster $j$ are computed in batch mode, whereas the variables of cluster $i$ are computed incrementally).
\end{enumerate}

The iCVIs are used to compute $T^b_i$ in Eq.~(\ref{Eq:ARTb_yF2}), thus effectively acting as fitness functions. The cases in which a sample maintains its current cluster assignment (no operation, i.e., the iCVI is not recomputed) and it is swapped from cluster $i$ to cluster $j$ (remove and add operations) are used to compute $T^b_i$, whereas the merge and split operations are used for the merge and split strategy at the end of each epoch. In all cases, with the exception of new clusters created when performing a split, the CVI computation is carried out incrementally (i.e., we make use of iCVIs), and intermediate quantities are cached to avoid unnecessary re-computations, thereby further reducing the computational burden. For instance, like in~\cite{leonardo.2020a}, only the columns and rows of dissimilarities matrices (if any) associated with cluster swaps, splits and merges are recomputed. This work further extends the set of incremental updates of hard compactness to include the three previously mentioned cases; the other cluster statistics use well-known incremental updates. The recursive computations for the iCVI variables of frequency ($n$), mean ($\bm{\mu}$), compactness ($CP$) and covariance matrix ($\bm{\Sigma}$) are listed below for all five possible cases (add, remove, merge, split and no operation):
\pagebreak
\begin{itemize}
\item Frequency ($n \in \mathbb{R}^1$):
\end{itemize}
\begin{equation}
n_i(t+1) =
\begin{cases}
n_i(t) + 1
& \quad \text{, add} \\
n_i(t) - 1 
& \quad \text{, remove} \\
n_i(t) + n_j(t)
& \quad \text{, merge}  \\
n_i(t) - n_j(t)
& \quad \text{, split} \\
n_i(t)
& \quad \text{, no operation}
\end{cases}
\label{Eq:freq_update}
\end{equation}

\begin{itemize}
\item Mean ($\bm{\mu} \in \mathbb{R}^d$):
\end{itemize}
\begin{equation}
\bm{\mu}_i(t+1) =
\begin{cases}
\dfrac{n_i(t)}{n_i(t+1)}\bm{\mu}_i(t) + \dfrac{1}{n_i(t+1)}\bm{x}^b
& \quad \text{, add} \\
\dfrac{n_i(t)}{n_i(t+1)}\bm{\mu}_i(t) - \dfrac{1}{n_i(t+1)}\bm{x}^b
& \quad \text{, remove} \\
\dfrac{n_i(t)}{n_i(t+1)} \bm{\mu}_i(t) + \dfrac{n_j(t)}{n_i(t+1)} \bm{\mu}_j(t)
& \quad \text{, merge}  \\
\dfrac{n_i(t)}{n_i(t+1)} \bm{\mu}_i(t) - \dfrac{n_j(t)}{n_i(t+1)} \bm{\mu}_j(t)
& \quad \text{, split} \\
\bm{\mu}_i(t)
& \quad \text{, no operation}
\end{cases}
\label{Eq:mean_update}
\end{equation}

\begin{itemize}
\item Compactness ($CP \in \mathbb{R}^1$):
\end{itemize}
\begin{equation}
CP_i(t+1) =
\begin{cases}
CP_i(t) + \dfrac{n_i(t)}{n_i(t) + 1} || \bm{x}^b - \bm{\mu}_i(t) ||_2^2 
& \quad \text{, add} \\
CP_i(t) - \dfrac{n_i(t)}{n_i(t) - 1} || \bm{x}^b - \bm{\mu}_i(t)||_2^2 
& \quad \text{, remove} \\
CP_i(t) + CP_j(t) + \dfrac{n_i(t)n_j(t)}{n_i(t) + n_j(t)} ||\bm{\mu}_j(t) - \bm{\mu}_i(t)||_2^2 
& \quad \text{, merge}  \\
CP_i(t) - CP_j(t) - \dfrac{n_i(t) n_j(t)}{n_i(t) - n_j(t)}||\bm{\mu}_j(t) - \bm{\mu}_i(t)||_2^2
& \quad \text{, split} \\
CP_i(t)
& \quad \text{, no operation}
\end{cases}
\label{Eq:comp_update}
\end{equation}

\begin{itemize}
\item Covariance matrix ($\bm{\Sigma} \in \mathbb{R}^{d \times d}$):
\end{itemize}
\begin{equation}
\bm{\Sigma}_i(t+1) =
\begin{cases}
\dfrac{n_i(t) - 1}{n_i(t)} \bm{\Sigma}_i(t) + \dfrac{1}{n_i(t) + 1} \left[ \bm{x}^b - \bm{\mu}_i(t) \right] \left[\bm{x}^b - \bm{\mu}_i(t) \right]^T
& \quad \text{, add} \\
\dfrac{n_i(t) - 1}{n_i(t) - 2} \bm{\Sigma}_i(t) - \dfrac{n_i(t)}{[n_i(t)-1][n_i(t)-2]}  \left[ \bm{x}^b - \bm{\mu}_i(t) \right] \left[\bm{x}^b - \bm{\mu}_i(t) \right]^T
& \quad \text{, remove} \\
\dfrac{n_i(t) - 1}{n_i(t) + n_j(t) - 1} \bm{\Sigma}_i(t) 
+ \dfrac{n_j(t) - 1}{n_i(t) + n_j(t) - 1} \bm{\Sigma}_j(t) \\
+ \dfrac{n_i(t) n_j(t)}{[n_i(t) + n_j(t)] [n_i(t) + n_j(t) - 1]}  \left[ \bm{\mu}_j(t) - \bm{\mu}_i(t) \right] \left[\bm{\mu}_j(t) - \bm{\mu}_i(t) \right]^T
& \quad \text{, merge}  \\
\dfrac{n_i(t) - 1}{n_i(t) - n_j(t) - 1} \bm{\Sigma}_i(t)  
- \dfrac{n_j(t) - 1}{n_i(t) - n_j(t) - 1} \bm{\Sigma}_j(t) \\
- \dfrac{n_i(t) n_j(t)}{[n_i(t) - n_j(t)] [n_i(t) - n_j(t) - 1]} \left[ \bm{\mu}_j(t) - \bm{\mu}_i(t) \right] \left[\bm{\mu}_j(t) - \bm{\mu}_i(t) \right]^T
& \quad \text{, split} \\
\bm{\Sigma}_i(t)
& \quad \text{, no operation}
\end{cases}
\label{Eq:cov_update}
\end{equation}

Note that the add and remove operations are specific cases of merge and split, i.e., the latter reduce to the former when $\bm{\mu}_j = \bm{x}^b$, $n_j = 1$ ($CP_j = 0$ and defining $\bm{\Sigma}_j = 0$). Also, note that the incremental update for the add operation is much simpler than the hard version of~\cite{Moshtaghi2018b} in~\cite{leonardo.2020a} (cf. Eq.~(\ref{Eq:iCP_1}) and the add case of Eq.~(\ref{Eq:comp_update})) because for the definition of $\bm{g}$ of the iCVIs used in this work, $\bm{g} \triangleq \Vec{\bm{0}}$. In particular, we employ these updates for the sum-of-squares-based (iCH, iWB, iDB, iXB, and iPBM) and information-theoretic-based (iNI) iCVIs listed in Table~\ref{Tab:CVI_summary}.

\subsection{Training} \label{sec:training}

The original data set $\bm{X} \in \mathbb{R}^{N \times d}$ is first duplicated. The first copy undergoes min-max normalization ($\bm{X}_{min-max} \in [0,1]^d$) and complement coding~\cite{carpenter1992} ($\bm{X}_{cc} = \left[ \bm{X}_{min-max}, 1 - \bm{X}_{min-max} \right] \in \mathbb{R}^{N \times 2d}$) in order to be presented to ARTa ($\bm{X}^a = \bm{X}_{cc}$). The other copy undergoes standardization ($\bm{X}_{std}$ with zero mean and unit variance) and is used to perform the iCVI computations ($\bm{X}^b = \bm{X}_{std}$). Next, like some other iterative algorithms, initial prototypes and an initial partition must be defined. Specifically, $k$ ARTa categories are initialized using kmeans on $\bm{X}_{std}$ with k-means++ initialization~\cite{David.2007a}. The $k$ obtained centroids $\bm{\mu} \in \mathbb{R}^{k \times d}$ undergo inverse standardization (using the scaling parameters that generated $\bm{X}_{std}$), min-max normalization (using the scaling parameters that generated $\bm{X}_{min-max}$) and complement coding; then, the ARTa categories ($\bm{W}_{a} \in \mathbb{R}^{k \times 2d}$) are set to these transformed centroids. An initial partition is defined by assigning all samples to their closest category using Eq.~(\ref{Eq:FA_T}). Finally, the map field mapping matrix is initialized to a matrix of 1s (i.e., $\bm{W}^{ab} = \bm{1} \in \mathbb{R}^{k \times k}$), and the iCVI value and variables are initialized in batch mode using the initial partition $\Omega(0)$ and $\bm{X}_{cc}$. Note that the variables in this step may differ according to the chosen iCVI (frequencies, means, compactness or covariances, pairwise distances, etc.).

During the training of the iCVI-ARTMAP model, for each sample presentation, the iCVI-framework computes the value of assigning such sample to each one of the existing clusters using the incremental formulations described in the previous section, along with the cached variables. The optimal assignment at that time step is used to generate a one-hot encoding vector $\bm{y}$ representing the class of such sample at that particular time. Next, iCVI-ARTMAP follows the dynamics described in Section~\ref{sec:ART}. The new label for that presented sample is set using the map field weights:
\begin{equation}
l = \argmax\limits_i \bm{w}_{J,i}^{ab},
\label{Eq:mf_pred}
\end{equation}
\noindent where $l$ is the new label assigned to the presented sample $\bm{x}$ that resonated with category $J$ of ARTa (i.e., the category that satisfied both $\rho_a$ and $\rho_{ab}$). Note that $l$ is not necessarily equal to the label encoded by $\bm{y}$. 

The iCVI-ARTMAP model also features shrinkage and pruning of the fuzzy ARTa's categories. If the current sample swaps assignment from ARTa's category $r$ to category $J$ (i.e., category $r$ is no longer the best match for that sample), then the category $r$ weight vector shrinks to:
\begin{equation}
\bm{w}_r = \bigwedge\limits_{\bm{x}_i \in \bm{w}_r}\bm{x}_i,
\label{Eq:shrinkage}
\end{equation}
\noindent where $\bm{x}_i$ represents all the samples that remain assigned to category~$r$; and as a consequence, fuzzy ARTa loses its stability property. If ARTa's category $r$ does not have any samples assigned to it, then it is pruned. At the end of each epoch, clusters are merged and split as discussed in Section~\ref{Sec:split_and_merge}. When performing shrinking, pruning, merging or splitting, the iCVI-ARTMAP variables must be appropriately adjusted to reflect the resulting changes. Finally, the following stopping criteria are checked: 
\begin{enumerate}
\item Reaching a predefined maximum number of epochs~($E$).
\item No change in fuzzy ARTa's weight vectors ($\bm{w}$) between two consecutive epochs. \item The difference between the iCVI values between two consecutive epochs is less than or equal to a predefined convergence parameter~($tol$).
\end{enumerate}

The pseudo-code in Algorithm~\ref{alg:icvi-artmap} summarizes the main steps of the iCVI-ARTMAP training procedure.

\begin{algorithm}[!p]
\DontPrintSemicolon
\SetKwInOut{KwIn}{Input}
\SetKwInOut{KwOut}{Output}
\SetKwComment{Comment}{}{}
\LinesNumbered
\KwIn{data ($\bm{X}$), number of clusters ($k$), ARTa parameters ($\rho_a$, $\beta_a$, $\alpha_a$), map field parameters ($\rho_{ab}$, $\beta_{ab}$, $\epsilon$), iCVI framework parameter (iCVI), as well as $tol$ and $E$.}
\KwOut{cluster labels}
\algrule
\tcc{Notation}
\Comment*[l]{$t$: iteration.}
\Comment*[l]{$\Omega(t)$: data partition at iteration $t$.}
\Comment*[l]{$J(\bm{x}^a, t)$: ARTa resonant category for sample $\bm{x}^a$ at time $t$.} 
\Comment*[l]{$\mathcal{J} = \{J(\bm{x}_i^a, t)\}_{i=1}^N$: ARTa resonant categories for each sample in $\bm{X}^a$ at iteration $t$.} 
\Comment*[l]{$k'(t)$: number of clusters in iCVI-ARTMAP at iteration $t$ ($\bm{W}^{ab} \in \mathbb{R}^{C_a \times k'(t)}$).} 
\tcc{Pre-processing}
\label{ARTa_pre}Generate ARTa inputs: $\bm{X}^{a} = \bm{X}_{cc}$ (min-max normalization and complement coding). \\
\label{iCVI_pre}Generate iCVI framework inputs: $\bm{X}^{b} = \bm{X}_{std}$ (standardization). \\
\tcc{Initialization}
$t \leftarrow 0$, $k'(t) \leftarrow k$\\
Initialize ARTa categories ($\bm{W}_a$) using the processed centroids from kmeans. \\ 
Initialize $\Omega(t)$ and $\mathcal{J}$ by presenting $\bm{X}^{a}$ to ARTa on feedforward mode without resonance. \\ 
Initialize ARTa instance counting of categories using $\Omega(t)$. \\ 
Initialize map field mapping matrix ($\bm{W}^{ab}$). \\
Initialize iCVI value and variables in batch mode using $\Omega(t)$ and $\bm{X}^{b}$.\\
\tcc{Training}
\While{stopping conditions not satisfied}{
\For{$(\bm{x}^a,\bm{x}^b) \in (\bm{X}^a,\bm{X}^b)$}{
$t \leftarrow t + 1$\\
Present $\bm{x}^b$ to iCVI-framework to generate the cluster label $\bm{y}(t)$. \\
Use the pair ($\bm{x}^a$, $\bm{y}(t)$) to carry out the learning dynamics of ARTa and map field.\\
Update $\Omega(t)$ using the prediction of the map field for $\bm{x}^a$ \\
\If(\tcp*[h]{i.e. sample $(\bm{x}^a,\bm{x}^b)$ changed cluster assignment}){$\Omega(t) \neq \Omega(t-1)$}{
Update iCVI value and variables. \\
}
\If{$J(\bm{x}^a, t) \neq J(\bm{x}^a, t-1)$}{
\uIf{$J(\bm{x}^a, t-1) \notin \mathcal{J}$}{
Prune category $J(\bm{x}^a, t-1)$. \\
}
\Else{
Shrink category $J(\bm{x}^a, t-1)$. \\
}
}
\If(\tcp*[h]{i.e. cluster disappeared}){$\exists i \in \{1,...,k\} ~:~ i~\notin~\Omega(t)$}{
Adjust the appropriate iCVI-ARTMAP variables. \\
}
}
\While(\tcp*[h]{i.e. merge improves the iCVI value}){$k'(t) > 2$ AND it is possible to merge}{
Merge clusters. \\
}
\While(\tcp*[h]{i.e. multi-prototype clusters}){$k - k'(t) \neq 0$ AND it is possible to split}{
Split clusters. \\
}
}
\caption{iCVI-ARTMAP}\label{alg:icvi-artmap}
\end{algorithm}

\section{Experiments} \label{sec:experiments}

\subsection{Data sets} \label{sec:data}

\begin{figure}[!t]
\centering
\begin{subfigure}[b]{0.24\textwidth}
\centering
\includegraphics[width=\textwidth]{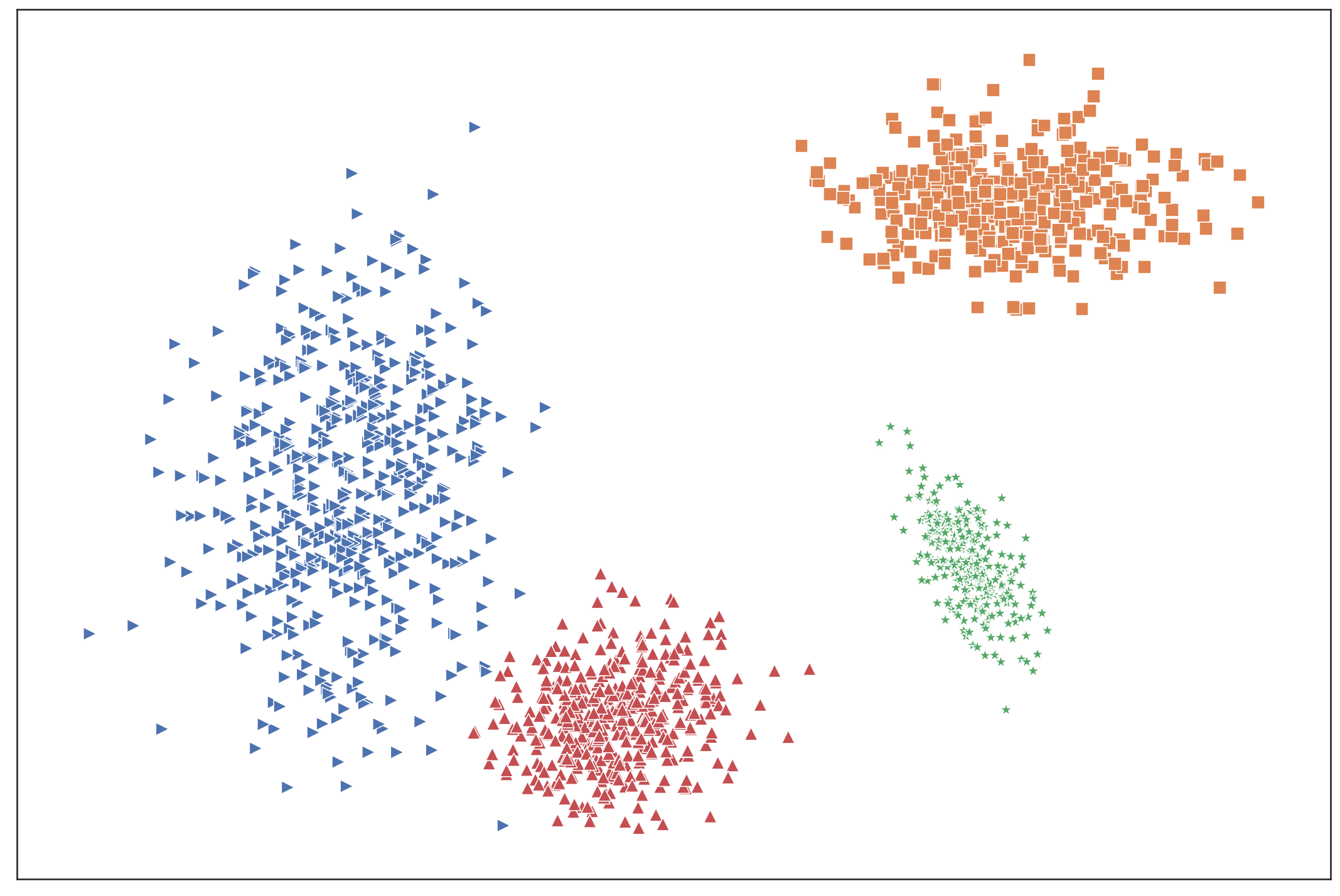}
\caption{2d-4c-no0}
\end{subfigure}
\hfill
\begin{subfigure}[b]{0.24\textwidth}
\centering
\includegraphics[width=\textwidth]{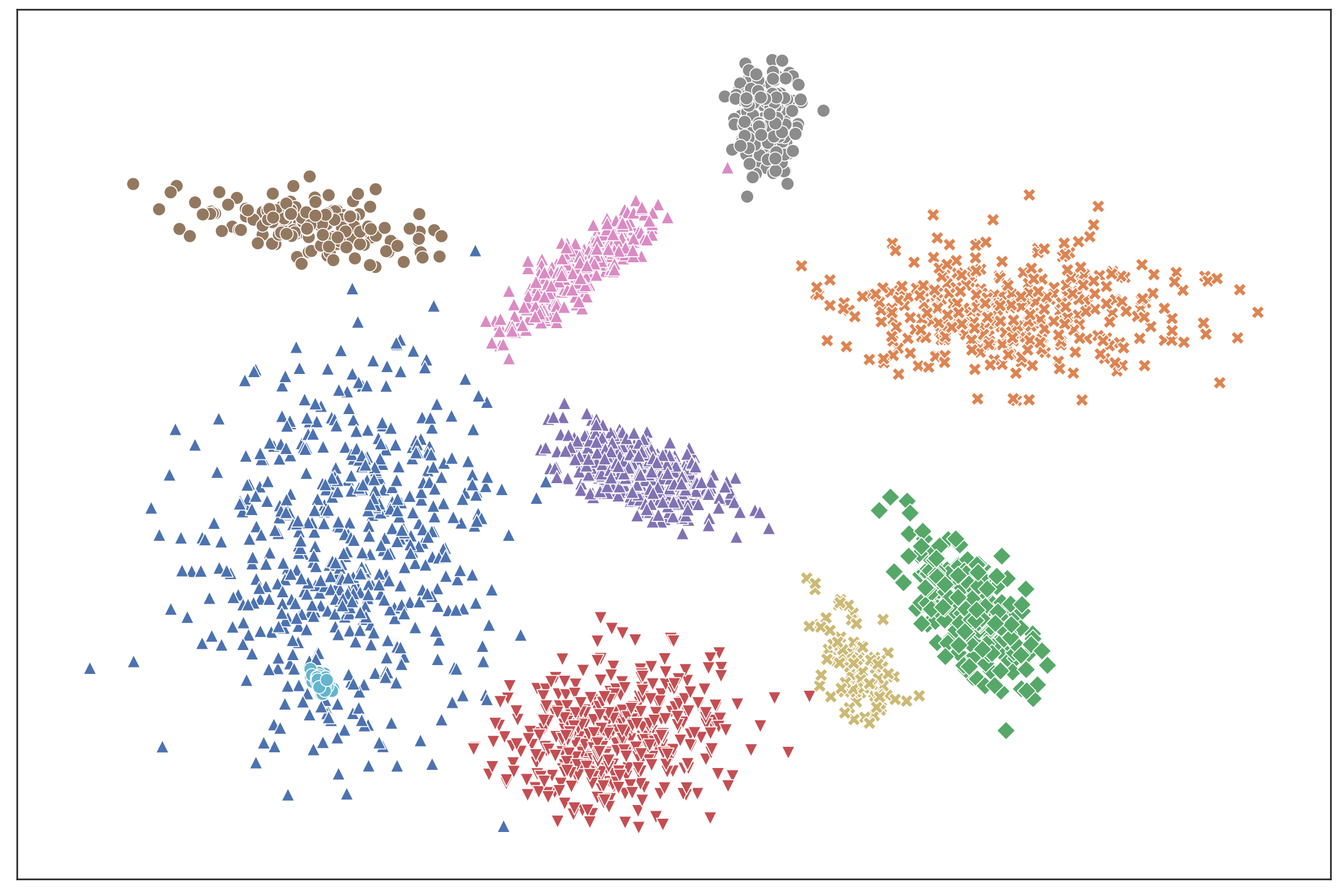}
\caption{2d-10c-no0}
\end{subfigure}
\hfill
\begin{subfigure}[b]{0.24\textwidth}
\centering
\includegraphics[width=\textwidth]{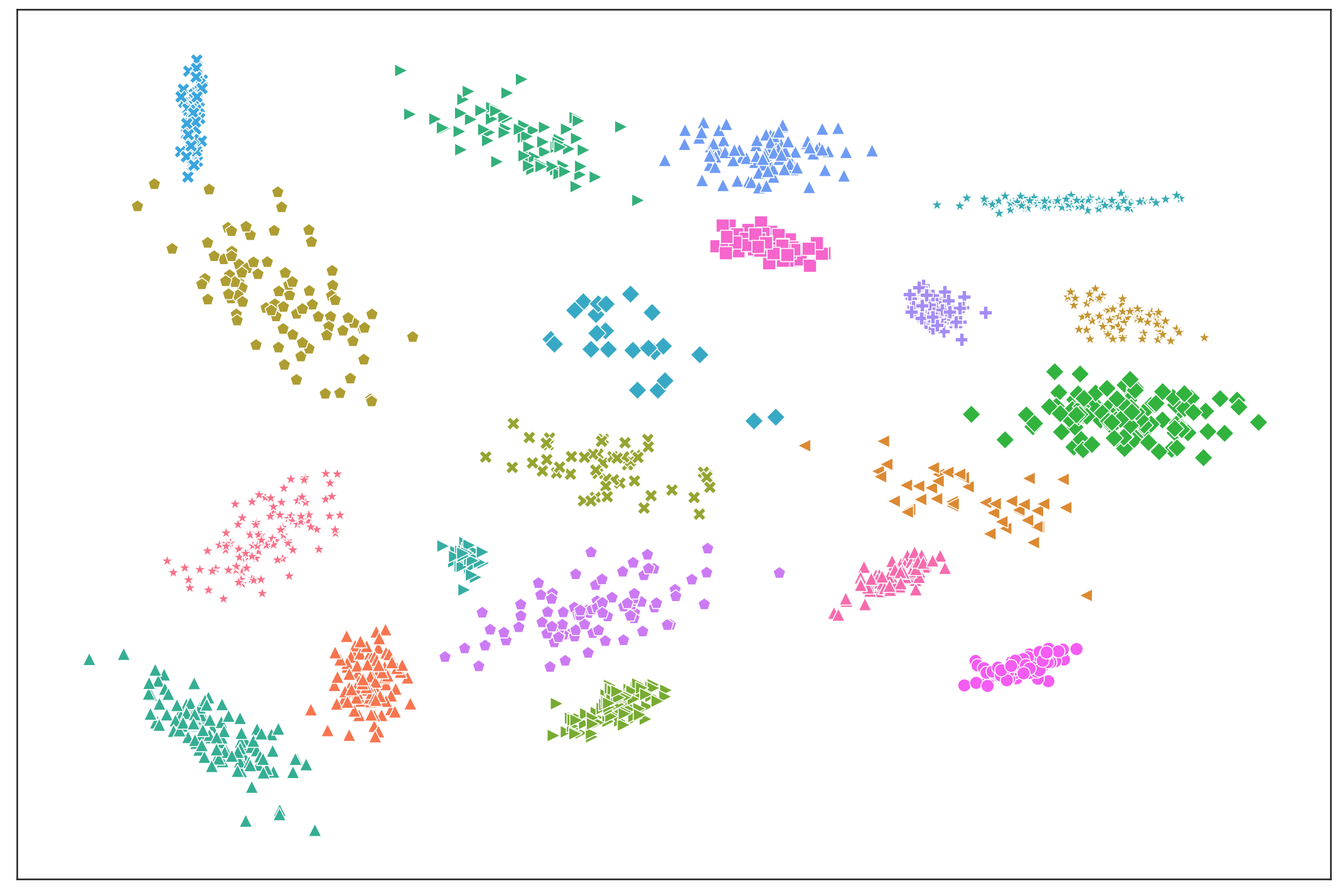}
\caption{2d-20c-no0}
\end{subfigure}
\hfill
\begin{subfigure}[b]{0.24\textwidth}
\centering
\includegraphics[width=\textwidth]{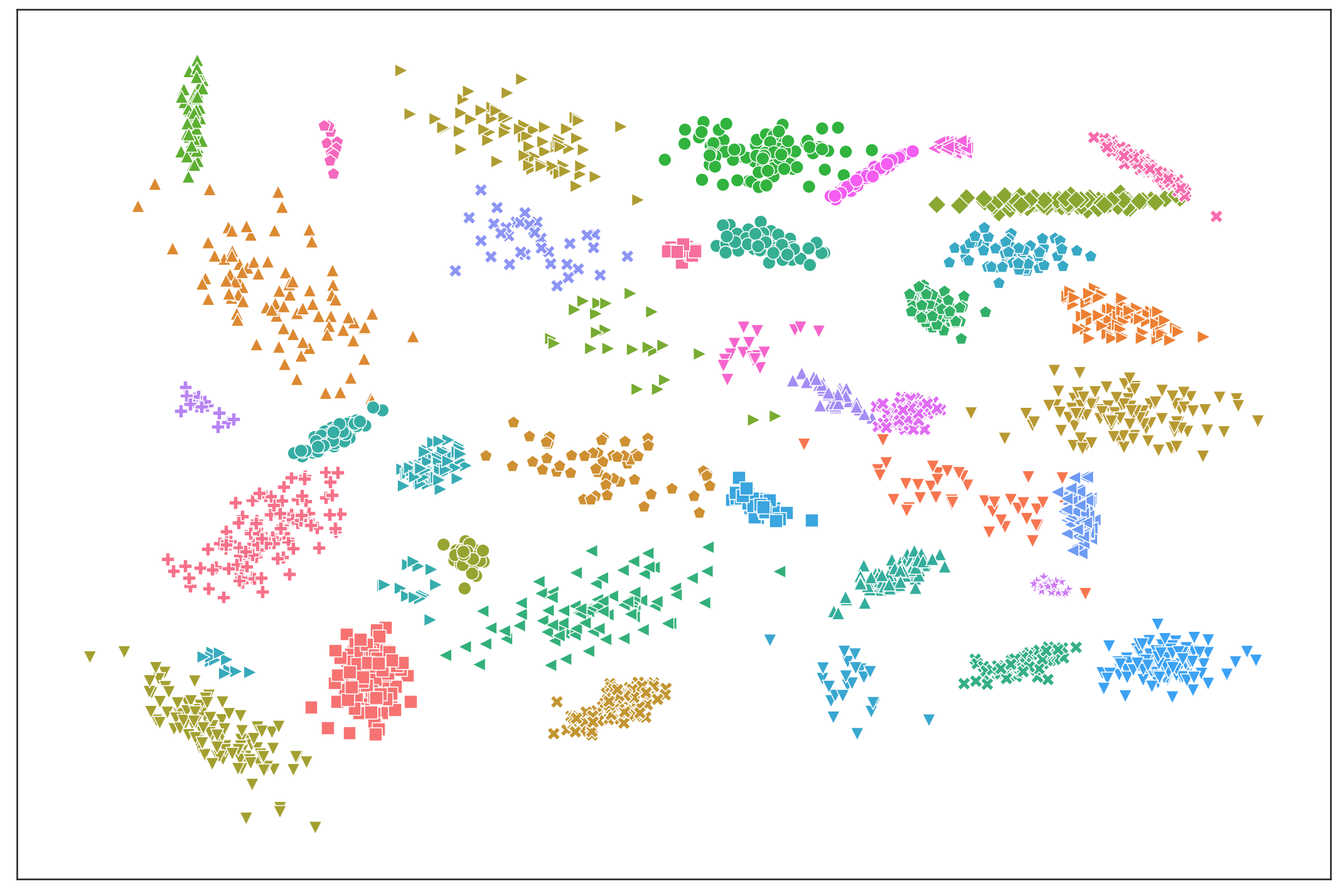}
\caption{2d-40c-no0}
\end{subfigure}

\begin{subfigure}[b]{0.24\textwidth}
\centering
\includegraphics[width=\textwidth]{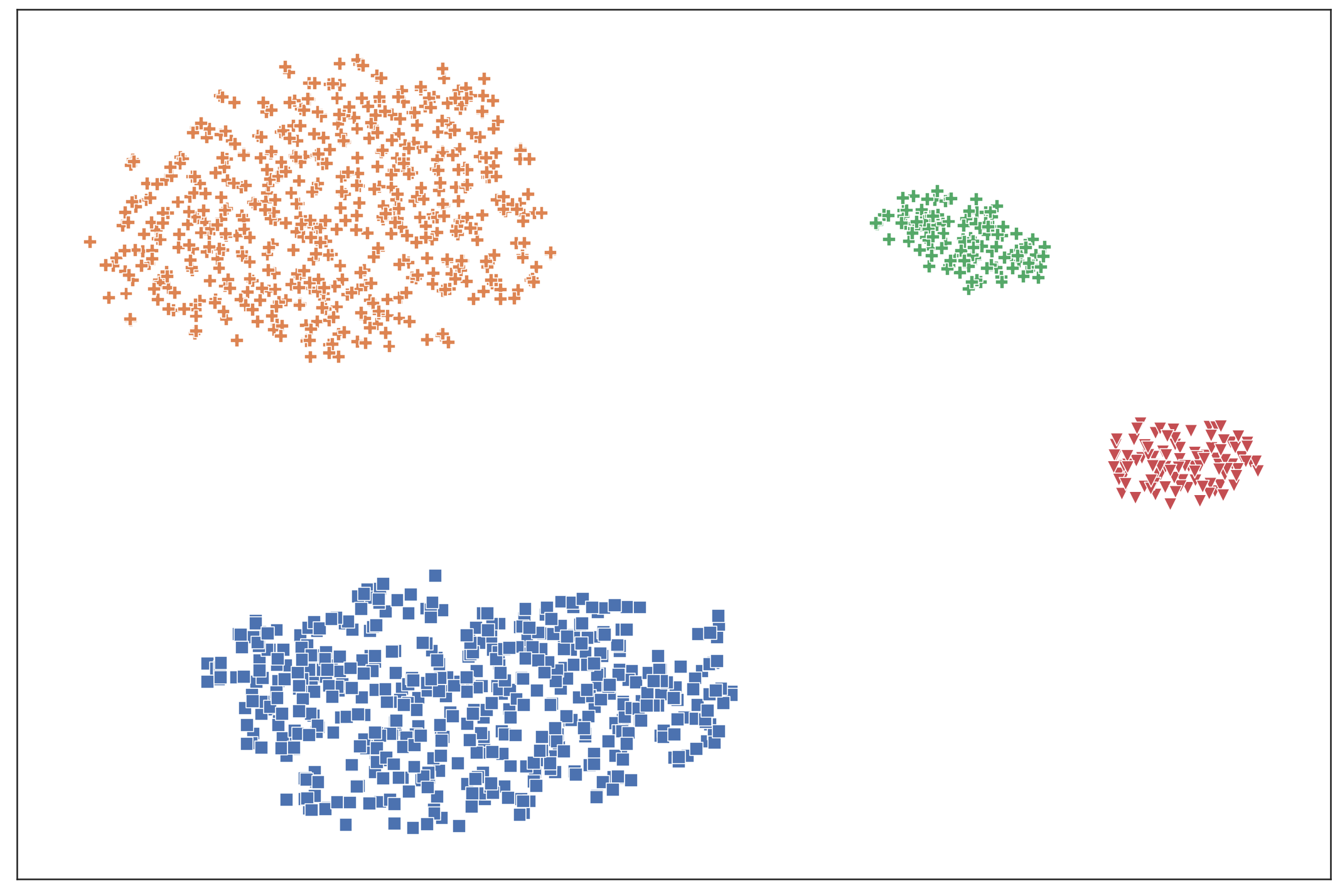}
\caption{10d-4c-no0}
\end{subfigure}
\hfill
\begin{subfigure}[b]{0.24\textwidth}
\centering
\includegraphics[width=\textwidth]{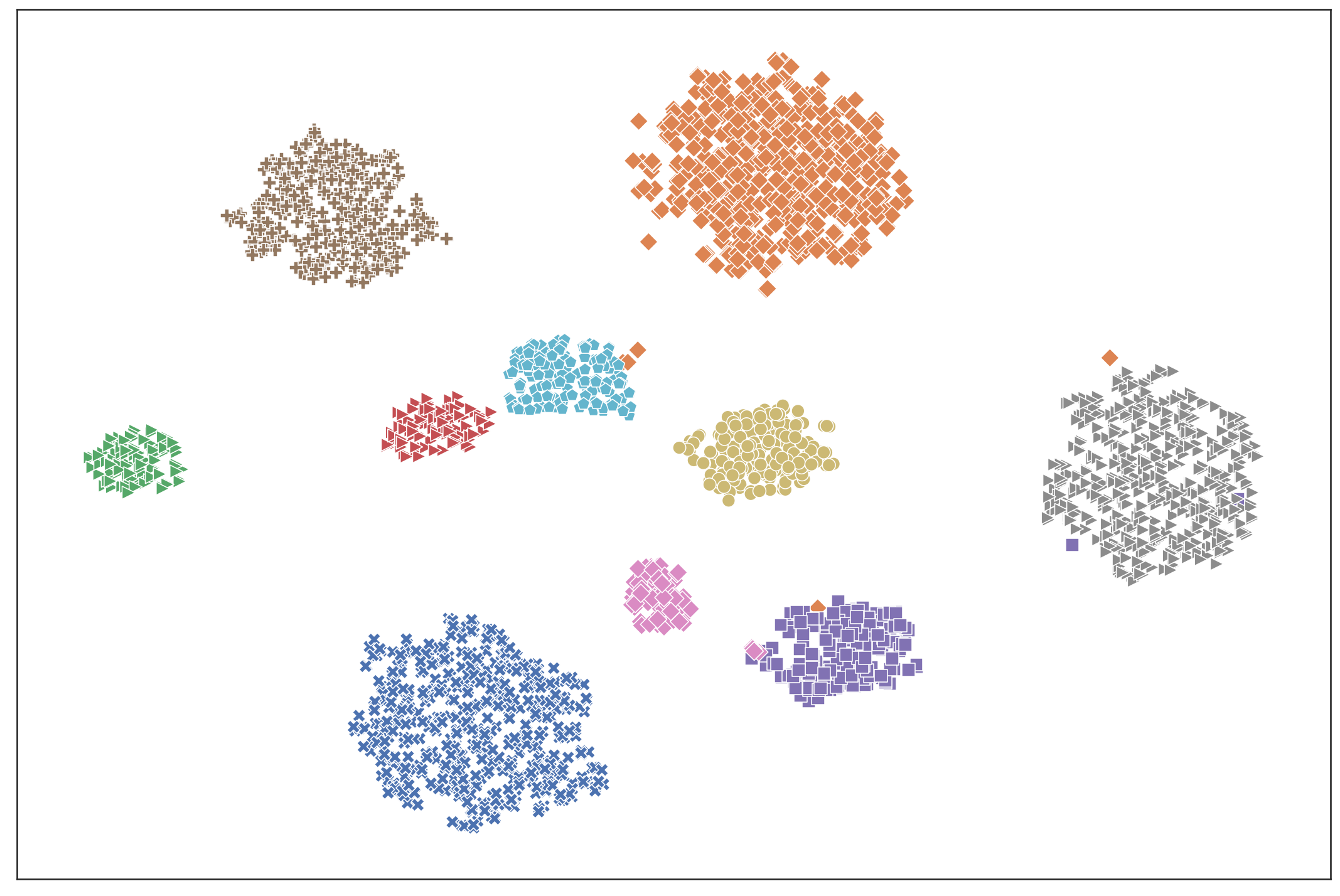}
\caption{10d-10c-no0}
\end{subfigure}
\hfill
\begin{subfigure}[b]{0.24\textwidth}
\centering
\includegraphics[width=\textwidth]{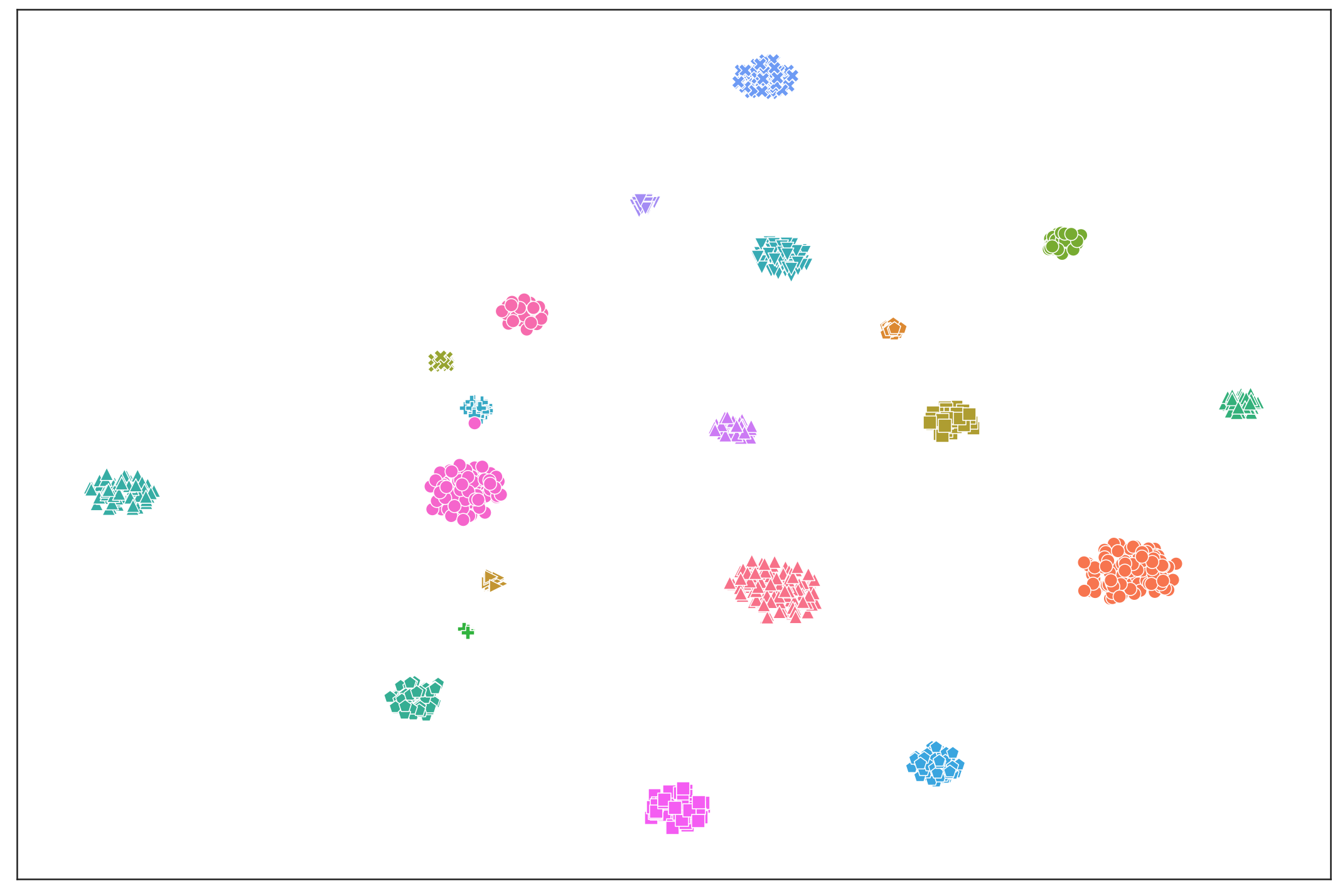}
\caption{10d-20c-no0}
\end{subfigure}
\hfill
\begin{subfigure}[b]{0.24\textwidth}
\centering
\includegraphics[width=\textwidth]{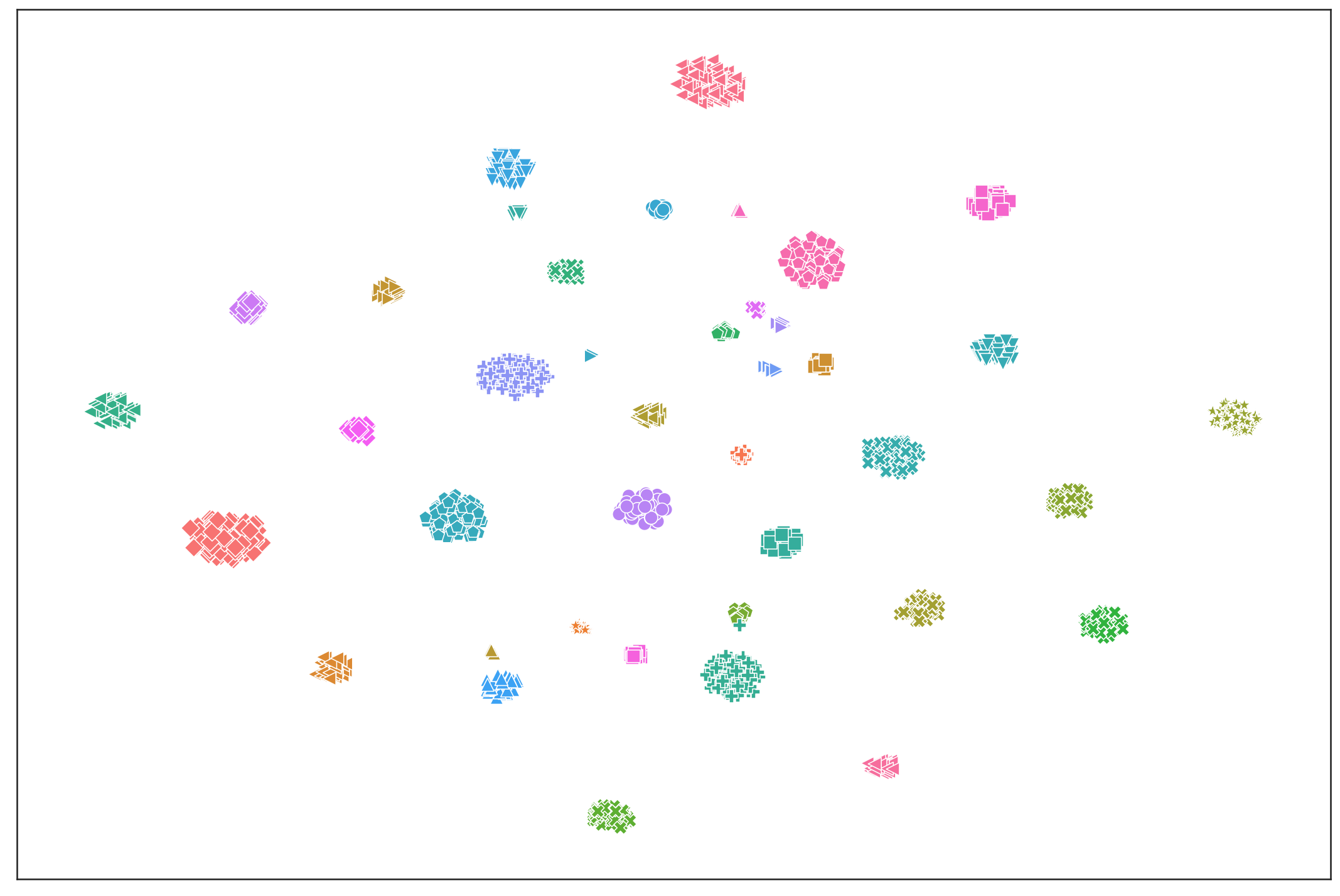}
\caption{10d-40c-no0}
\end{subfigure}

\begin{subfigure}[b]{0.24\textwidth}
\centering
\includegraphics[width=\textwidth]{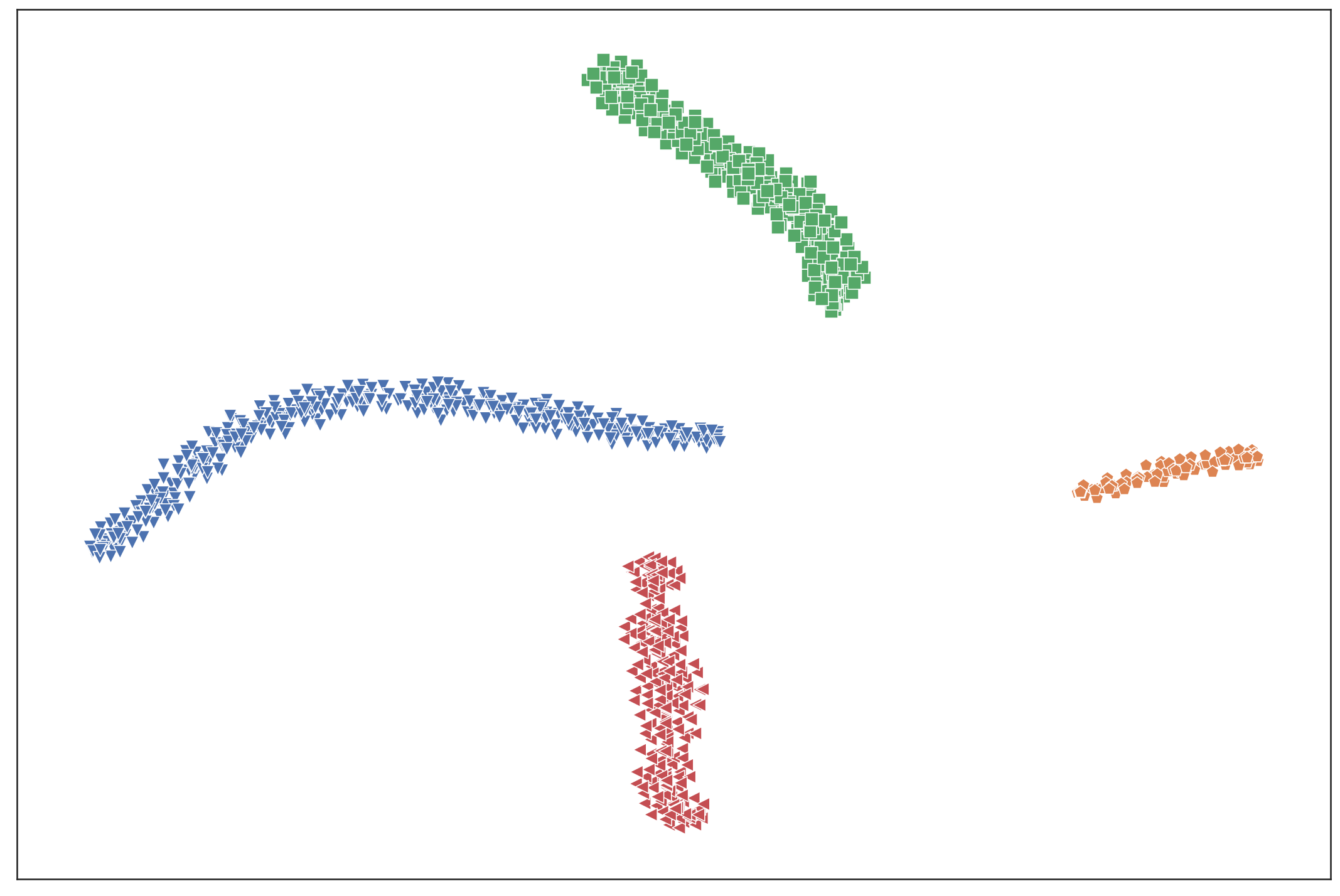}
\caption{ellipsoid.50d4c.1}
\end{subfigure}
\hfill
\begin{subfigure}[b]{0.24\textwidth}
\centering
\includegraphics[width=\textwidth]{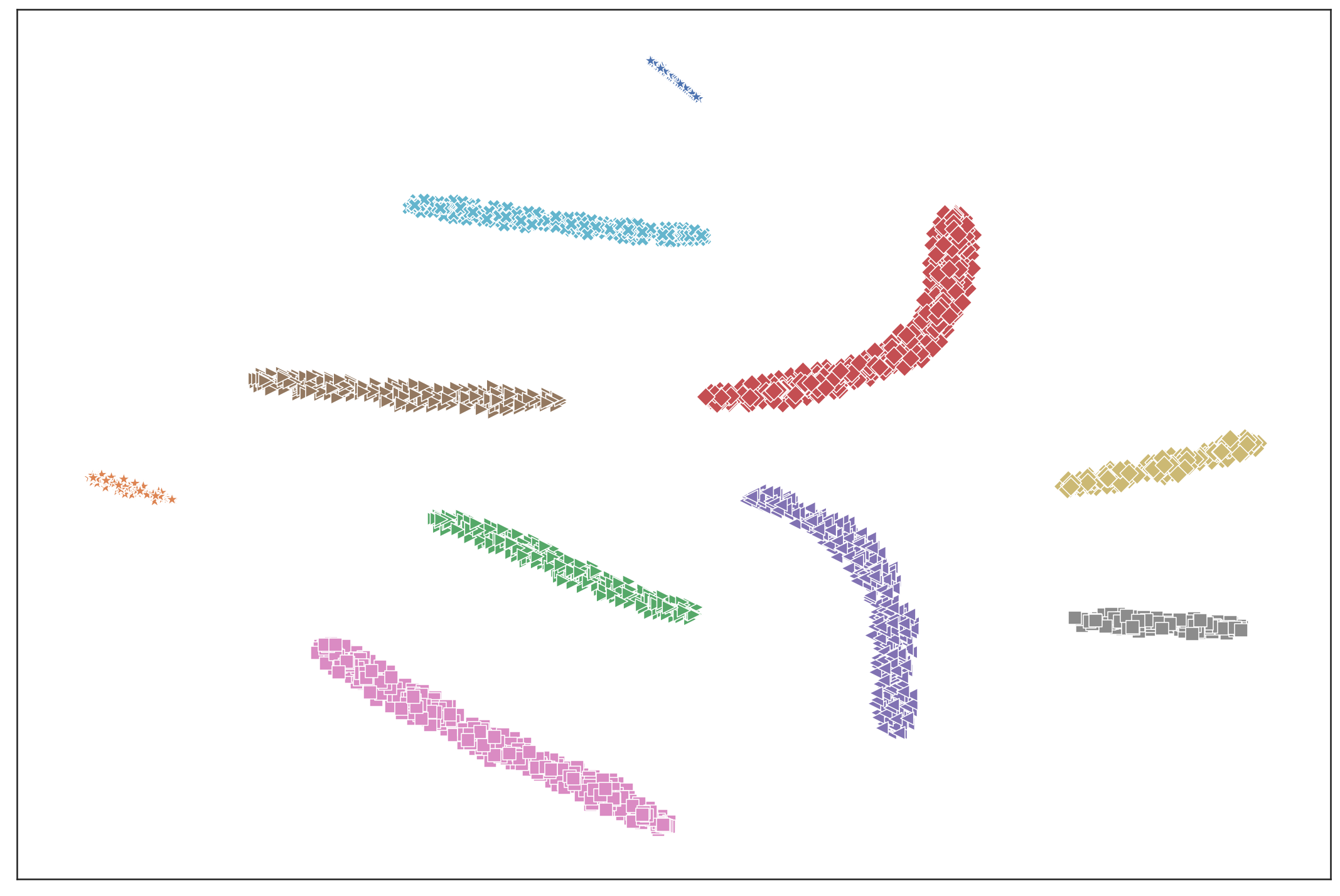}
\caption{ellipsoid.50d10c.1}
\end{subfigure}
\hfill
\begin{subfigure}[b]{0.24\textwidth}
\centering
\includegraphics[width=\textwidth]{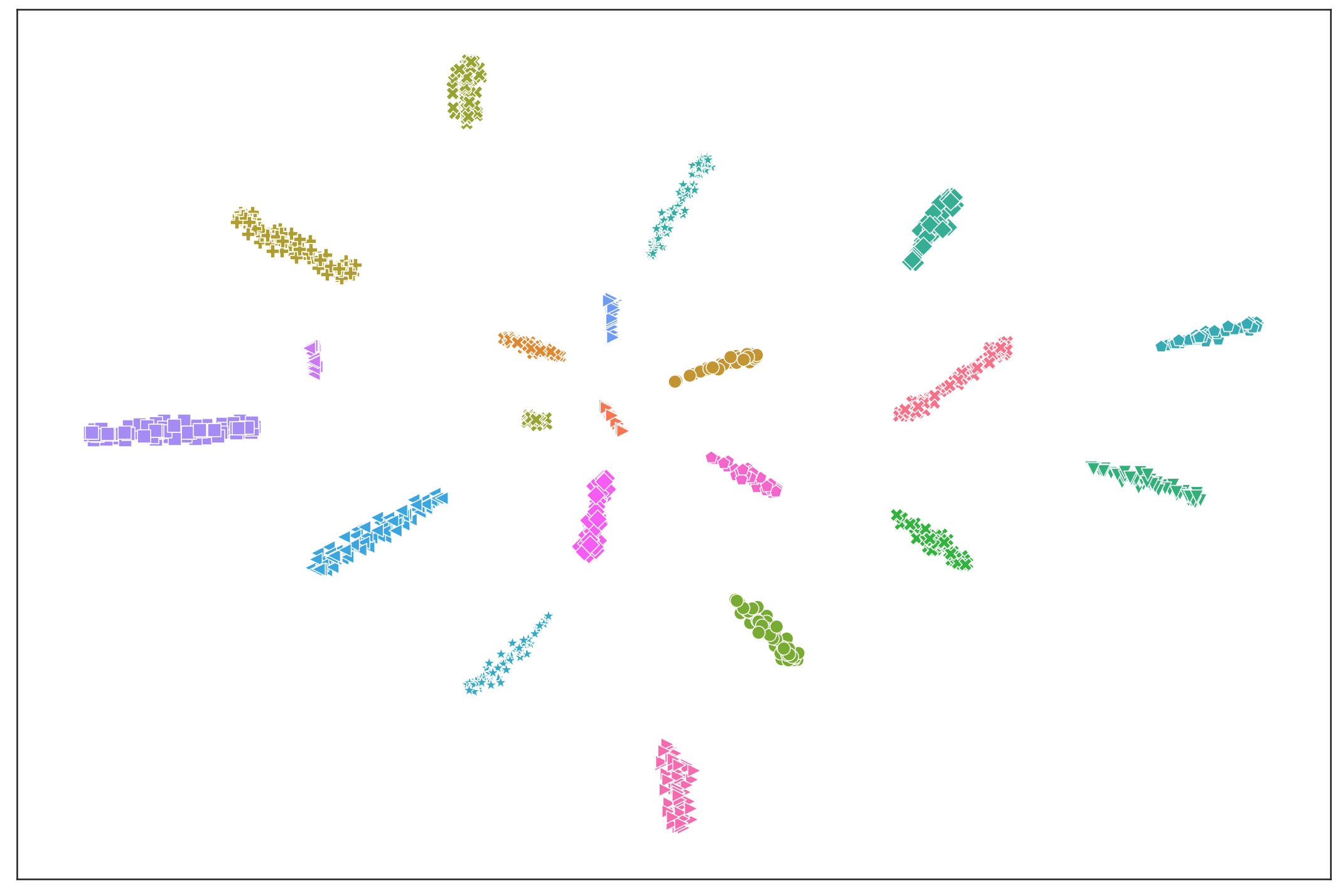}
\caption{ellipsoid.50d20c.1}
\end{subfigure}
\hfill
\begin{subfigure}[b]{0.24\textwidth}
\centering
\includegraphics[width=\textwidth]{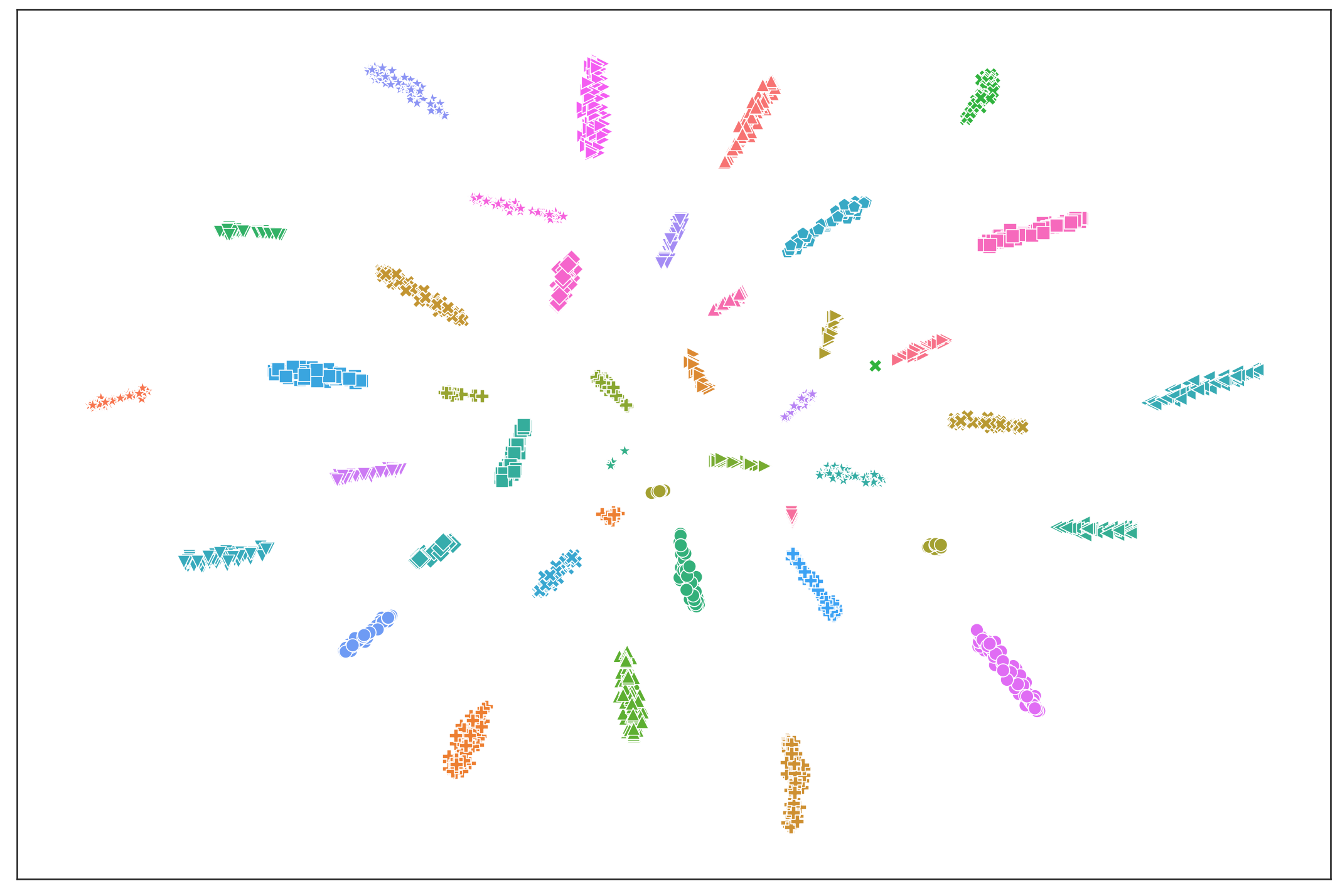}
\caption{ellipsoid.50d40c.1}
\end{subfigure}

\begin{subfigure}[b]{0.24\textwidth}
\centering
\includegraphics[width=\textwidth]{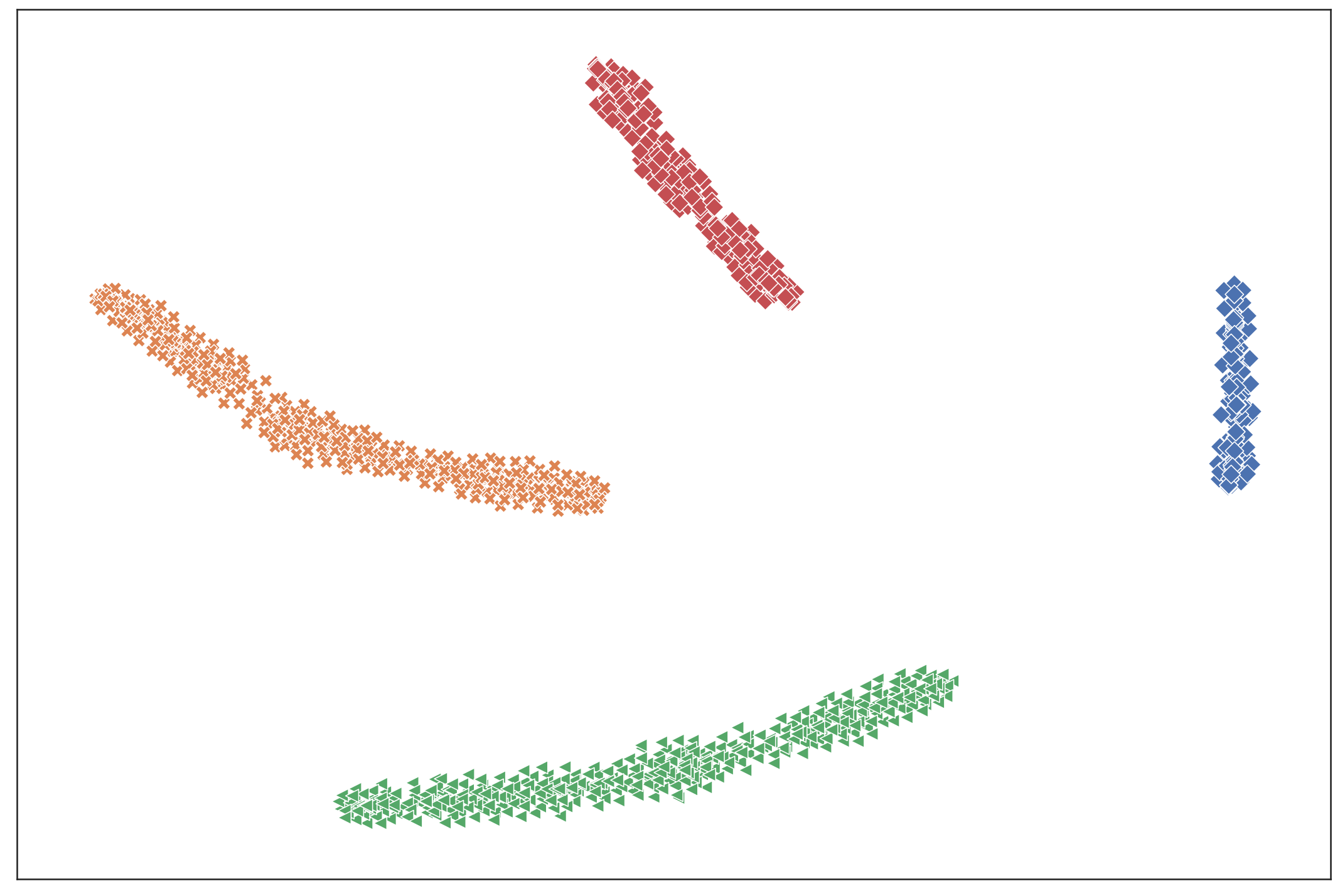}
\caption{ellipsoid.100d4c.1}
\end{subfigure}
\hfill
\begin{subfigure}[b]{0.24\textwidth}
\centering
\includegraphics[width=\textwidth]{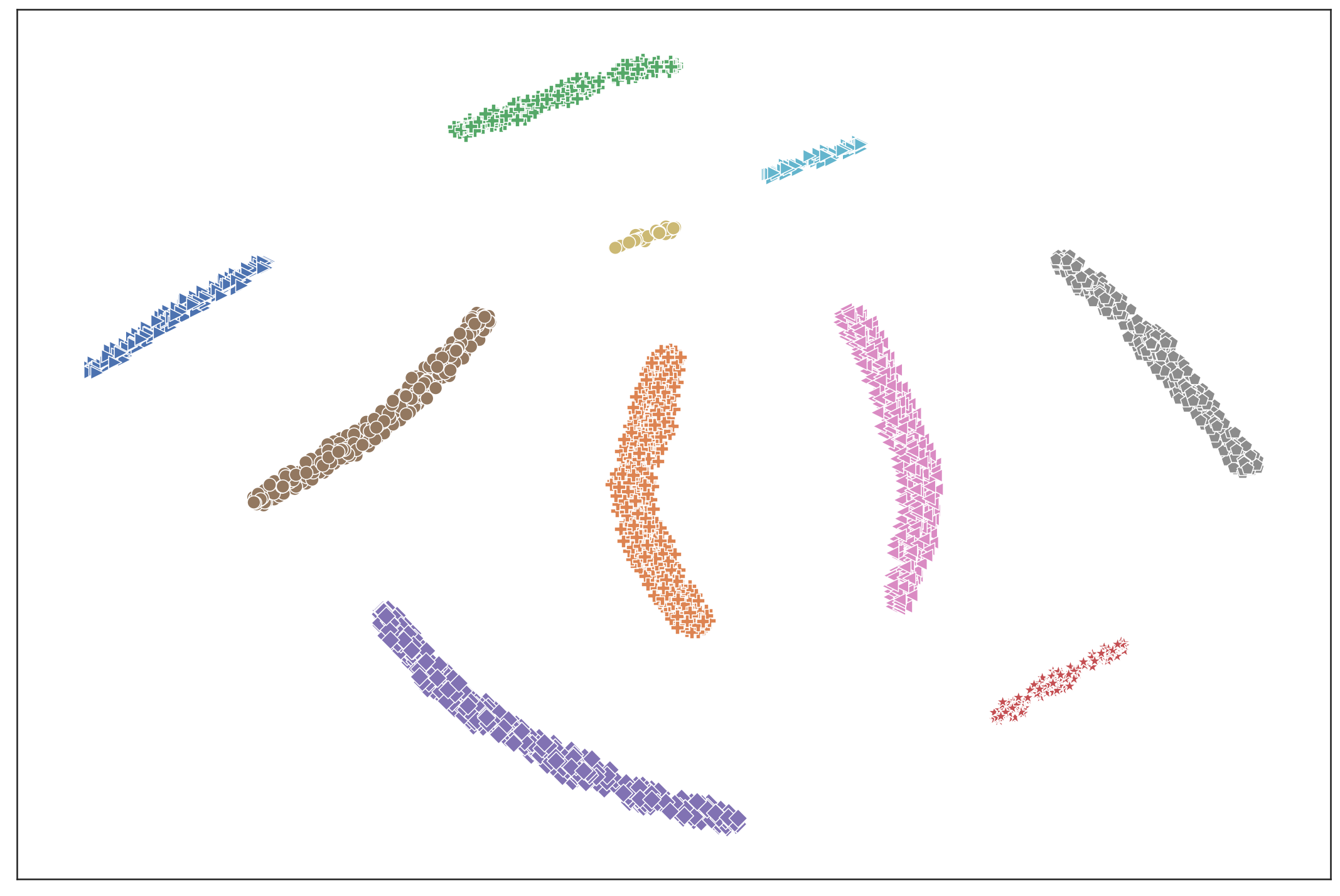}
\caption{ellipsoid.100d10c.1}
\end{subfigure}
\hfill
\begin{subfigure}[b]{0.24\textwidth}
\centering
\includegraphics[width=\textwidth]{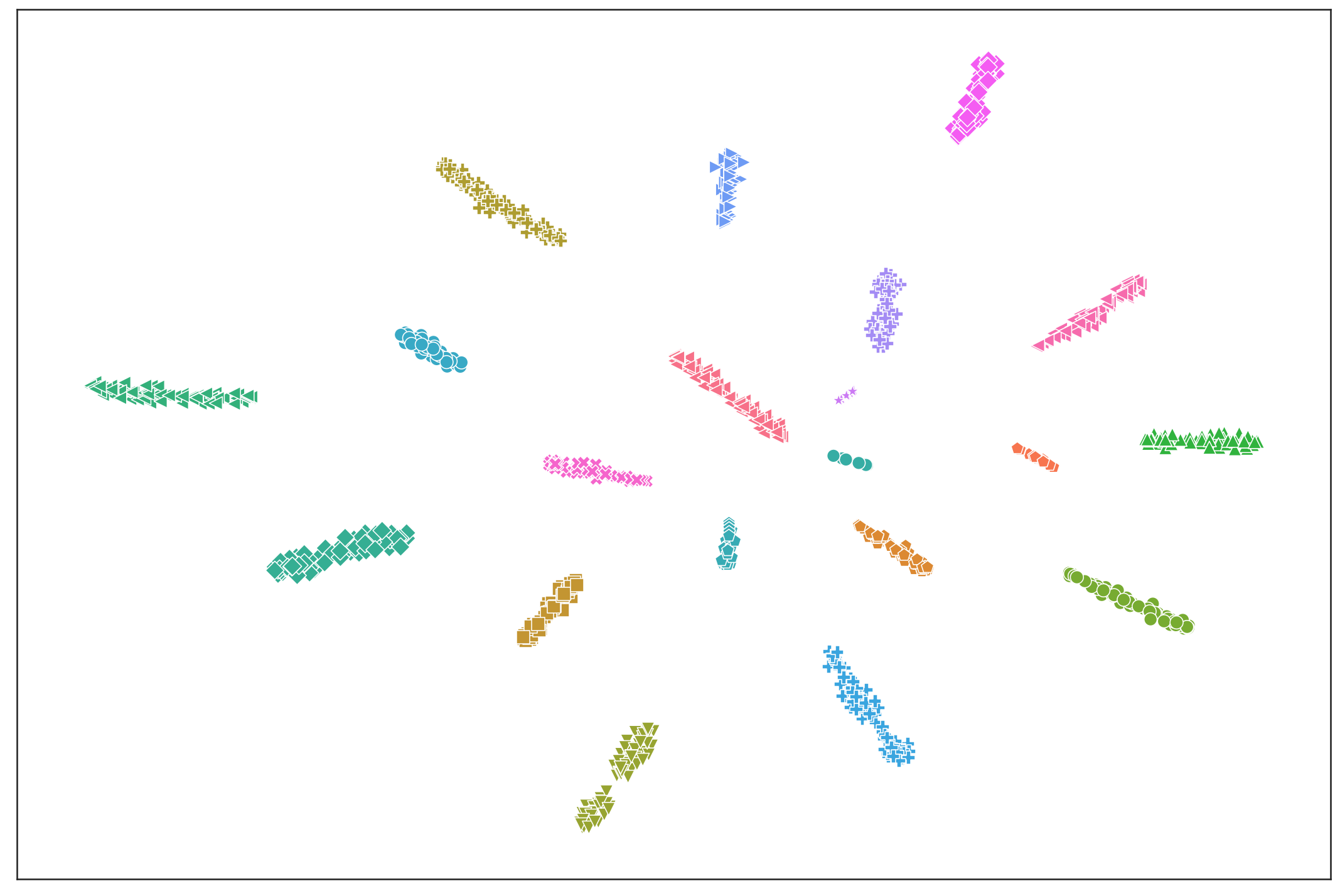}
\caption{ellipsoid.100d20c.1}
\end{subfigure}
\hfill
\begin{subfigure}[b]{0.24\textwidth}
\centering
\includegraphics[width=\textwidth]{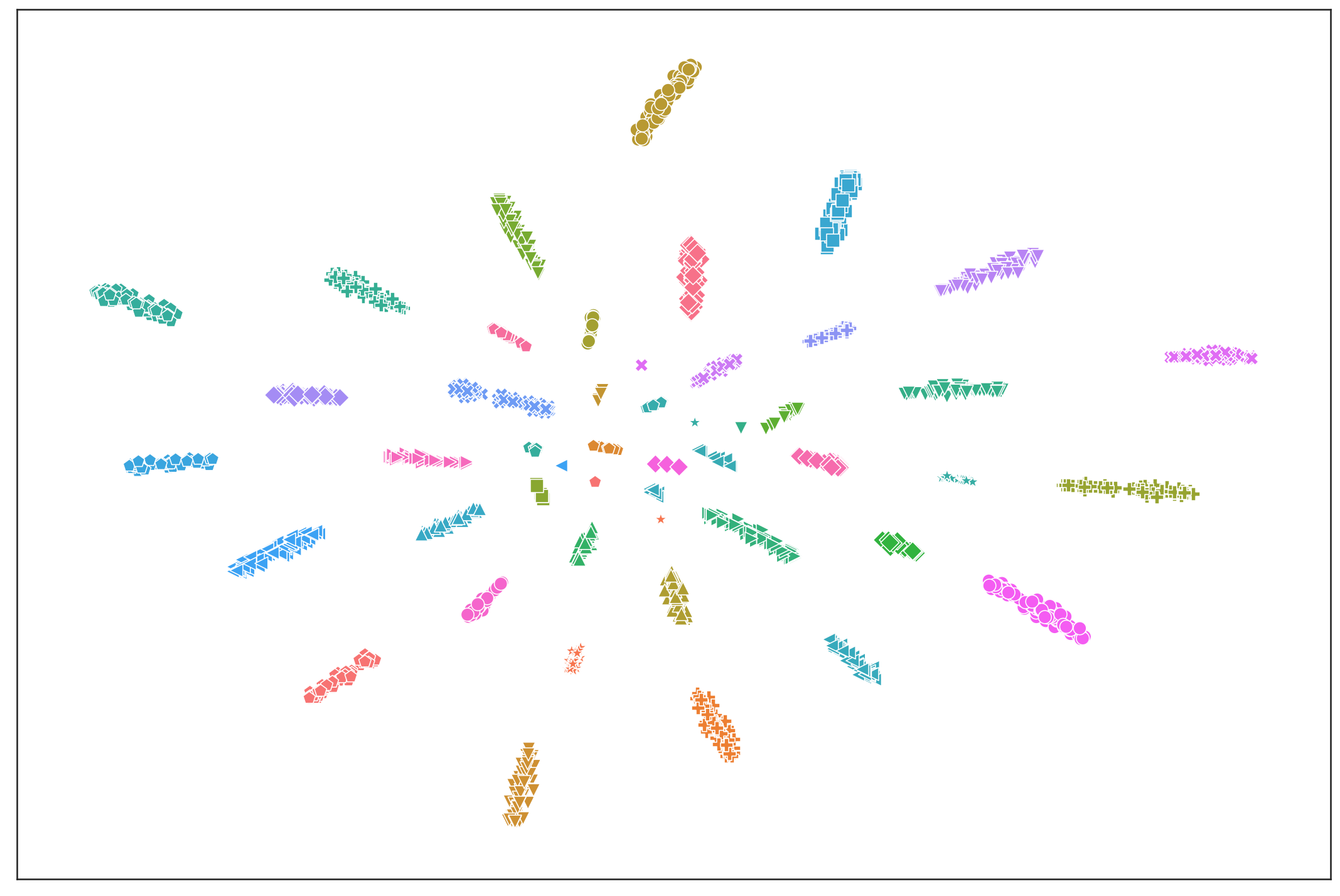}
\caption{ellipsoid.100d40c.1}
\end{subfigure}
\caption{Synthetic data sets used in the experiments. The t-SNE~\cite{Maaten.2008a} method was used to project data sets with dimensionality greater than 2.}
\label{fig:datasets_1}
\end{figure}

\begin{figure}[!t]
\centering
\begin{subfigure}[b]{0.24\textwidth}
\centering
\includegraphics[width=\textwidth]{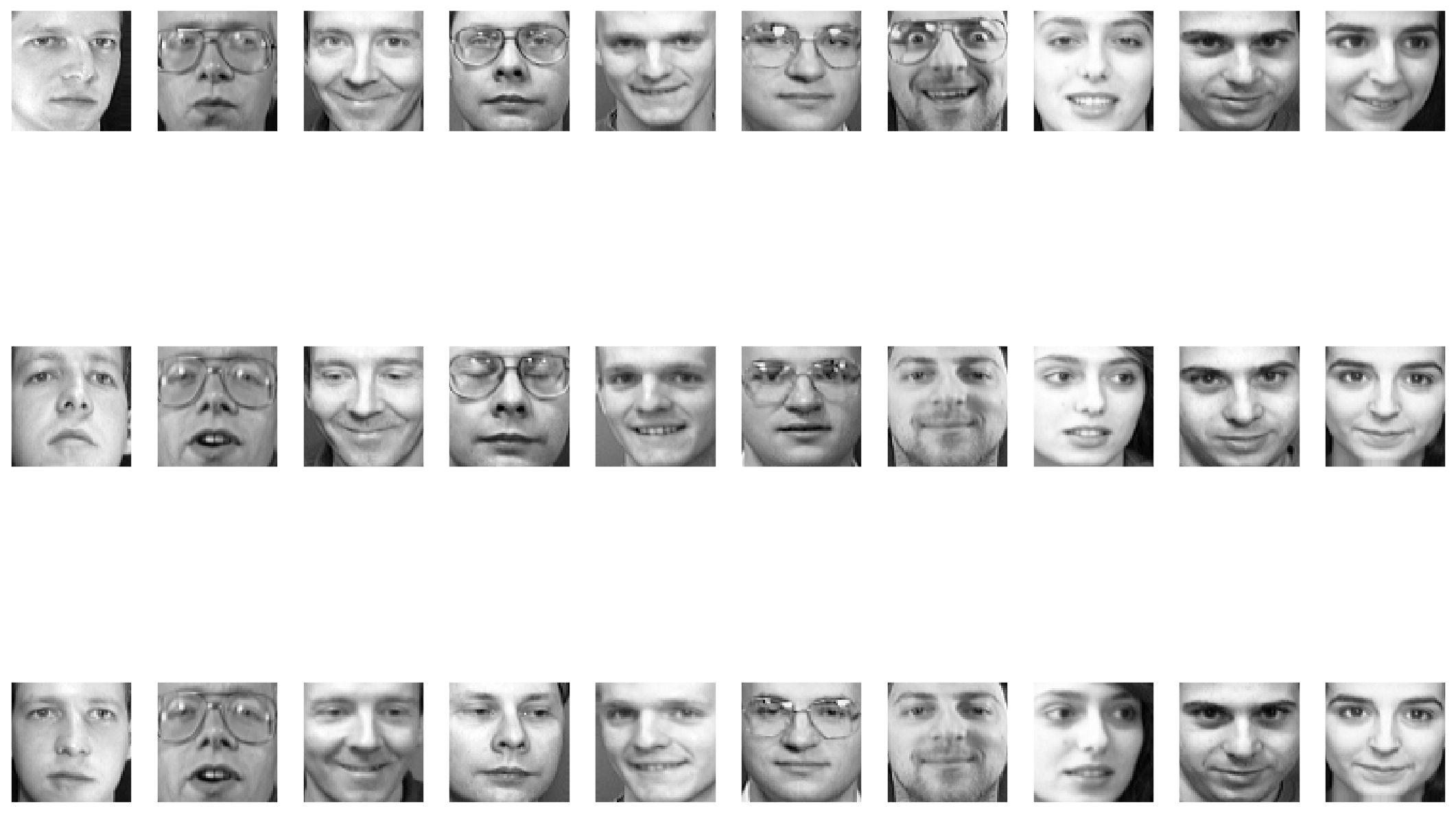}
\caption{Olivetti faces}
\end{subfigure}
\hfill
\begin{subfigure}[b]{0.24\textwidth}
\centering
\includegraphics[width=\textwidth]{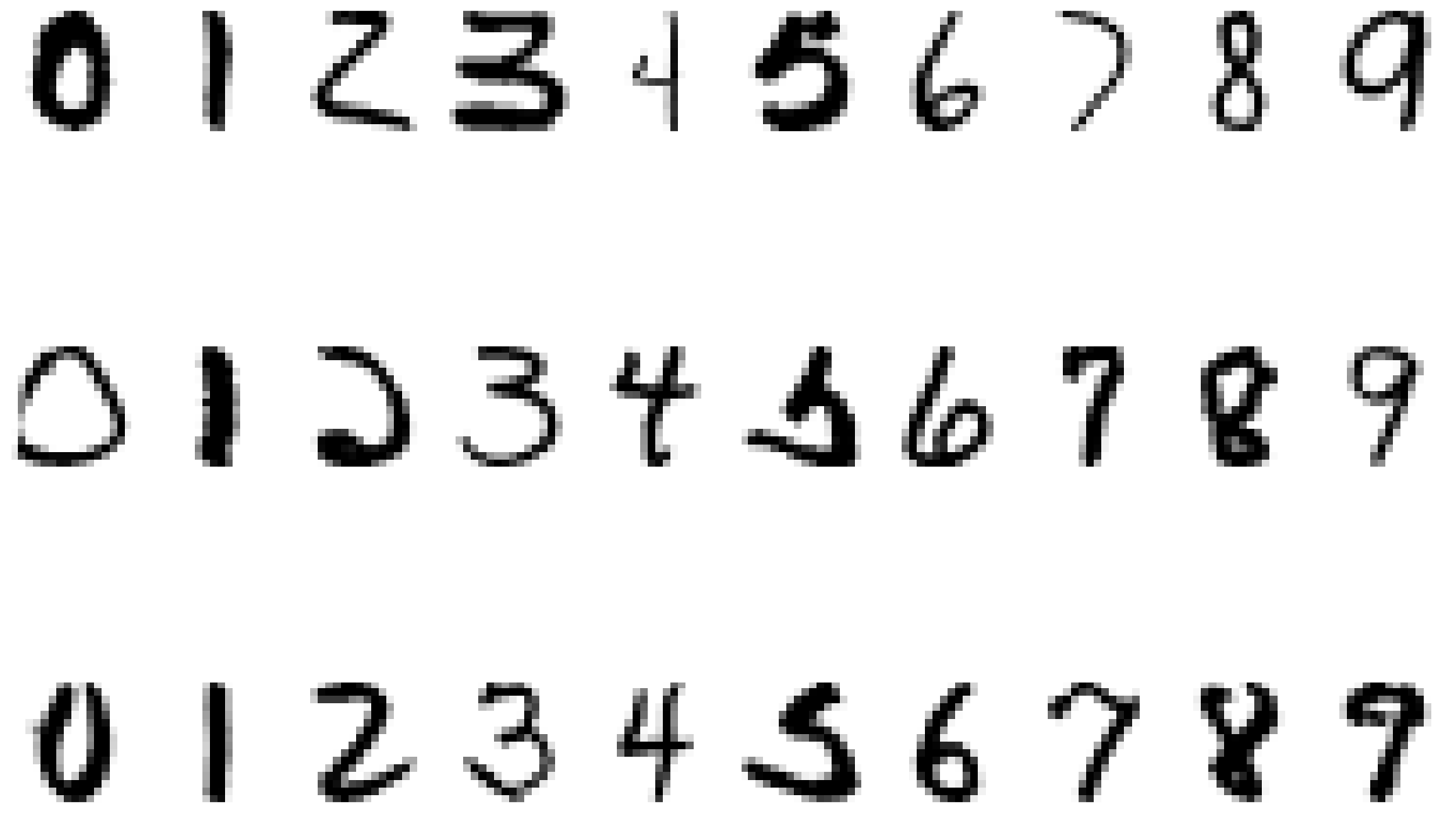}
\caption{USPS}
\end{subfigure}
\hfill
\begin{subfigure}[b]{0.24\textwidth}
\centering
\includegraphics[width=\textwidth]{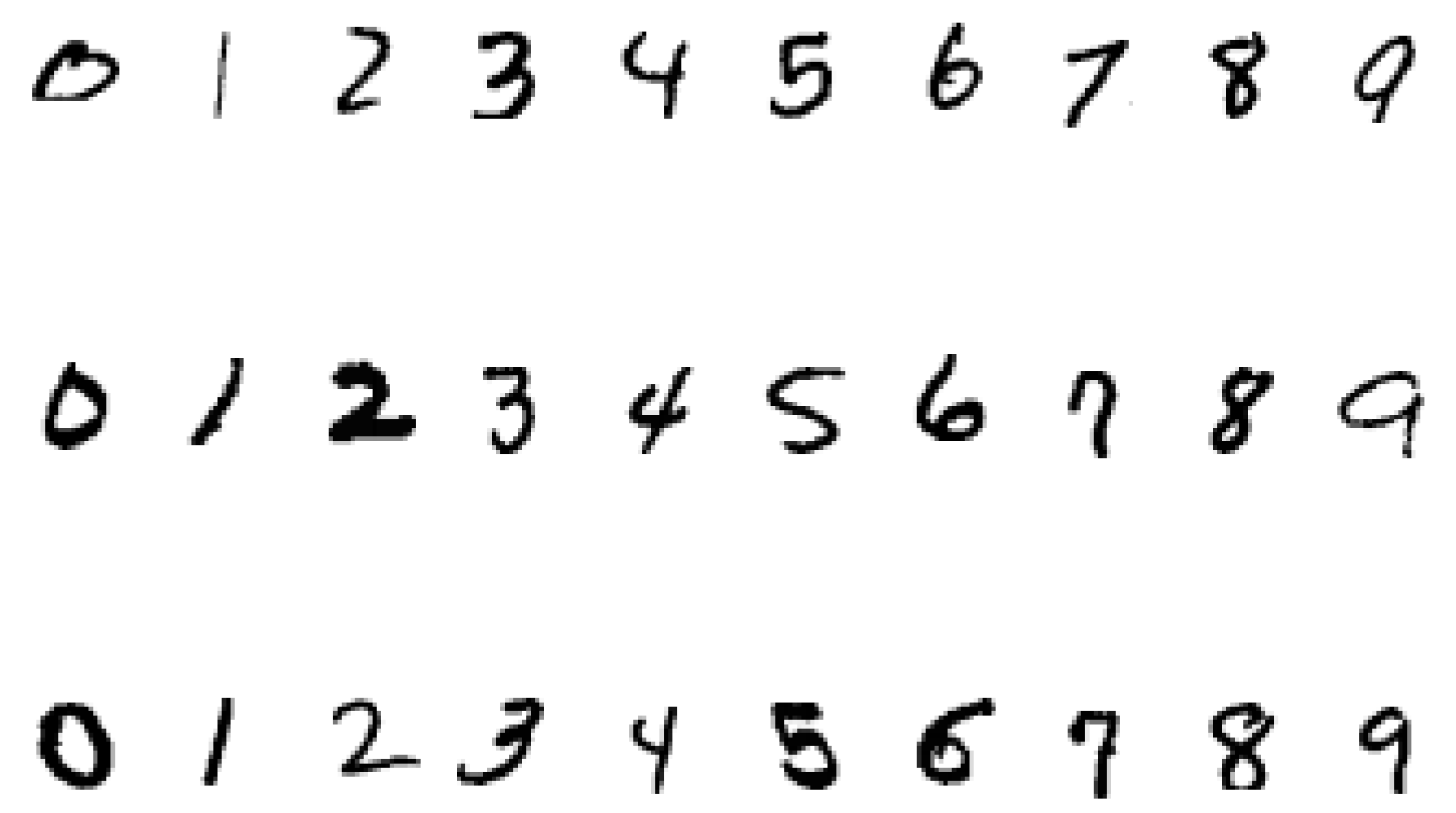}
\caption{MNIST}
\end{subfigure}
\hfill
\begin{subfigure}[b]{0.24\textwidth}
\centering
\includegraphics[width=\textwidth]{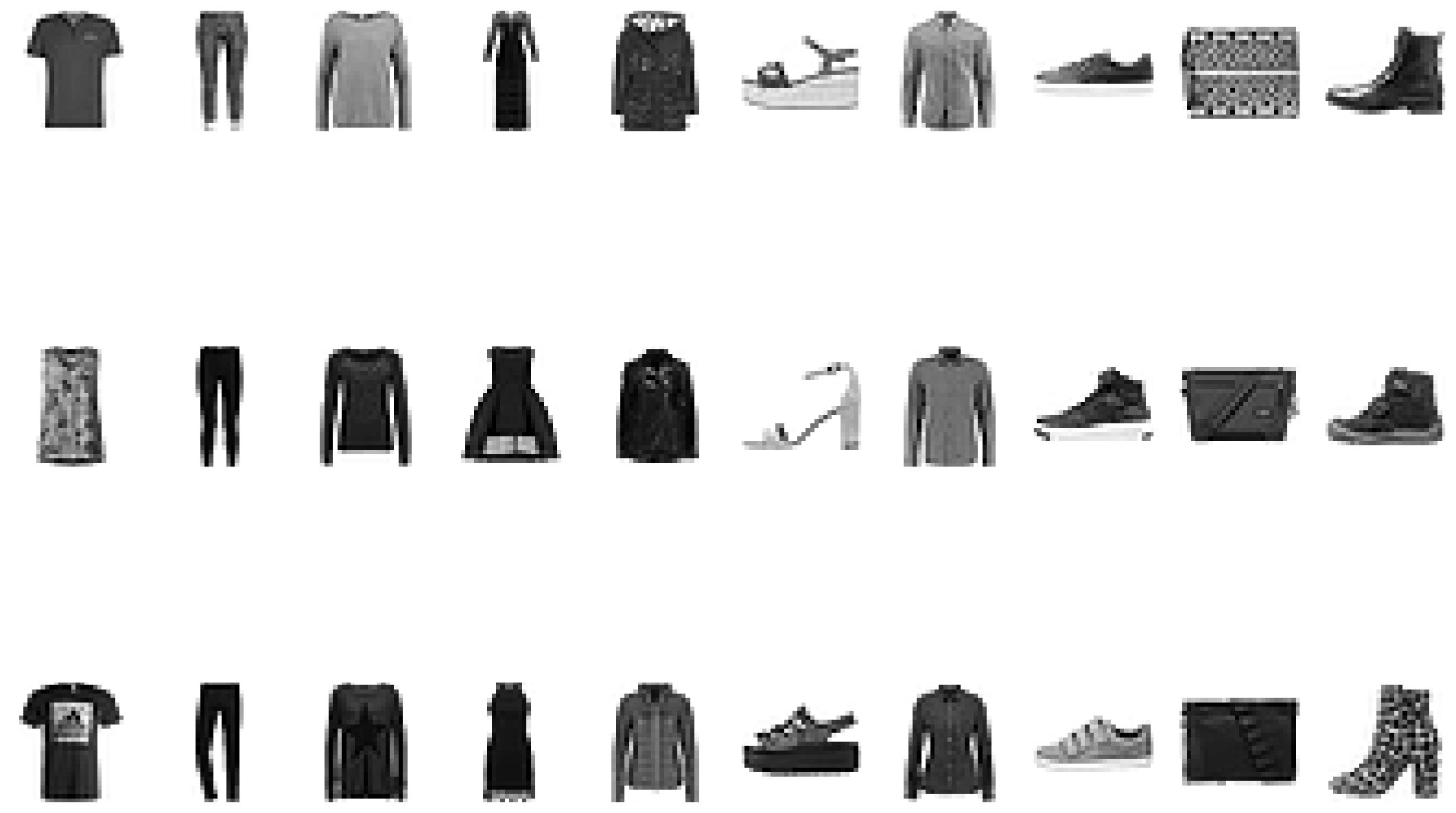}
\caption{Fashion MNIST}
\end{subfigure}
\caption{Real world data sets used in the experiments. Examples of samples from different classes are shown in the columns of (a)-(d).}
\label{fig:datasets_2}
\end{figure}

The experiments in this work were conducted with the benchmark data sets listed in Table~\ref{tab:datasets} and depicted in Figs.~\ref{fig:datasets_1} and~\ref{fig:datasets_2}, which comprise 16 synthetic (Gaussian-like) data sets with a varied number of clusters, dimensionalities and cluster covariances, as well as 4 real world image data sets. 

\begin{table}[!t]
\centering
\caption{Summary of the datasets.}
\begin{threeparttable}
\begin{tabular*}{\textwidth}{@{\extracolsep{\fill}}llllll@{}}
\toprule
dataset & type  & \#samples & \#dimensions & \#clusters & reference(s) \\
\midrule
\midrule
2d-4c-no0\tnote{a}				& synthetic 	& 1572  & 2     & 4     & \cite{Handl2005}		\\
2d-10c-no0\tnote{a}				& synthetic 	& 2972  & 2     & 10    & \cite{Handl2005}		\\
2d-20c-no0\tnote{a} 			& synthetic 	& 1517  & 2     & 20    & \cite{Handl2005}		\\
2d-40c-no0\tnote{a} 			& synthetic 	& 2563  & 2     & 40    & \cite{Handl2005} 		\\
10d-4c-no0\tnote{a} 			& synthetic 	& 1289  & 10    & 4     & \cite{Handl2005}		\\
10d-10c-no0\tnote{a} 			& synthetic 	& 2729  & 10    & 10    & \cite{Handl2005} 		\\
10d-20c-no0\tnote{a} 			& synthetic 	& 1013  & 10    & 20    & \cite{Handl2005} 		\\
10d-40c-no0\tnote{a} 			& synthetic 	& 1937  & 10    & 40    & \cite{Handl2005} 		\\
ellipsoid.50d4c.1\tnote{a} 		& synthetic 	& 1064  & 50    & 4     & \cite{Handl2005} 		\\
ellipsoid.50d10c.1\tnote{a} 	& synthetic 	& 2698  & 50    & 10    & \cite{Handl2005} 		\\
ellipsoid.50d20c.1\tnote{a} 	& synthetic 	& 1254  & 50    & 20    & \cite{Handl2005} 		\\
ellipsoid.50d40c.1\tnote{a} 	& synthetic 	& 2334  & 50    & 40    & \cite{Handl2005} 		\\
ellipsoid.100d4c.1\tnote{a} 	& synthetic 	& 1286  & 100   & 4     & \cite{Handl2005}		\\
ellipsoid.100d10c.1\tnote{a} 	& synthetic 	& 2892  & 100   & 10    & \cite{Handl2005} 		\\
ellipsoid.100d20c.1\tnote{a} 	& synthetic 	& 1338  & 100   & 20    & \cite{Handl2005} 		\\
ellipsoid.100d40c.1\tnote{a} 	& synthetic 	& 2211  & 100   & 40    & \cite{Handl2005} 		\\
Olivetti faces\tnote{b} 		& real world 	& 400  	& 4096  & 40    & \cite{Samaria.1994a} 	\\
USPS\tnote{c}  					& real world 	& 9298  & 256   & 10    & \cite{LeCun.1989a} 	\\
MNIST-test\tnote{d} 			& real world 	& 10000 & 784   & 10    & \cite{Lecun.1998a} 	\\
MNIST\tnote{d}					& real world 	& 70000 & 784   & 10    & \cite{Lecun.1998a} 	\\
Fashion MNIST\tnote{d}			& real world 	& 70000 & 784   & 10    & \cite{xiao2017} 		\\
\bottomrule
\end{tabular*}
\begin{tablenotes}[normal,flushleft]
\item[a]Cluster generators (sample data sets): \\ \url{personalpages.manchester.ac.uk/staff/Julia.Handl/generators.html}.
\item[b]Scikit-learn data sets (AT\&T Laboratories Cambridge): \\ \url{scikit-learn.org/stable/modules/classes.html#module-sklearn.datasets}.
\item[c]DynAE (data): \url{github.com/nairouz/DynAE}
\item[d]TensorFlow data sets: \url{tensorflow.org/datasets}.
\end{tablenotes}
\end{threeparttable}
\label{tab:datasets}
\end{table}

\subsection{Clustering algorithms} \label{sec:algs}

iCVI-ARTMAP was compared to the ART-based clustering methods of fuzzy ART (FA)~\cite{Carpenter1991} and dual vigilance fuzzy ART (DVFA)~\cite{leonardo.2018b}. Additionally, it was also compared to the following non-ART-based baselines: k-means~\cite{kmeans}, Gaussian mixture models (GMM)~\cite{xu2009}, spectral clustering (SC)~\cite{Luxburg.2007a} and hierarchical agglomerative clustering (HAC) methods (Ward, average, complete, single)~\cite{xu2009}. These non-ART-based methods were chosen for comparison purposes because all of them require the number of clusters ($k$) as an input parameter; moreover, they comprise a diverse and representative subset of traditional clustering algorithms.

\subsection{Quantitative evaluation} \label{sec:assess}

The performance of all clustering algorithms was measured using the adjusted rand index (ARI)~\cite{hubert1985}.

\subsection{Parameter setting} \label{sec:param_set}

For iCVI-ARTMAP and all non-ART-based clustering algorithms, the number of clusters parameter ($k$) was set to the ground truth value for the respective data set, and whenever possible, a grid search was performed to tune the remaining parameters. iCVI-ARTMAP has the same parameters as fuzzy ARTMAP: ARTa parameters (vigilance parameter $\rho_a$, learning rate $\beta_a$, choice parameter $\alpha_a$), map field parameters (vigilance parameter $\rho_{ab}$, learning rate $\beta_{ab}$, match tracking $\epsilon$) and the number of epochs for training ($E$). The additional parameters are the number of clusters ($k$), the iCVI and the tolerance parameter ($tol$). In particular, Table~\ref{tab:gridsearch} details the parameter search strategy employed for iCVI-ARTMAP, where the notation $[a,b] @ c$ corresponds to a grid search in the closed interval $[a, b]$ with a step size of $c$. Suitably setting the vigilance parameter is crucial in ART models~\cite{leonardo.2019b}. Thus, for simplicity, only the vigilance parameters of ARTa ($\rho_a$) and the map field ($\rho_{ab}$) were varied; all the remaining parameters were fixed. For k-means and GMM, the maximum number of epochs and the tolerance parameter for the convergence check were set to 300 and $10^{-6}$, respectively. For numerical stability, the term $\delta I$ was added into the computation of covariance matrices of the GMM, where $\delta = 10^{-\frac{12}{d}}$ and $d$ is the dimension of the data set. The same term is used in the incremental updates of covariance matrices of the iNI-ARTMAP variant as in~\cite{leonardo.2020a}. The SC method used a rbf kernel (for which a grid search using $[0.5, 1.5] @ 0.01$ was conducted), and kmeans with $10$ different initializations was used to perform the clustering task. Finally, for all $4$ HAC methods, the correct number of clusters $k$ was used to cut the dendrogram.

Regarding the ART-based clustering algorithms, the vigilance parameter ($\rho$) of FA and the upper bound vigilance parameter ($\rho_{ub}$) of DVFA were searched using $[0.1, 0.95] @ 0.01$, whereas for each $\rho_{ub}$ of DVFA, the corresponding lower bound vigilance parameter ($\rho_{lb}$) was searched using $[\rho_{ub} - 0.1, \rho_{ub}] @ 0.01$.

\begin{table}[!t]
\centering
\caption{iCVI-ARTMAP grid search details. ARTa's is parameterized by vigilance ($\rho_a$), learning rate ($\beta_a$), choice parameter ($\alpha_a$). Map field is parameterized by vigilance ($\rho_{ab}$), learning rate ($\beta_{ab}$), and match tracking parameter ($\epsilon$). iCVI-ARTMAP's addtional parameters are the number of clusters ($k$), number of epochs ($E$) and tolerance ($tol$) for convergence check of iCVI values. Except for $\rho_a$ and $\rho_{ab}$, all remaining parameters were fixed across all experiments.}
\begin{tabular}{m{11em}lrrrllll}
\toprule
\multirow{2}[4]{*}{dataset} & \multicolumn{8}{c}{iCVI-ARTMAP parameters} \\
\cmidrule{2-9}          
& $\rho_a$ & $\beta_a$ & $\alpha_a$ & $\epsilon$ & $\rho_{ab}$ & $\beta_{ab}$ & $tol$ & $E$ \\
\midrule
\midrule
synthetic data (2d) & [0.0, 0.95] @ 0.05 & 1.0   & 0.001 & 0.01  & [0.1, 1.0] @ 0.1 & 0.001  & 1e-6 & 20 \\
synthetic data (10d, 50d, 100d) & [0.0, 0.7] @ 0.05 & 1.0   & 0.001 & 0.01  & [0.1, 1.0] @ 0.1 & 0.001  & 1e-6 & 20 \\
Olivetti faces  & [0.0, 0.9] @ 0.05 & 1.0   & 0.001 & 0.01  & [0.1, 1.0] @ 0.1 & 0.001  & 1e-6 & 20 \\
USPS, MNIST-test  & [0.0, 0.9] @ 0.1 & 1.0   & 0.001 & 0.01  & [0.1, 1.0] @ 0.1 & 0.001  & 1e-6 & 20 \\
MNIST, Fashion MNIST & [0.0, 0.2] @ 0.1 & 1.0   & 0.001 & 0.01  & [0.1, 1.0] @ 0.1 & 0.001  & 1e-6 & 20 \\
\bottomrule
\end{tabular}
\label{tab:gridsearch}
\end{table}

\subsection{Experimental protocol} \label{sec:protocol}

Each data set was shuffled, and all algorithms were fed the samples in the same order (including those invulnerable to ordering effects). Because the clustering task is carried out in offline mode, the visual assessment of cluster tendency (VAT)~\cite{bezdek2002,Bezdek2017,Kumar.2020a} was used to sort each of the shuffled data sets prior to their presentation to FA and DVFA to improve their performance~\cite{leonardo2018,leonardo.2018b}. 

The baseline kmeans was initialized with the kmeans++ method~\cite{David.2007a}, and the best solution out of $10$ trials was selected. This clustering algorithm was also used to provide the same initial prototypes for GMM (means) and iCVI-ARTMAP (ARTa's categories - see Section~\ref{sec:training}), such that these two methods are initialized under conditions as similar as possible. 

Recent advances in deep learning have enabled the learning of representations that are more suitable for performing clustering. Therefore, the auto-encoder-based deep clustering method of 
Dynamic Autoencoder (DynAE)~\cite{Mrabah.2020a} was used as a pre-processor for the real world data sets of USPS, MNIST-test, MNIST and Fashion MNIST, so as to generate 10-dimensional latent spaces in which the clustering task was carried out. Eigenfaces~\cite{Kirby.1990a,Zhao.2003a}, a principal component analysis (PCA)-based classic face recognition method, has yielded reasonable classification accuracy for the Olivetti faces~\cite{Zhang.1997a}. Thus, PCA was used to project the latter data set to a reduced 20-dimensional space. 
 
\subsection{Implementation and reproducibility} \label{sec:code}

The experiments were carried using python. The iCVI-ARTMAP source code is provided at Guise's GitHub repository\footnote{available for academic purposes only at \url{https://github.com/GoGetter-Inc/iCVI-ARTMAP}}. Some of its components are based on the iCVI-toolbox for Matlab~\cite{leonardo.2020a} from the ACIL's GitHub repository\footnote{available at \url{https://github.com/ACIL-Group/iCVI-toolbox}}. The code for the non-ART-based baseline clustering algorithms and ARI are from scikit-learn\footnote{available at \url{https://scikit-learn.org}}~\cite{scikit-learn} whereas the code for VAT, FA and DVFA are from the ACIL's GitHub repository (\mbox{NuART-Py})\footnote{available at \url{https://github.com/ACIL-Group/NuART-Py}}~\cite{Elnabarawy.2019b}. The code for the DynAE is from\footnote{available at \url{https://github.com/nairouz/DynAE}}~\cite{Mrabah.2020a}.

\subsection{Results and discussion} \label{sec:results}

\begin{figure}[!p]
\centering
\subcaptionbox{iNI-ARTMAP}{\includegraphics[width=0.3\textwidth]{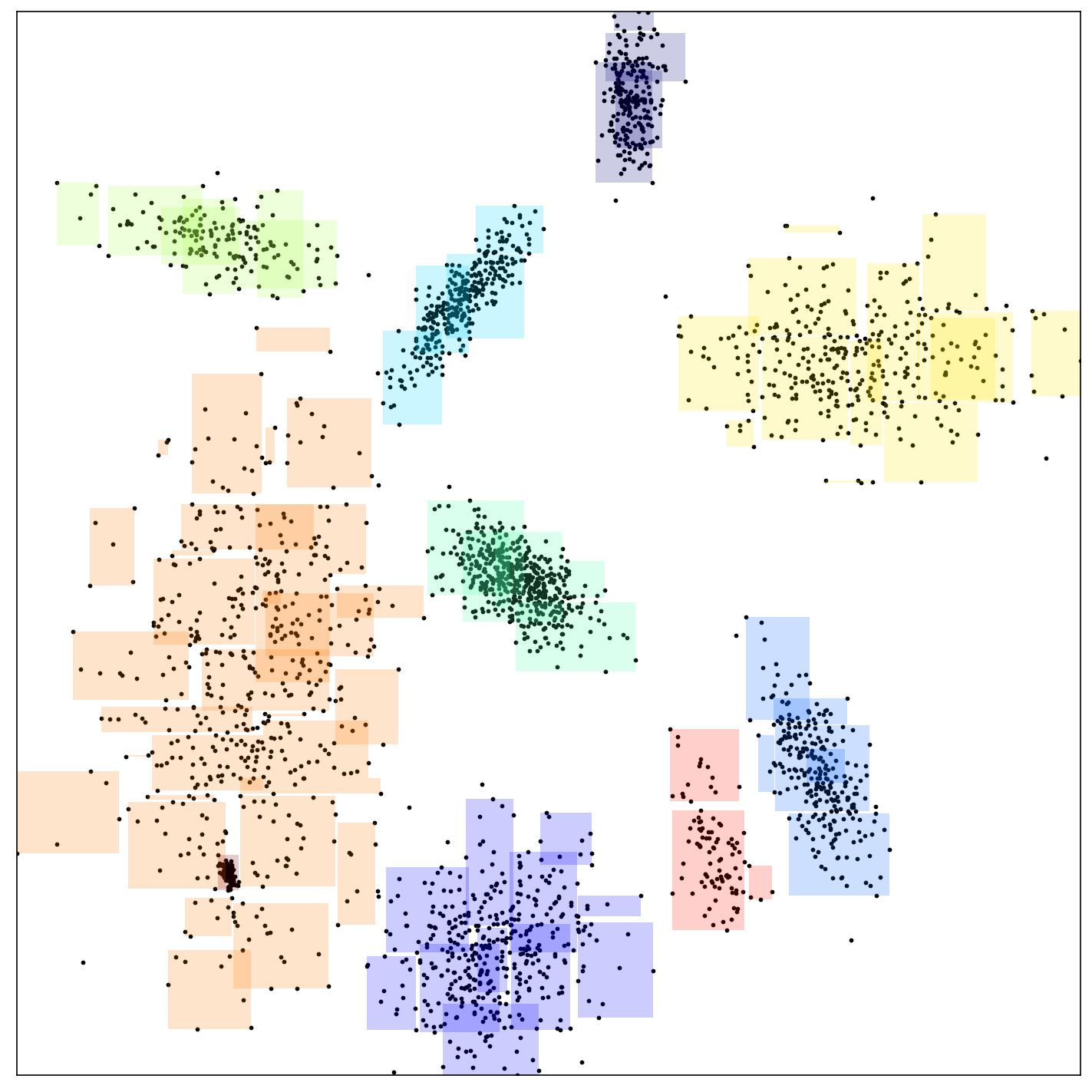}} 
\hfill
\subcaptionbox{iCH-ARTMAP}{\includegraphics[width=0.3\textwidth]{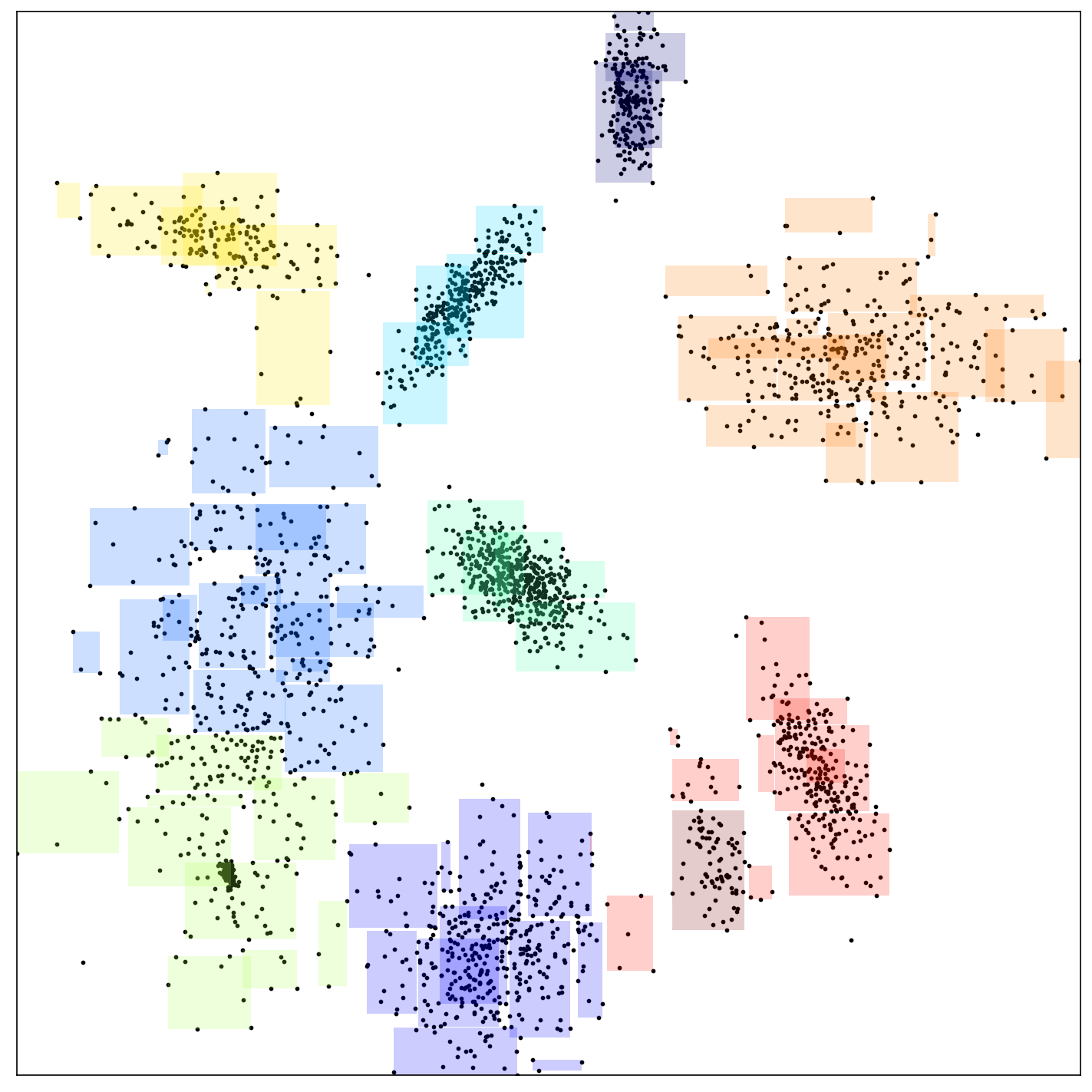}}
\hfill
\subcaptionbox{iWB-ARTMAP}{\includegraphics[width=0.3\textwidth]{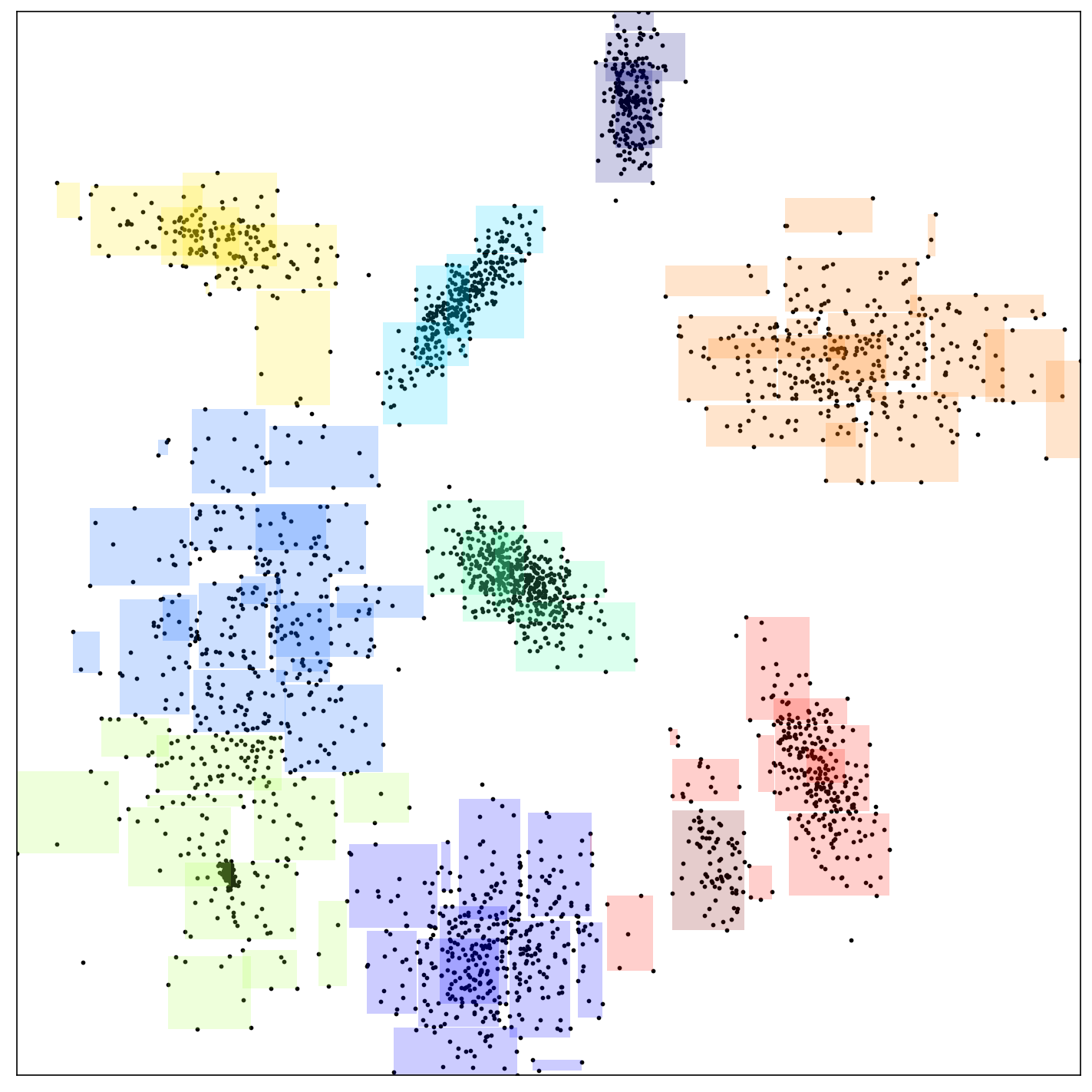}}

\subcaptionbox{iXB-ARTMAP}{\includegraphics[width=0.3\textwidth]{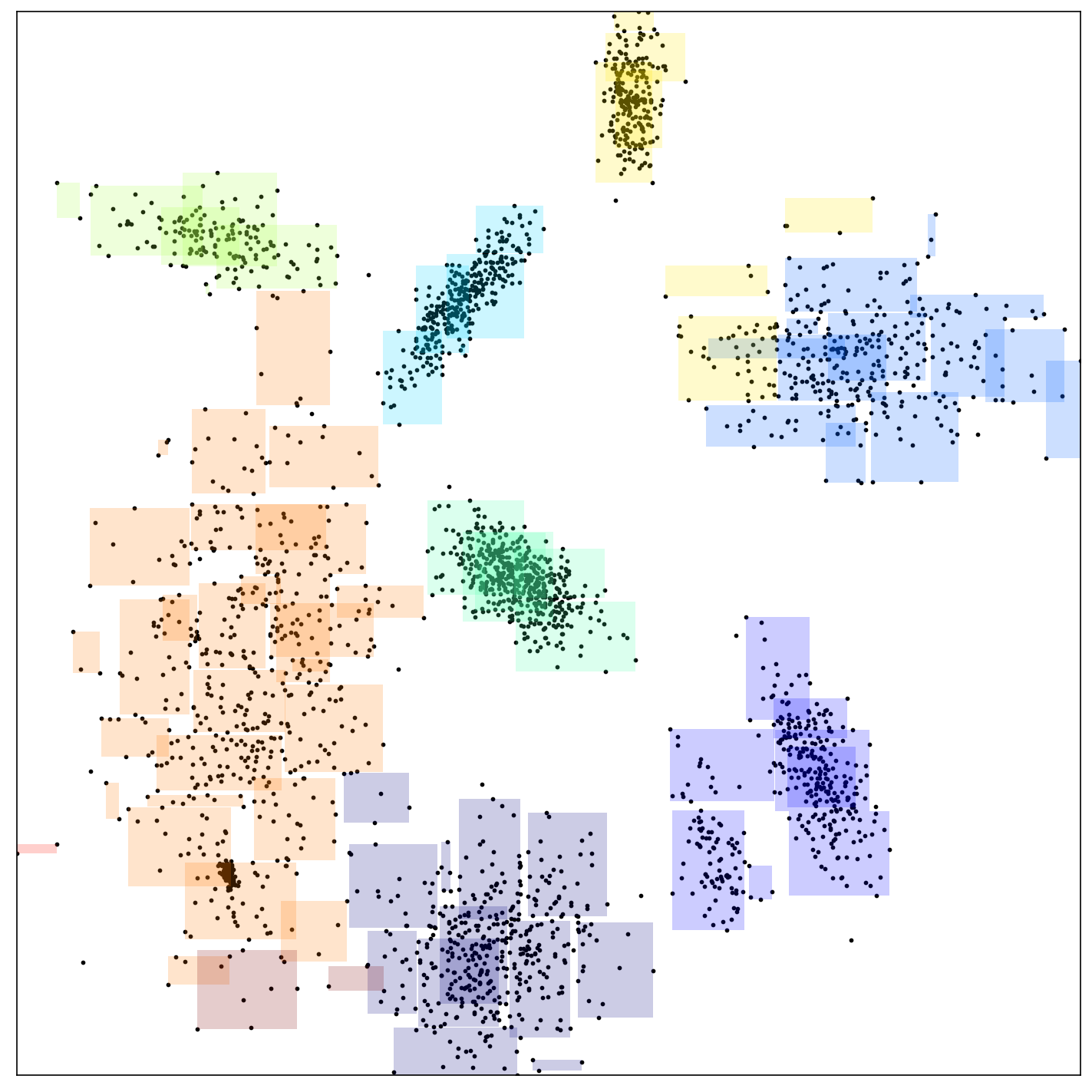}} 
\hfill
\subcaptionbox{iDB-ARTMAP}{\includegraphics[width=0.3\textwidth]{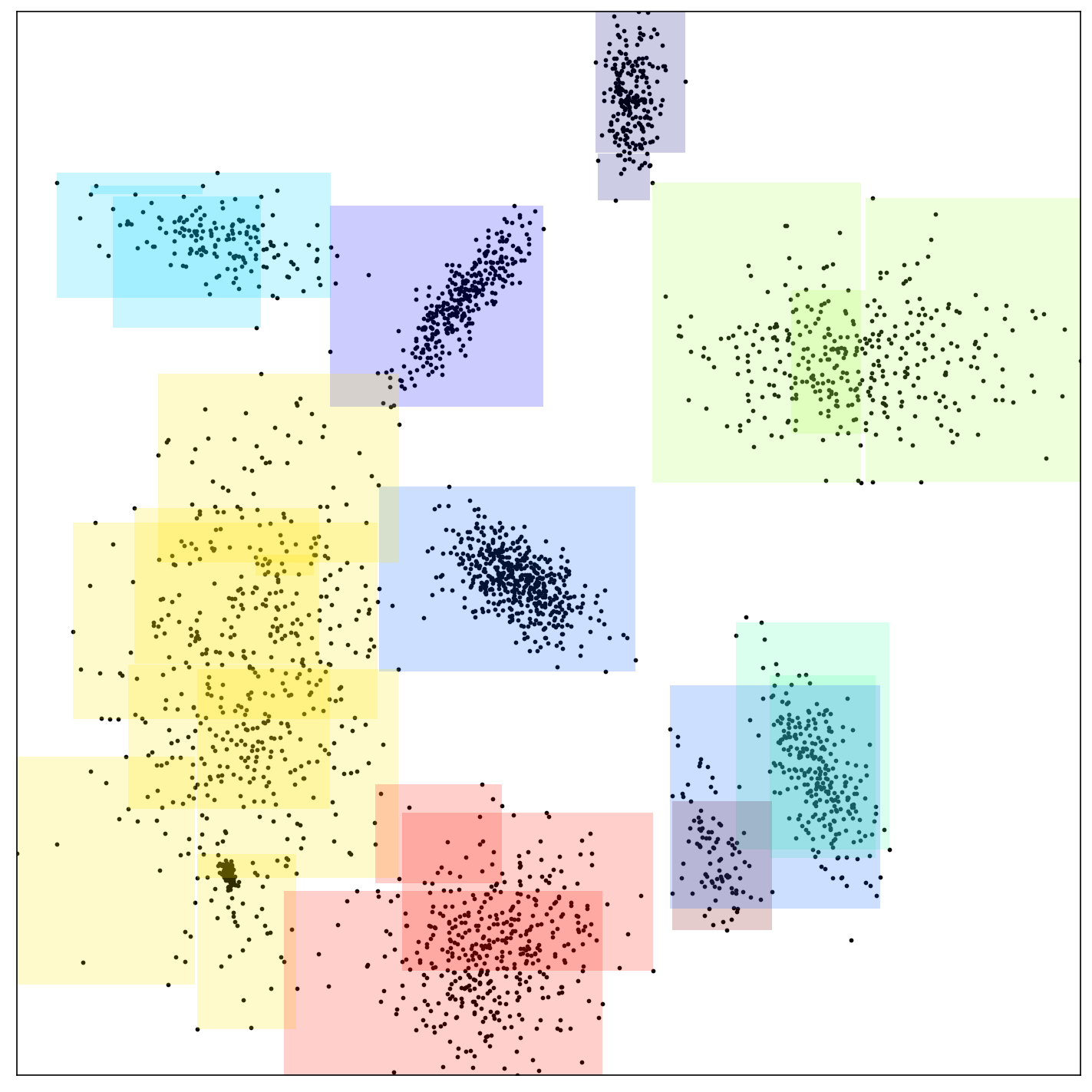}}
\hfill
\subcaptionbox{iPBM-ARTMAP}{\includegraphics[width=0.3\textwidth]{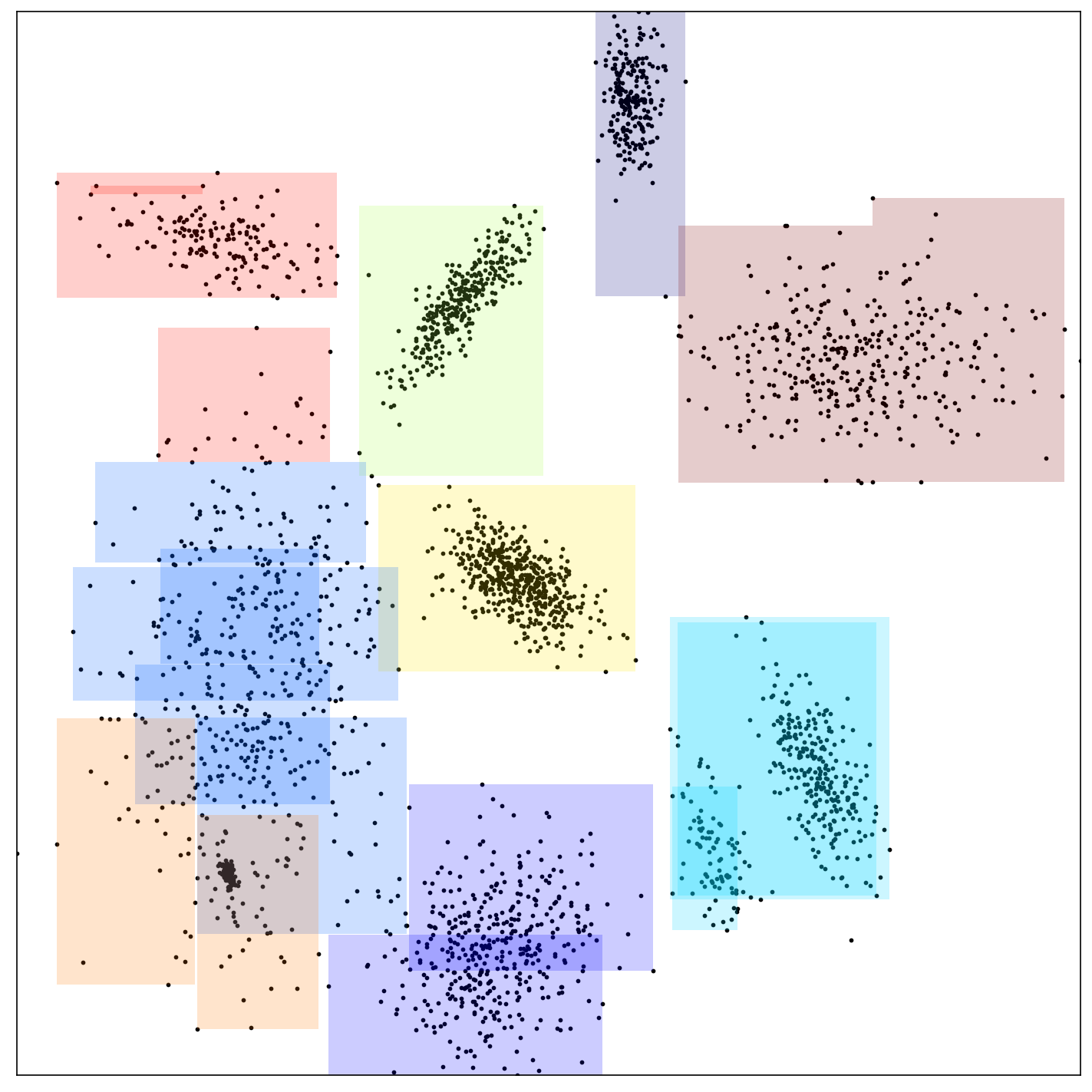}}
\caption{Best data partitions yielded by iCVI-ARTMAP variants for the 2d-10c-no0 data set reported on Table~\ref{tab:synthetic_results}.}
\label{Fig:partition_a}
\end{figure}

\begin{figure}[!p]
\centering
\subcaptionbox{iNI-ARTMAP}{\includegraphics[width=0.31\textwidth]{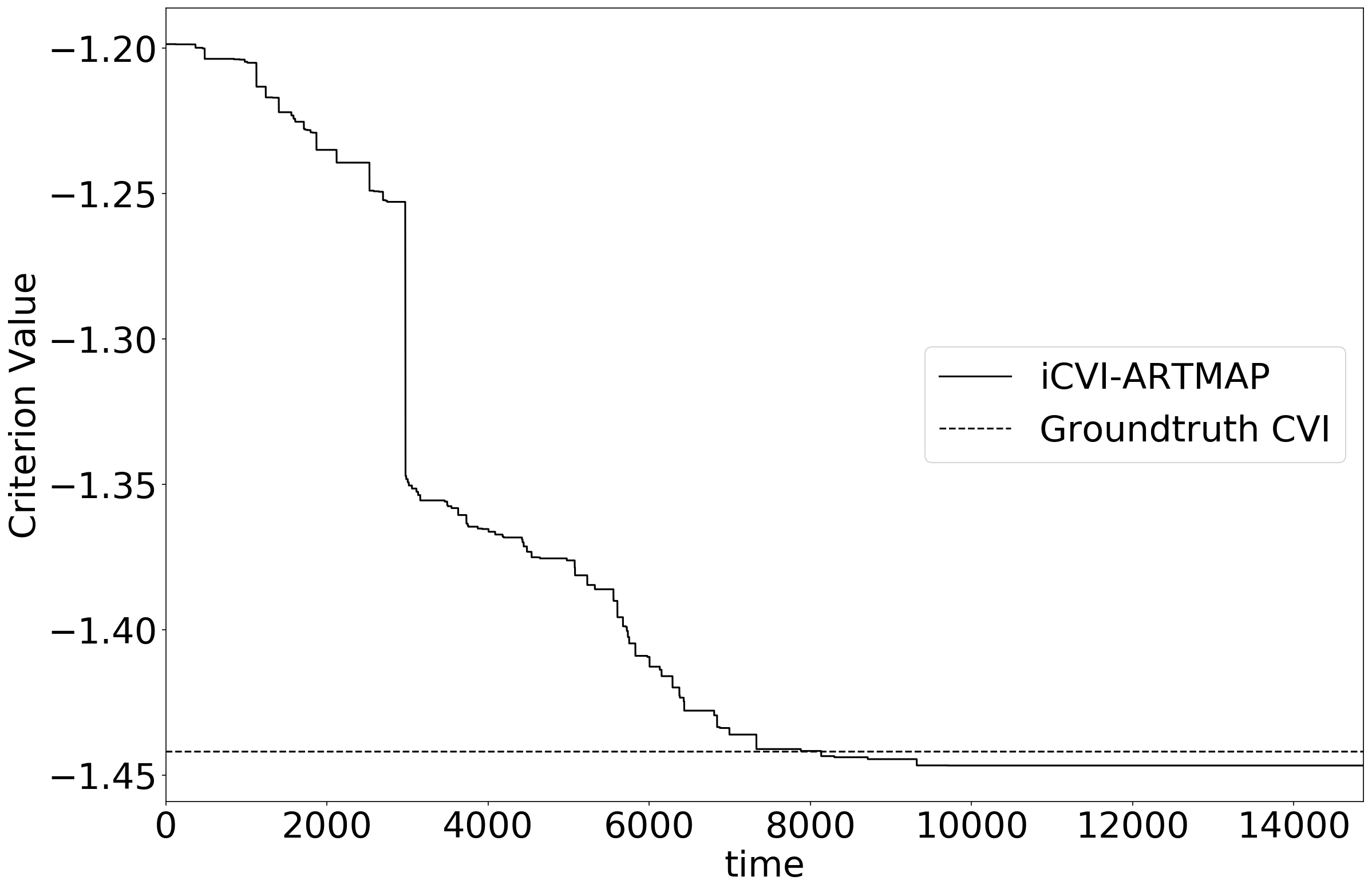}} 
\hfill
\subcaptionbox{iCH-ARTMAP}{\includegraphics[width=0.31\textwidth]{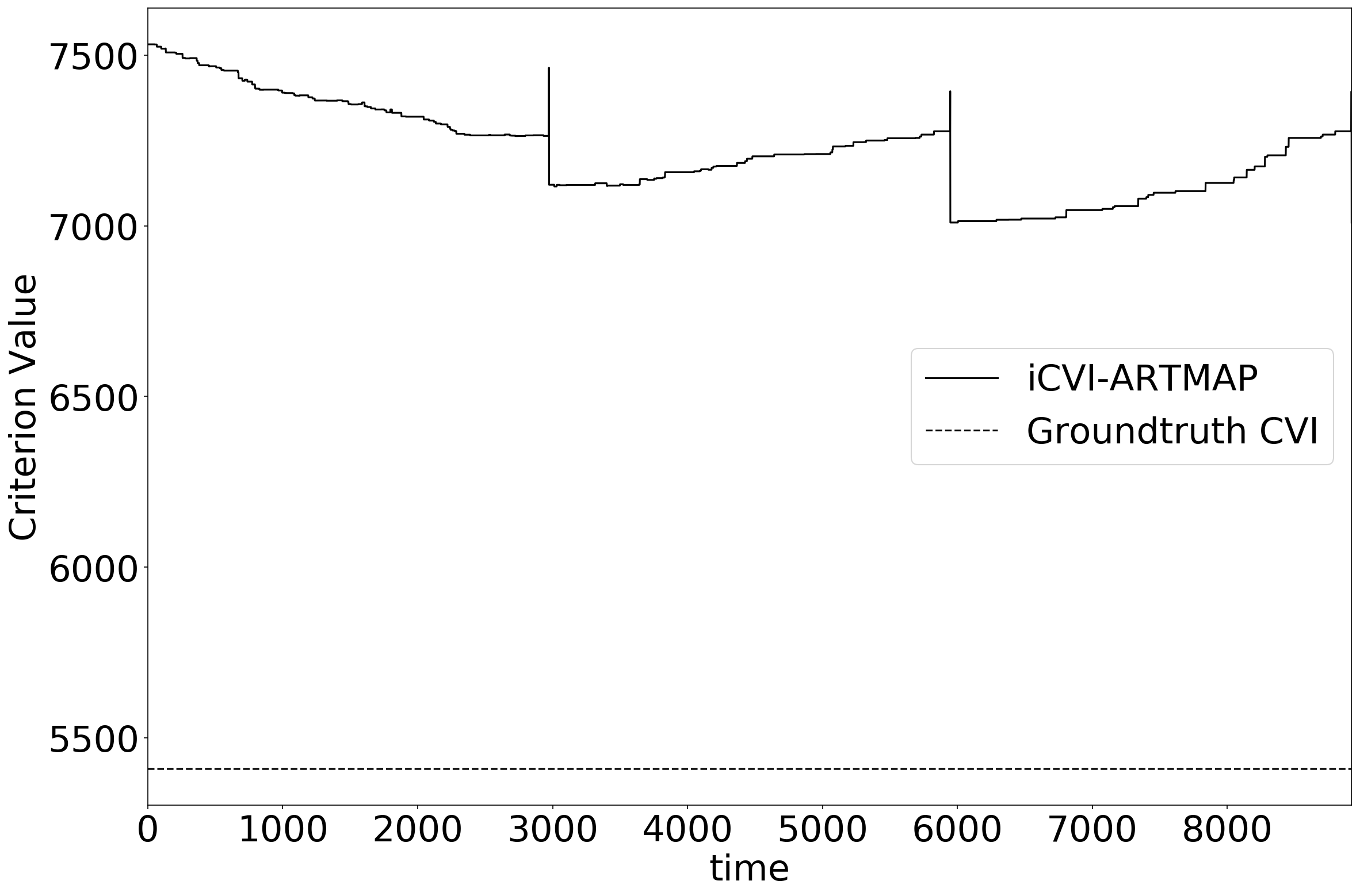}}
\hfill
\subcaptionbox{iWB-ARTMAP}{\includegraphics[width=0.31\textwidth]{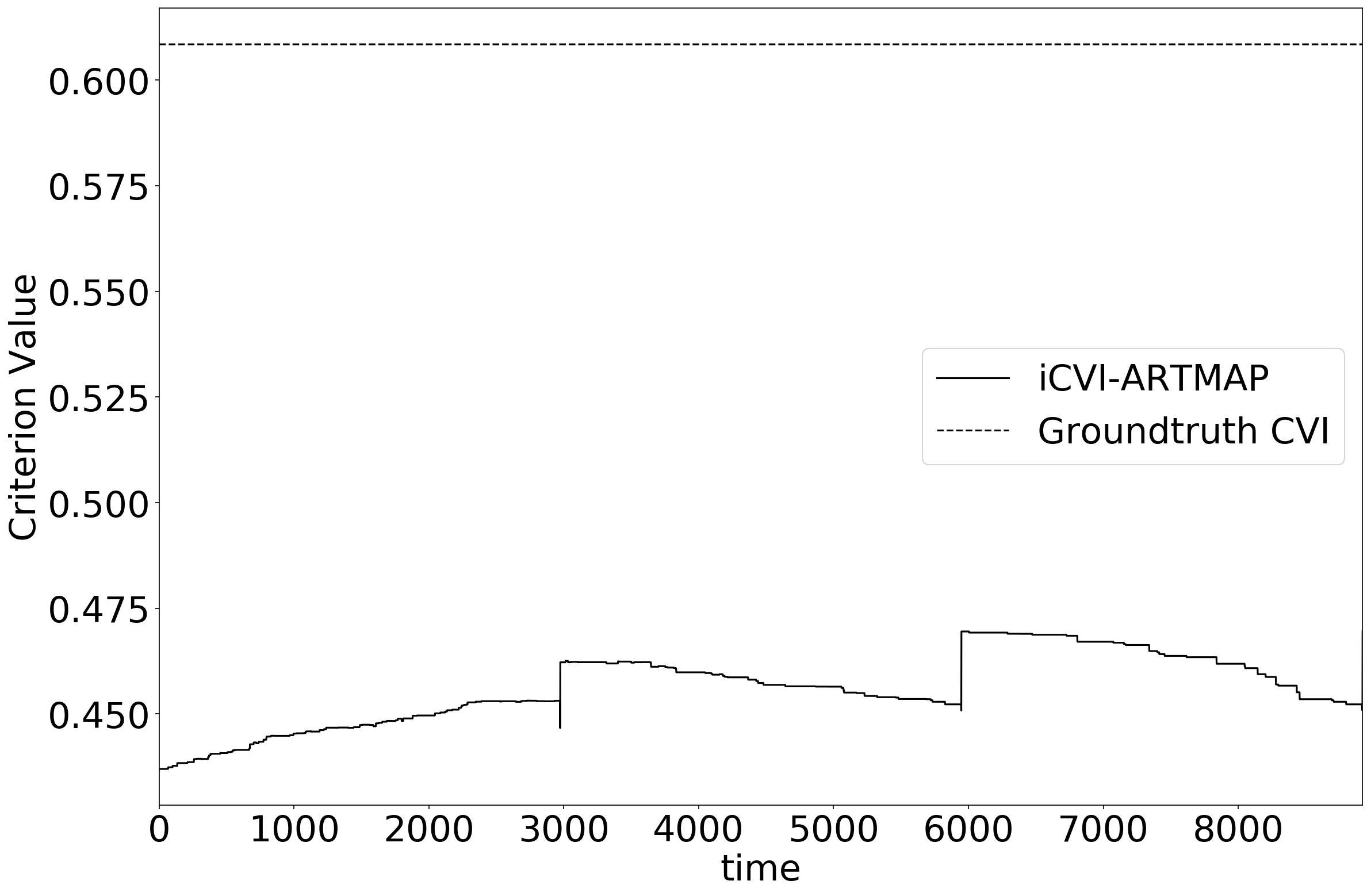}}

\subcaptionbox{iXB-ARTMAP}{\includegraphics[width=0.31\textwidth]{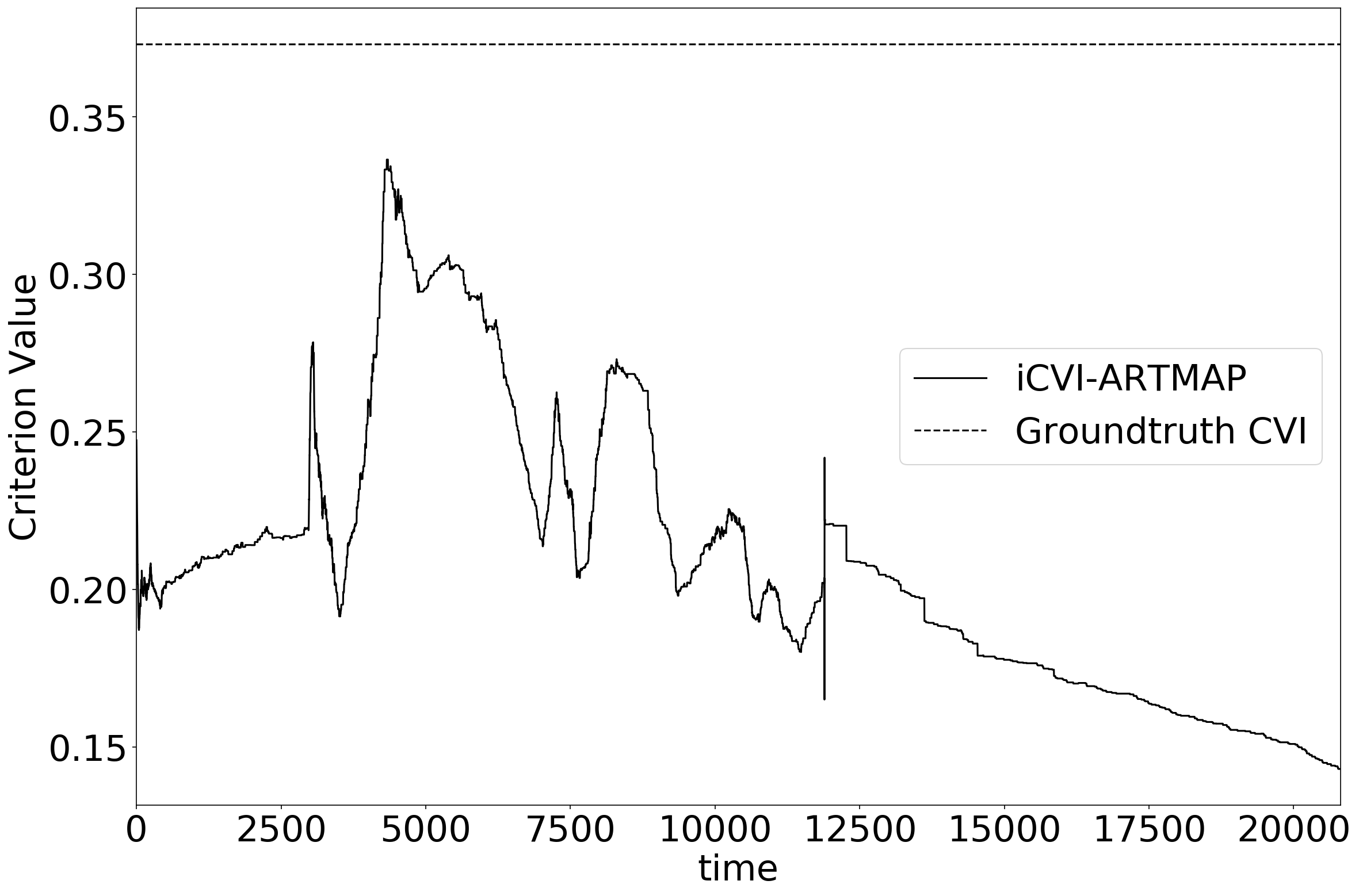}} 
\hfill
\subcaptionbox{iDB-ARTMAP}{\includegraphics[width=0.31\textwidth]{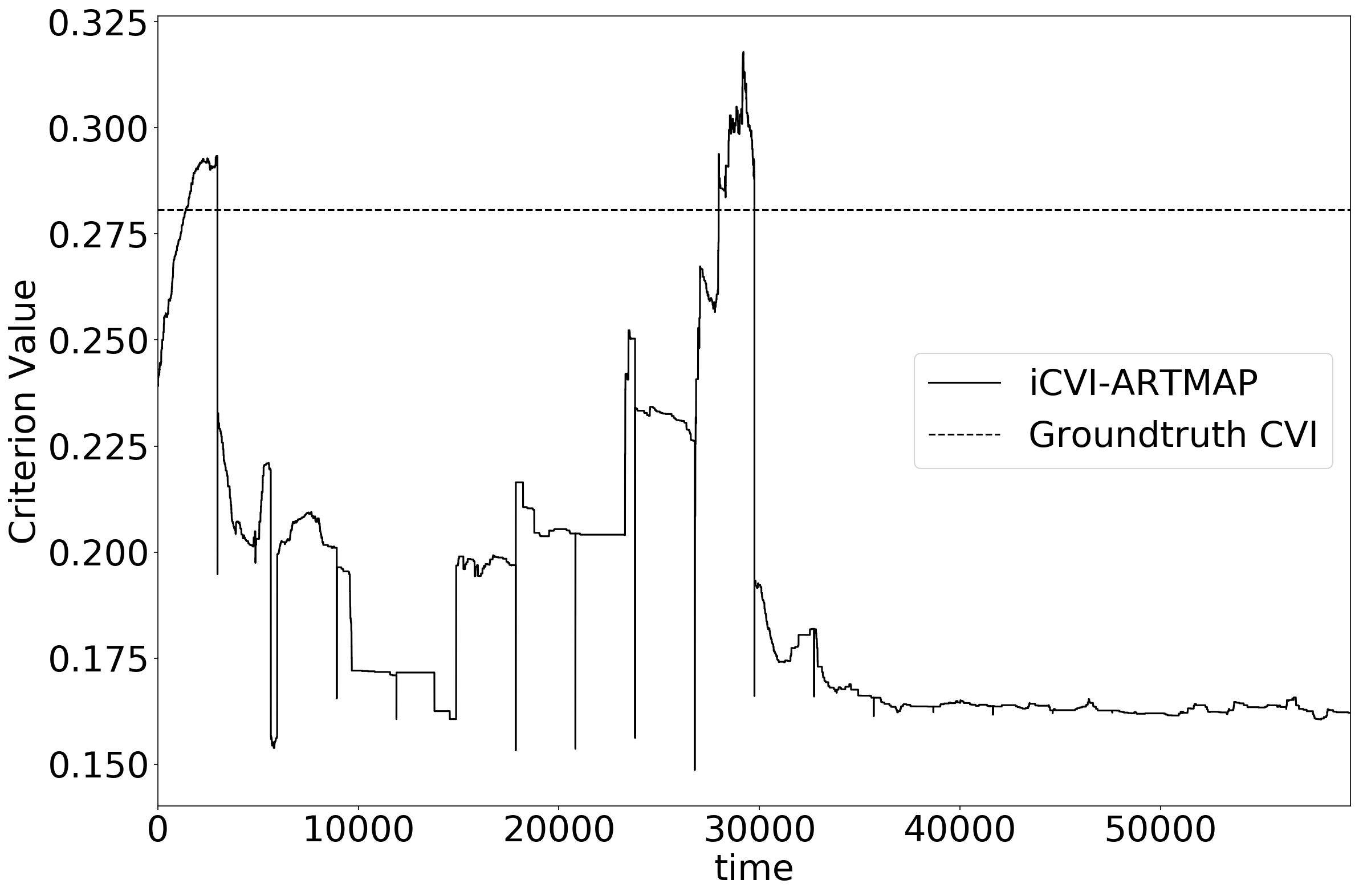}}
\hfill
\subcaptionbox{iPBM-ARTMAP}{\includegraphics[width=0.31\textwidth]{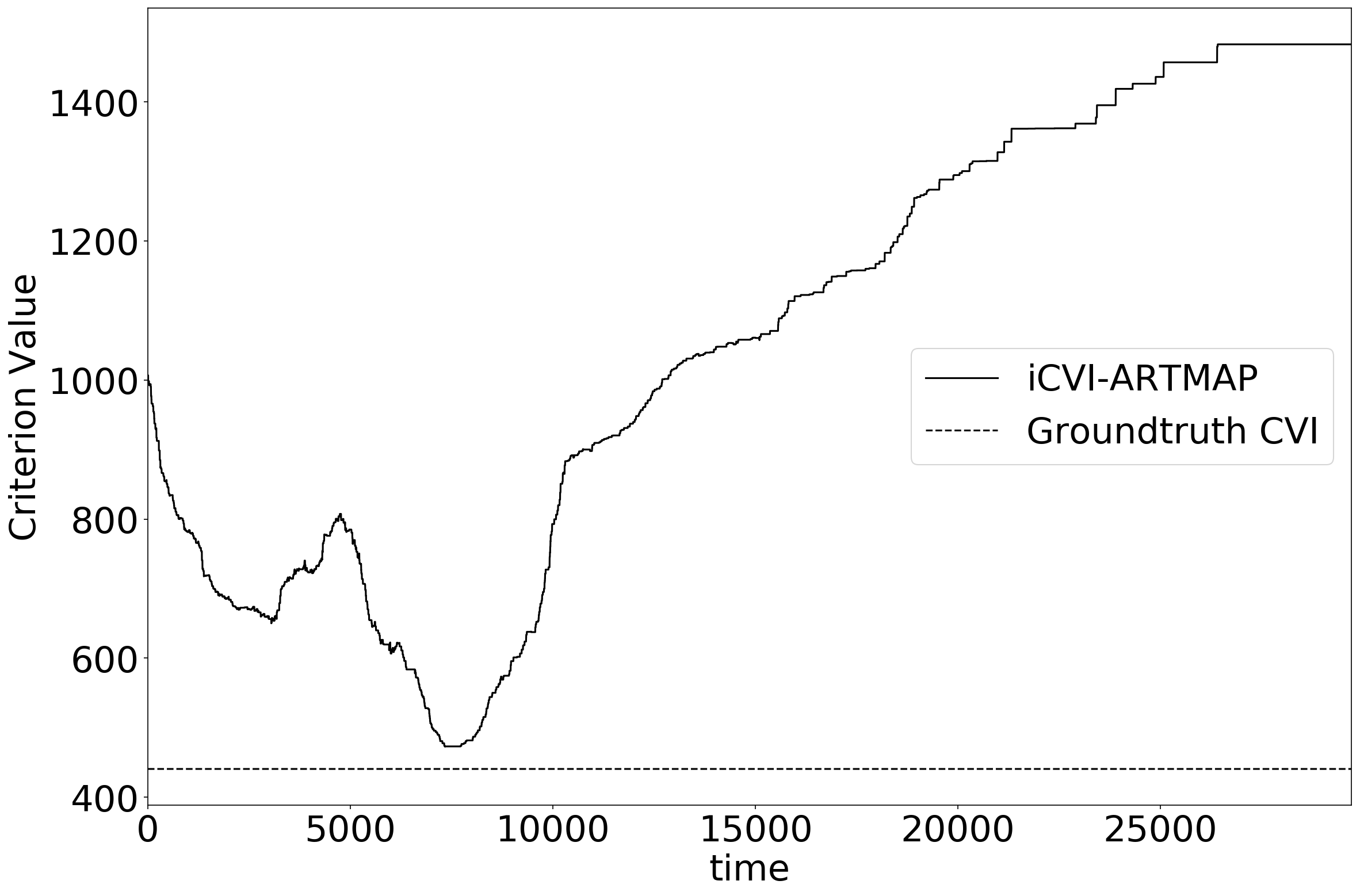}}
\caption{iCVI values versus iterations for the parameterization of iCVI-ARTMAP variants that partitioned the 2d-10c-no0 data set as shown in Fig.~\ref{Fig:partition_a} and whose performances are reported on Table~\ref{tab:synthetic_results}. Each time step corresponds to the presentation of one sample. The iCVI value of the ground truth partition is also shown for comparative purposes.}
\label{Fig:partition_b}
\end{figure}

Table~\ref{tab:synthetic_results} reports the best performance (in terms of ARI) of all clustering algorithms with respect to the synthetic benchmark data sets while following the parameter-setting strategy described in Section~\ref{sec:param_set}. In addition, Fig.~\ref{Fig:partition_a} illustrates the output partitions yielded by the iCVI-ARTMAP variants for the 2d-10c-no0 data set, as reported in Table~\ref{tab:synthetic_results}, while Fig.~\ref{Fig:partition_b} shows the values of their iCVIs varying during the training. A reference partition of a given data set may not map to the optimal value of a given CVI~\cite{Kim2016}, and Fig.~\ref{Fig:partition_b} shows that this is the case for the 2d-10c-no0 data set given the respective iCVI values for the ground truth partition (constant dashed lines). Table~\ref{tab:synthetic_results} shows that iNI-ARTMAP outperformed the other contenders in the vast majority of these data sets, thereby achieving the best average rank. GMM and SC obtained the second and third best average ranks, respectively. Notably, all the remaining iCVI-ARTMAP variants relying on sum-of-squares-based iCVIs (i.e., iCH, iWB, iXB, iDB and iPBM) achieved better average ranks than k-means (which is also based on compactness and was used for initialization purposes - see Section~\ref{sec:training}). In particular, iCH-ARTMAP, iWB-ARTMAP and iXB-ARTMAP achieved the best average ranks among the latter variants, while also surpassing Ward's HAC. Moreover, except for iDB-ARTMAP, all iCVI-ARTMAP variants achieved better average ranks than the baseline ART-based clustering algorithms combined with VAT. Notably, the iXB-ARTMAP was the most robust (performance-wise) sum-of-squares-based iCVI-ARTMAP variant in clustering the higher-dimensional data sets. As a trade-off for better accuracy, iCVI-ARTMAP's execution time is longer than that of the contenders; it computes the iCVI values for assigning every single sample presented to each existing cluster across epochs. However, as shown in Section~\ref{sec:speed}, the iCVI formulation is orders of magnitude faster than using batch CVIs for improving cluster selection. 

\begin{table}[!b]
\centering
\caption{Performance of clustering algorithms on benchmark synthetic data sets; best results in terms of ARI are reported in bold.}
\resizebox{\textwidth}{!}{
\begin{tabular}{lrrrrrrrrrrrrrrr}
\toprule
\multirow{2}[4]{*}{\textbf{data set}} & 
\multicolumn{6}{c}{iCVI-ARTMAP} & 
\multirow{2}[4]{*}{\textbf{VAT+FA}} & 
\multirow{2}[4]{*}{\textbf{VAT+DVFA}} & 
\multirow{2}[4]{*}{\textbf{kmeans}} & 
\multirow{2}[4]{*}{\textbf{GMM}} & 
\multirow{2}[4]{*}{\textbf{SC}} & 
\multicolumn{4}{c}{Hierarchical Agglomerative Clustering} \\
\cmidrule{2-7}
\cmidrule{13-16} & 
\textbf{iNI} & 
\textbf{iCH} & 
\textbf{iWB} & 
\textbf{iXB} & 
\textbf{iDB} & 
\textbf{iPBM} &       &       &       &       &       & 
\textbf{Ward} & 
\textbf{complete} & 
\textbf{average} & 
\textbf{single} \\
\midrule
2d-4c-no0 & \textbf{0.9941} & 0.9843 & 0.9843 & 0.9843 & 0.9806 & 0.9843 & 0.8478 & 0.8478 & 0.8817 & \textbf{0.9941} & 0.8899 & 0.9843 & 0.6446 & 0.7696 & 0.6172 \\
2d-10c-no0 & \textbf{0.9941} & 0.8880 & 0.8880 & 0.8673 & 0.9038 & 0.8876 & 0.8044 & 0.8044 & 0.8288 & 0.9280 & 0.8400 & 0.8447 & 0.8301 & 0.8591 & 0.2318 \\
2d-20c-no0 & \textbf{0.9990} & 0.9843 & 0.9843 & 0.9801 & 0.9804 & 0.9843 & 0.9209 & 0.9209 & 0.9770 & 0.9918 & 0.8927 & 0.9905 & 0.7955 & 0.8485 & 0.6560 \\
2d-40c-no0 & \textbf{0.9863} & 0.8898 & 0.8898 & 0.8733 & 0.8658 & 0.8999 & 0.7986 & 0.7986 & 0.8820 & 0.9536 & 0.8285 & 0.8994 & 0.7746 & 0.8528 & 0.5826 \\
10d-4c-no0 & \textbf{1.0000} & \textbf{1.0000} & \textbf{1.0000} & 0.9989 & 0.9994 & 0.9679 & 0.9530 & 0.9638 & 0.9860 & \textbf{1.0000} & \textbf{1.0000} & \textbf{1.0000} & 0.4218 & 0.9707 & 0.0024 \\
10d-10c-no0 & \textbf{0.9981} & 0.9624 & 0.9601 & 0.9012 & 0.8652 & 0.9196 & 0.6392 & 0.6392 & 0.8681 & 0.8983 & 0.9848 & 0.9901 & 0.6622 & 0.6737 & 0.0008 \\
10d-20c-no0 & 0.9981 & 0.9963 & 0.9963 & 0.9962 & 0.9962 & 0.9963 & 0.9374 & 0.9374 & 0.9963 & 0.9981 & 0.9981 & \textbf{1.0000} & \textbf{1.0000} & \textbf{1.0000} & 0.8860 \\
10d-40c-no0 & 0.9962 & 0.9937 & 0.9917 & 0.9546 & 0.9694 & 0.9915 & 0.9436 & 0.9436 & 0.9793 & 0.9799 & 0.9981 & 0.9990 & 0.9981 & \textbf{1.0000} & 0.8465 \\
ellipsoid.50d4c.1 & \textbf{1.0000} & 0.6605 & 0.6605 & 0.6130 & 0.5636 & 0.5738 & 0.6621 & 0.6621 & 0.4539 & 0.6755 & \textbf{1.0000} & 0.6027 & 0.4151 & 0.5610 & 0.0006 \\
ellipsoid.50d10c.1 & \textbf{0.9995} & 0.4624 & 0.5542 & 0.6760 & 0.8399 & 0.4036 & 0.5363 & 0.5363 & 0.3995 & 0.8421 & 0.9739 & 0.4283 & 0.2824 & 0.3200 & 0.0004 \\
ellipsoid.50d20c.1 & \textbf{1.0000} & 0.4651 & 0.4651 & 0.6119 & 0.3827 & 0.4374 & 0.6935 & 0.6935 & 0.3365 & 0.9166 & 0.7288 & 0.3835 & 0.3347 & 0.1772 & 0.0004 \\
ellipsoid.50d40c.1 & \textbf{0.9645} & 0.3724 & 0.3724 & 0.6327 & 0.1198 & 0.2807 & 0.6948 & 0.6948 & 0.2515 & 0.9049 & 0.7576 & 0.3003 & 0.1866 & 0.1670 & 0.0004 \\
ellipsoid.100d4c.1 & \textbf{1.0000} & 0.4441 & 0.5295 & 0.8626 & 0.9540 & 0.3083 & 0.7055 & 0.7055 & 0.3942 & 0.5750 & \textbf{1.0000} & 0.2905 & 0.3868 & 0.1869 & -0.0010 \\
ellipsoid.100d10c.1 & \textbf{1.0000} & 0.6061 & 0.6061 & 0.8037 & 0.5537 & 0.5307 & 0.7397 & 0.7397 & 0.5001 & 0.9554 & \textbf{1.0000} & 0.4026 & 0.1973 & 0.2353 & -0.0006 \\
ellipsoid.100d20c.1 & \textbf{0.9572} & 0.5772 & 0.5772 & 0.7231 & 0.1747 & 0.3629 & 0.6931 & 0.6931 & 0.3545 & 0.9211 & 0.8894 & 0.5504 & 0.1598 & 0.1801 & 0.0018 \\
ellipsoid.100d40c.1 & \textbf{0.9750} & 0.3974 & 0.3974 & 0.6947 & 0.1161 & 0.3945 & 0.6525 & 0.6525 & 0.3989 & 0.9471 & 0.5261 & 0.2870 & 0.1792 & 0.1345 & 0.0013 \\
\midrule
\textbf{Average rank} & 1.78  & 6.56  & 6.44  & 6.66  & 9.09  & 8.41  & 8.97  & 8.91  & 10.09 & 3.56  & 4.88  & 7.09  & 11.81 & 10.75 & 15.00 \\
\bottomrule
\end{tabular}
}
\label{tab:synthetic_results}
\end{table}

Given the performances of the iNI- and iCH-ARTMAPs for the synthetic data sets, they are the only iCVI-ARTMAP variants chosen for the experiments with the real world image data sets. Table~\ref{tab:real_world_results} reports the experimental results consisting of the best performance in terms of ARI. Empty cells indicate that the corresponding method was not run due to execution time and/or memory constraints. In addition, these cells are disregarded for the rank and average rank computations (row-wise and column-wise, respectively). As discussed in Section~\ref{sec:protocol}, the clustering task was carried out in the latent spaces of trained DynAEs or PCA projection. Note that the goal is not to assess the individual deep clustering methods on their ability to learn representations or to evaluate the merits of dimensionality reduction techniques; rather, it is to evaluate and compare the clustering algorithms. Table~\ref{tab:real_world_results} shows that iNI-ARTMAP continues to yield the best performance among the clustering methods for most of the real world data sets under experimentation. The performance difference is more expressive for Olivetti faces and Fashion MNIST, whereas for the other data sets, the performance is very similar; iNI-ARTMAP outperforms the close contenders by a very small margin for the USPS and MNIST-test data sets, and GMM does similarly for the MNIST data set. The iCH-ARTMAP outperformed k-means in the experiments with Olivetti faces and USPS (for a small margin), while performing similarly for the remaining data sets. Both iCVI-ARTMAP variants outperformed the other ART-based methods in all data sets for which comparison was possible, with the exception of the Olivetti faces data set, for which iCH-ARTMAP was outperformed.

\begin{table}[!t]
\centering
\caption{Performance of clustering algorithms on benchmark real world data sets; best results in terms of ARI are reported in bold. The notation ``($dim-D$)'' shown in parentheses under the ``pre-processing'' column indicates that the data set was transformed to a $dim$-dimensional space using the corresponding technique.}
\resizebox{\textwidth}{!}{
\begin{tabular}{lrrrrrrrrrrrr}
\toprule
\multirow{2}[4]{*}{\textbf{data set}} & 
\multirow{2}[4]{*}{\textbf{Pre-processing}} & 
\multicolumn{2}{c}{iCVI-ARTMAP} & 
\multirow{2}[4]{*}{\textbf{VAT+FA}} & 
\multirow{2}[4]{*}{\textbf{VAT+DVFA}} & 
\multirow{2}[4]{*}{\textbf{kmeans}} & 
\multirow{2}[4]{*}{\textbf{GMM}} & 
\multirow{2}[4]{*}{\textbf{SC}} & 
\multicolumn{4}{c}{Hierarchical Agglomerative Clustering} \\
\cmidrule{3-4}
\cmidrule{10-13} &       & 
\textbf{iNI} & 
\textbf{iCH} &       &       &       &       &       & 
\textbf{Ward} & 
\textbf{complete} & 
\textbf{average} & 
\textbf{single} \\
\midrule
Olivetti Faces & 
PCA (20-D) & \textbf{0.6374} & 0.5863 & 0.5939 & 0.5939 & 0.5578 & 0.5866 & 0.2327 & 0.6341 & 0.4369 & 0.3362 & 0.0503 \\
USPS  & 
DynAE (10-D) & \textbf{0.9649} & 0.9609 & 0.9260 & 0.9260 & 0.9599 & 0.9623 & 0.9614 & 0.9593 & 0.9473 & 0.8906 & 0.0000 \\
MNIST-test & 
DynAE (10-D) & \textbf{0.9643} & 0.9627 & 0.8429 & 0.8429 & 0.9627 & 0.9640 & 0.9628 & 0.9516 & 0.8698 & 0.8690 & 0.0000 \\
MNIST & 
DynAE (10-D) & 0.9698 & 0.9688 & - & - & 0.9688 & \textbf{0.9699} & - & - & - & - & - \\
Fashion MNIST & 
DynAE (10-D) & \textbf{0.4802} & 0.4495 & - & - & 0.4495 & 0.4704 & - & - & - & - & - \\
\midrule
\textbf{Average rank} &       & 1.20  & 4.30  & 7.17  & 7.17  & 4.70  & 2.40  & 5.33  & 4.67  & 7.33  & 9.00  & 11.00 \\
\bottomrule
\end{tabular}  
}
\label{tab:real_world_results}
\end{table}

\subsection{Speed advantages of iCVI-ARTMAP} \label{sec:speed}

The appeal of using iCVIs is the prospect of considerably decreasing the computational burden when performing this type of offline incremental clustering. To showcase this advantage, the elapsed times when imbuing ARTMAP with batch versus incremental CVIs (the latter also featuring caching variables) was investigated. These models are referred to as bCVI-ARTMAP and iCVI-ARTMAP, respectively. To this end, the data set ellipsoid.100d40c.1 was clustered by varying the number of clusters ($k$) in the closed interval $[2, 40]$ with a step size of $1$ and then recording the execution times. Specifically, most of the (b/i)CVI-ARTMAP parameters were set as reported in Table~\ref{tab:gridsearch}, with the exception of $\rho_a$, $\rho_{ab}$ and $E$, which were set to $0.7$, $1.0$ and $1$ (single epoch), respectively. 

\begin{figure}[!t]
\centering
\begin{subfigure}[b]{0.3\textwidth}
\centering
\includegraphics[width=\textwidth]{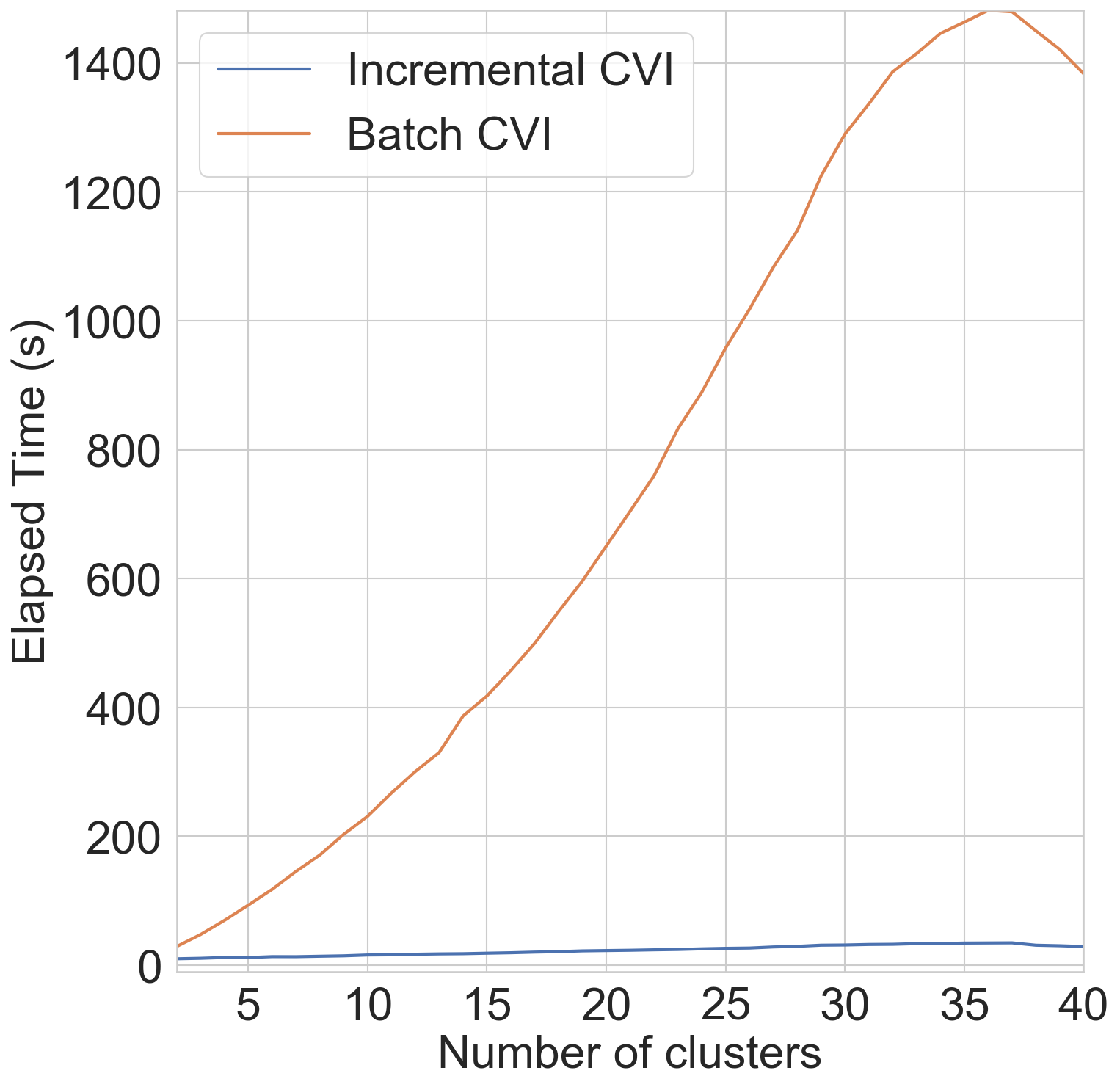}
\caption{Negentropy Increment}
\end{subfigure}
\hfill
\begin{subfigure}[b]{0.3\textwidth}
\centering
\includegraphics[width=\textwidth]{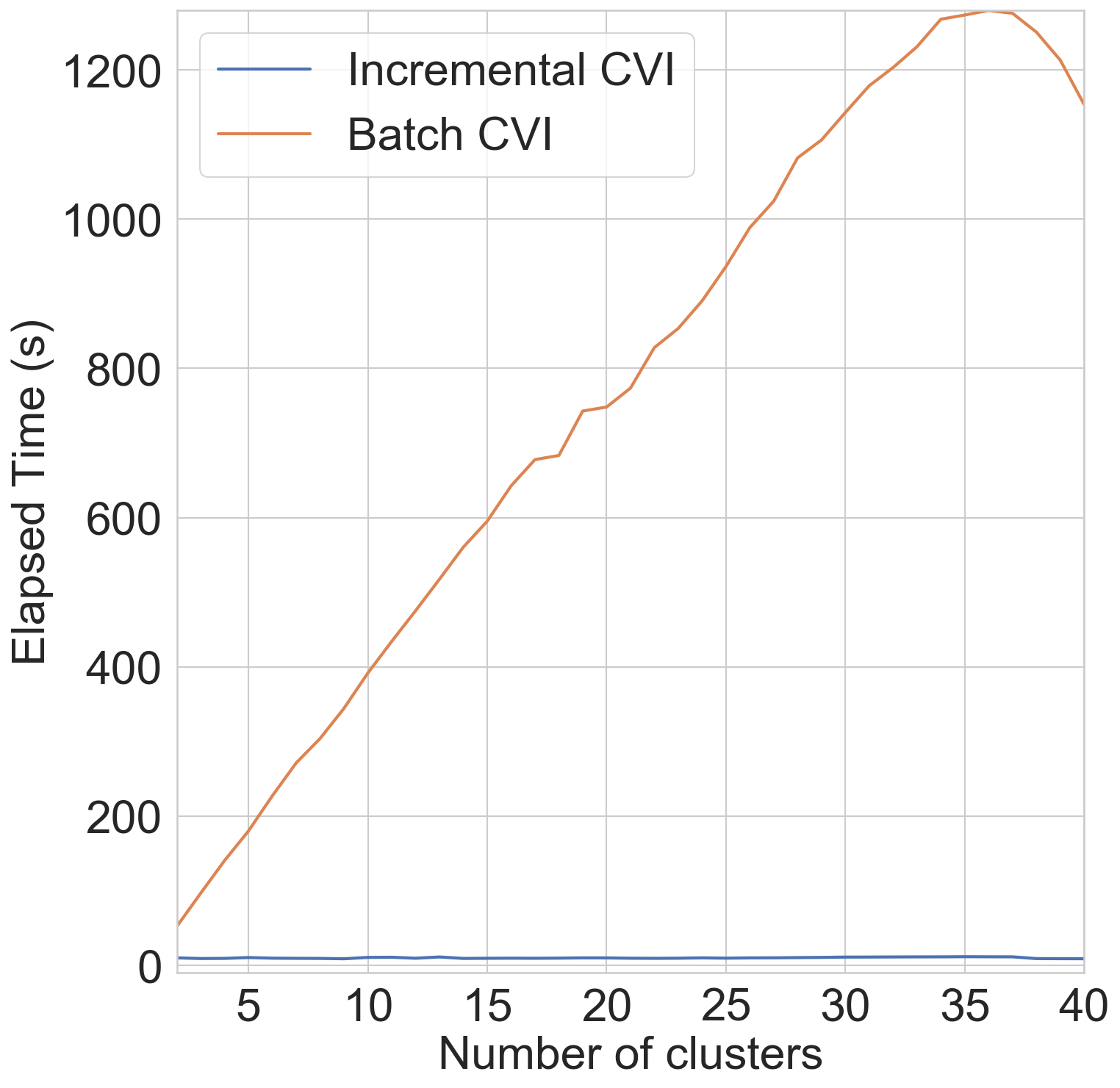}
\caption{Calinski Harabasz}
\end{subfigure}
\hfill
\begin{subfigure}[b]{0.3\textwidth}
\centering
\includegraphics[width=\textwidth]{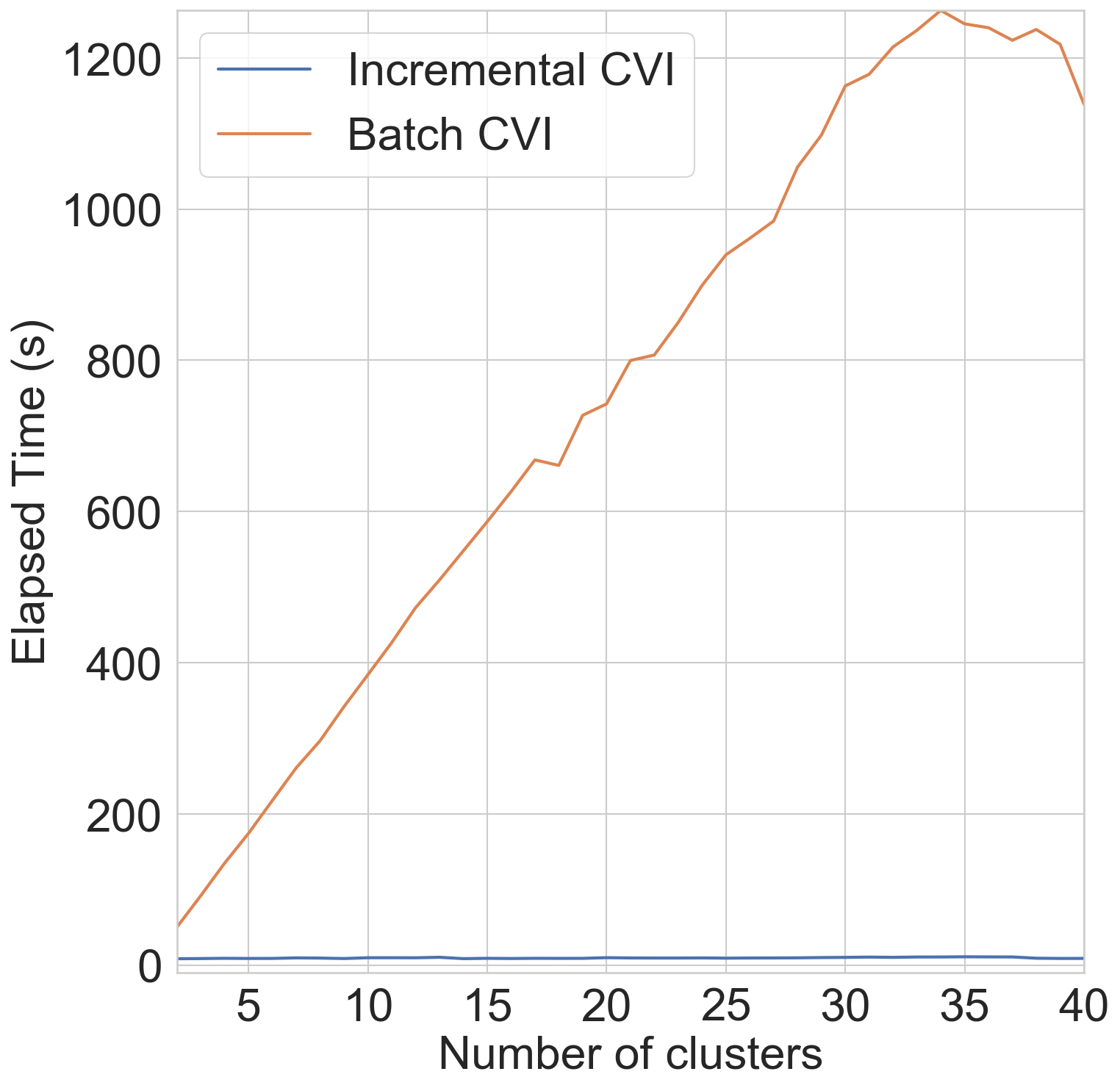}
\caption{WB-index}
\end{subfigure}
\centering
\begin{subfigure}[b]{0.3\textwidth}
\centering
\includegraphics[width=\textwidth]{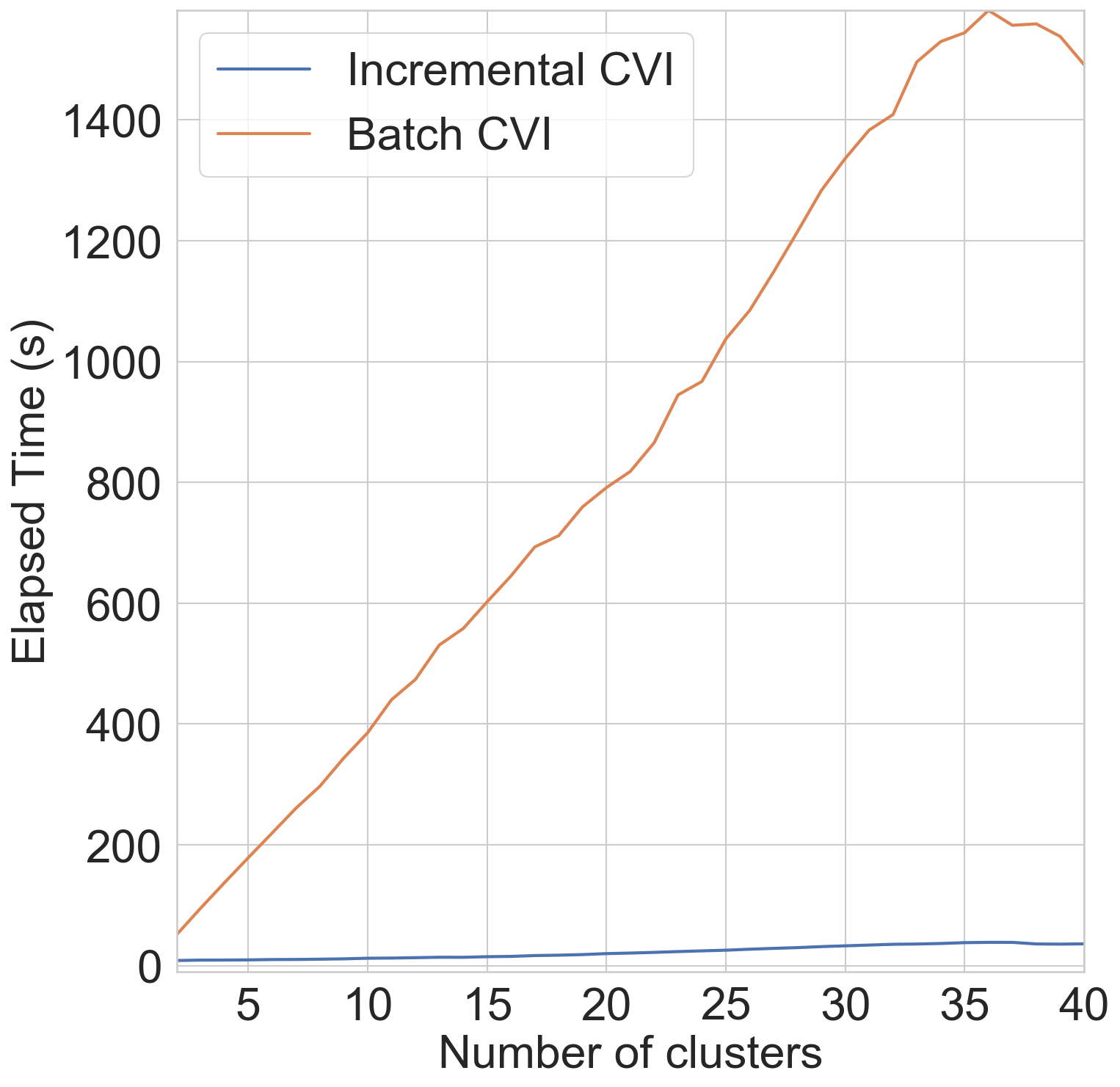}
\caption{Xie-Beni}
\end{subfigure}
\hfill
\begin{subfigure}[b]{0.3\textwidth}
\centering
\includegraphics[width=\textwidth]{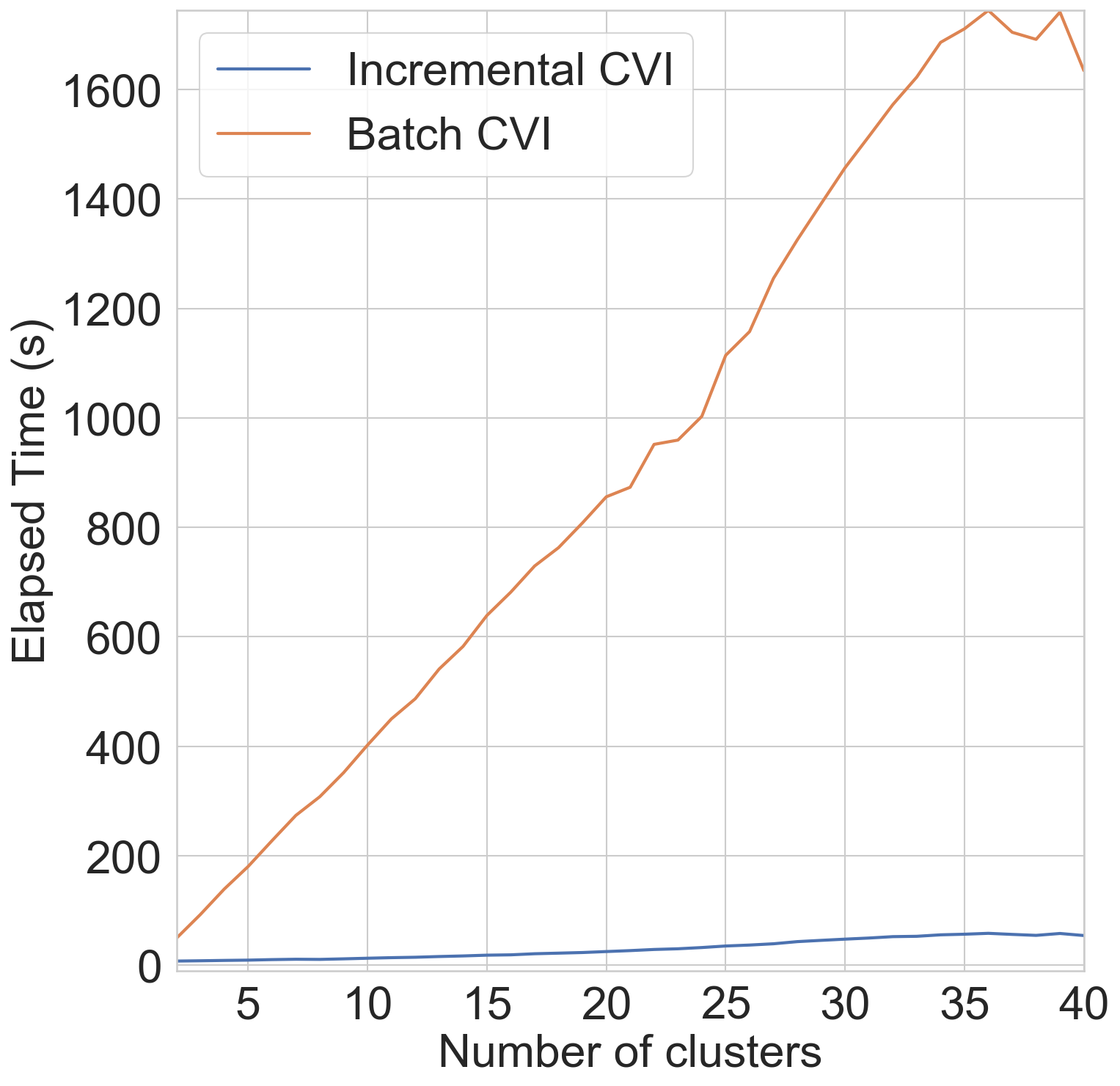}
\caption{Davies-Bouldin}
\end{subfigure}
\hfill
\begin{subfigure}[b]{0.3\textwidth}
\centering
\includegraphics[width=\textwidth]{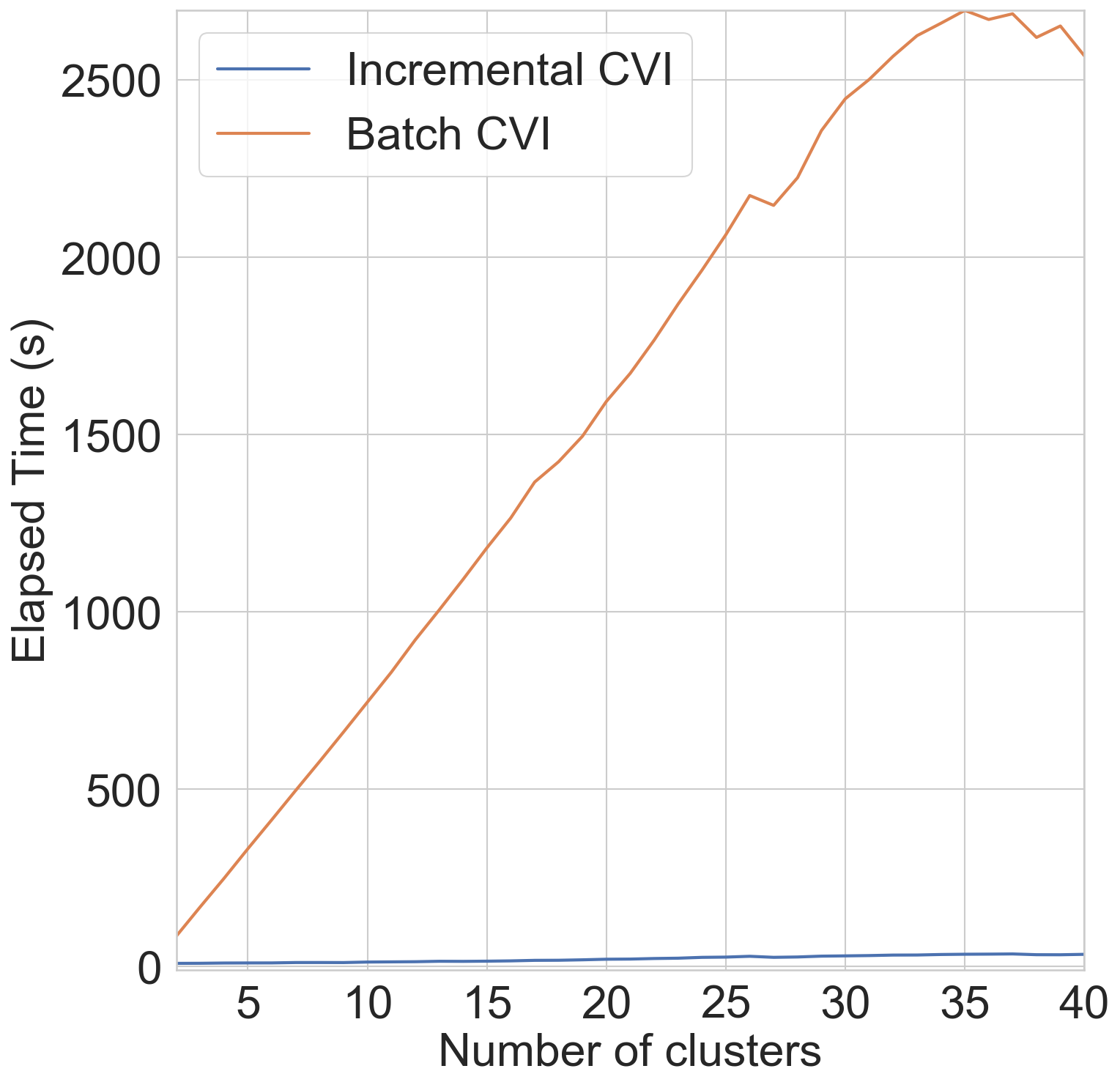}
\caption{Pakhira-Bandyopadhyay-Mauli}
\end{subfigure}
\caption{Elapsed time (in seconds) for running a single epoch of bCVI-ARTMAP (orange curve) and iCVI-ARTMAP (blue curve). The former computes the CVIs in batch mode whereas the latter in incremental mode and caches variables.}
\label{fig:elapsed_time}
\end{figure}

Fig.~\ref{fig:elapsed_time} shows that, for the ellipsoid.100d40c.1 data set, the computational advantage of the presented iCVI-ARTMAP over the corresponding bCVI-ARTMAP clustering variant reaches an average improvement of $53 \times$ across the aforementioned range of $k$ and all CVIs. This includes $28 \times$ for NI, $78 \times$ for CH, $79 \times$ for WB, $35 \times$ for XB, $29 \times$ for DB and $70 \times$ for PBM. As the number of clusters increases, the computational advantage increases even further. In particular, it reaches a maximum improvement of $47 \times$ for NI, $143 \times$ for CH, $140 \times$ for WB, $43 \times$ for XB, $35 \times$ for DB and $84 \times$ for PBM. These results contribute to the viability of this type of offline incremental clustering.

\section{Conclusion} \label{sec:conclusion}

This paper presented a novel ARTMAP model, iCVI-ARTMAP, which transforms ARTMAP into an offline clustering algorithm. This is accomplished by using iCVIs within ARTMAP to generate cluster labels during the training process. Because ART models such as ARTMAP are incremental (or sequential) learners, combining them with iCVIs (along with caching intermediate variables) makes iCVI-ARTMAP a natural approach that can perform all the required CVI computations incrementally, including updates of frequencies, means, compactness and covariance matrices for the operations of adding/removing one sample to/from a cluster, as well as splitting/merging entire clusters. This strategy greatly reduces the computational burden, especially for large values of the number of clusters, thereby permitting the computation of a large number CVI values in much shorter execution times compared to batch mode for each sample presentation across all clusters at all training epochs.

In this work, the following iCVIs were used: negentropy increment (iNI), Calinski-Harabasz (iCH), WB-index (iWB),  Pakhira-Bandyopadhyay-Maulik (iPBM), Xie-Beni (iXB), and Davies-Bouldin (iDB). Experimental results showed that while iNI-ARTMAP was the best-performing variant across the experiments, iCH-, iWB- and iXB-ARTMAP remained strong contenders among the sum-of-squares-based iCVIs. Moreover, iNI-ARTMAP outperformed fuzzy ART, dual vigilance fuzzy ART, k-means, Gaussian mixture models, spectral clustering and hierarchical clustering algorithms in the vast majority of the experiments carried out with synthetic benchmark data sets and yielded competitive performance across the real world data sets. Naturally, the performance of iCVI-ARTMAP depends on the selected iCVI and its suitability to the data at hand. iCVI-ARTMAP comprises a general framework that can be combined with other iCVIs; therefore, extensions to embed other sum-of-squares-based or information-theoretic-based iCVIs in ARTMAP are straightforward.

\appendix
\section*{Appendix} \label{sec:appendix}

To compare the performance of several clustering algorithms, this paper demonstrated their best performance when optimized for the ARI (Tables~\ref{tab:synthetic_results} and~\ref{tab:real_world_results}). It is also of interest to determine the ARI of algorithms that use iCVIs when only the iCVI is used to determine the parameter settings. This can occur in many applications. In the absence of parameter tuning information, iCVIs have significant efficacy in automatically optimizing parameter tuning. In this context, Tables~\ref{tab:iCVI_opt} and~\ref{tab:iCVI_opt_RW} reproduce the experiments on synthetic and real-world data respectively, except that the iCVI is used for parameter tuning in addition to adjusting clustering decisions in the iCVI-ARTMAP variants. Specifically, these Tables report the best results according to the selected iCVI, i.e., from all the vigilance parameter combinations ($\rho_a$, $\rho_{ab}$) with which a given iCVI-ARTMAP variant was trained (Table~\ref{tab:gridsearch}), the one that yielded the output partition with the best corresponding iCVI value is selected, and its ARI value is reported. Effectively, this corresponds to eliminating the requirement of carefully setting these two vigilance parameters; only their search ranges and step sizes are necessary. Note that the comparison to the other baseline clustering algorithms reported on Tables~\ref{tab:iCVI_opt} and~\ref{tab:iCVI_opt_RW}, in the form of average ranks, is based on their results as listed on Tables~\ref{tab:synthetic_results} and~\ref{tab:real_world_results}. Therefore, SC, VAT+FA and VAT+DVFA clustering methods still kept their ARI-optimized performances.  

Table~\ref{tab:iCVI_opt} shows that, in such a scenario, iNI-ARTMAP remained the best-performing method as shown by its average rank (even though some of the competitor algorithms had their parameters optimized for the ARI). The good performance of iNI-ARTMAP corroborates the results reported in~\cite{fernandez2010}, in which a genetic algorithm was used to find data partitions by optimizing batch CVIs. On the other hand, the performance of the sum-of-squares-based iCVI-ARTMAP variants suffered in different degrees. iXB and iDB were particularly affected, whereas iWB and iCH swapped relative positions but still achieved better average ranks than k-means, VAT+DVFA and VAT+FA (as previously mentioned, these two ART-based methods had their parameters adjusted using the ARI). GMM, SC and Ward's HAC yielded the second, third and fourth best average ranks in this experiment, respectively.

\begin{table}[!t]
\centering
\caption{Performance (ARI) of the iCVI-ARTMAP variants on benchmark synthetic data sets when the corresponding iCVIs are also used to tune ($\rho_a$, $\rho_{ab}$). The average ranks are computed considering all nine baseline clustering algorithms as in Table~\ref{tab:synthetic_results}.}
\begin{tabular}{lrrrrrr}
\toprule
\multirow{2}[4]{*}{\textbf{data set}} & 
\multicolumn{6}{c}{iCVI-ARTMAP} \\
\cmidrule{2-7}          
& 
\textbf{iNI} & 
\textbf{iCH} & 
\textbf{iWB} & 
\textbf{iXB} &
\textbf{iDB} &
\textbf{iPBM} \\
\midrule
2d-4c-no0 & 0.9941 & 0.8817 & 0.8817 & 0.8813 & 0.9404 & 0.8851 \\
2d-10c-no0 & 0.9941 & 0.8292 & 0.8292 & 0.7718 & 0.8677 & 0.8721 \\
2d-20c-no0 & 0.9979 & 0.9773 & 0.9773 & 0.9781 & 0.8221 & 0.9773 \\
2d-40c-no0 & 0.9863 & 0.8740 & 0.8898 & 0.8436 & 0.8167 & 0.8738 \\
10d-4c-no0 & 1.0000 & 0.9860 & 0.9860 & 0.9292 & 0.0024 & 0.6881 \\
10d-10c-no0 & 0.9981 & 0.8782 & 0.8802 & 0.8237 & 0.0004 & 0.9196 \\
10d-20c-no0 & 0.9981 & 0.9963 & 0.9963 & 0.9394 & 0.9473 & 0.9916 \\
10d-40c-no0 & 0.9940 & 0.9890 & 0.9890 & 0.9408 & 0.9669 & 0.9901 \\
ellipsoid.50d4c.1 & 1.0000 & 0.4731 & 0.4539 & 0.2899 & 0.0464 & 0.3453 \\
ellipsoid.50d10c.1 & 0.9995 & 0.3987 & 0.3987 & 0.3319 & 0.1874 & 0.3395 \\
ellipsoid.50d20c.1 & 1.0000 & 0.3404 & 0.3404 & 0.2851 & 0.0878 & 0.3002 \\
ellipsoid.50d40c.1 & 0.9645 & 0.3167 & 0.3167 & 0.2417 & 0.0733 & 0.2480 \\
ellipsoid.100d4c.1 & 1.0000 & 0.3957 & 0.3957 & 0.2490 & -0.0017 & 0.2245 \\
ellipsoid.100d10c.1 & 1.0000 & 0.5001 & 0.5001 & 0.2927 & 0.0398 & 0.4162 \\
ellipsoid.100d20c.1 & 0.9486 & 0.3261 & 0.3261 & 0.1832 & 0.0387 & 0.2993 \\
ellipsoid.100d40c.1 & 0.9750 & 0.2822 & 0.2822 & 0.1655 & 0.1075 & 0.2680 \\
\midrule
\textbf{Average rank} & 1.72  & 7.44  & 7.34  & 10.94 & 12.28 & 8.63 \\
\bottomrule
\end{tabular}
\label{tab:iCVI_opt}
\end{table}

\begin{table}[!t]
\centering
\caption{Performance (ARI) of the iCVI-ARTMAP variants on benchmark real world data sets when the corresponding iCVIs are also used to tune ($\rho_a$, $\rho_{ab}$). The average ranks are computed considering all nine baseline clustering algorithms as in Table~\ref{tab:real_world_results}. The notation ``($dim-D$)'' shown in parentheses under the ``pre-processing'' column indicates that the data set was transformed to a $dim$-dimensional space using the corresponding technique.}
\begin{tabular}{lrrr}
\toprule
\multirow{2}[4]{*}{\textbf{data set}} & 
\multirow{2}[4]{*}{\textbf{Pre-processing}} &
\multicolumn{2}{c}{iCVI-ARTMAP} \\
\cmidrule{3-4} 
&       & 
\textbf{iNI} & 
\textbf{iCH} \\
\midrule
Olivetti Faces & 
PCA (20-D) & 0.6374 & 0.5825 \\
USPS  & 
DynAE (10-D) & 0.8842 & 0.9599 \\
MNIST-test & 
DynAE (10-D) & 0.8634 & 0.9627 \\
MNIST & 
DynAE (10-D) & 0.9698 & 0.9688 \\
Fashion MNIST & 
DynAE (10-D) & 0.4802 & 0.4495 \\
\midrule
\textbf{Average rank} & & 4.40  & 4.00 \\
\bottomrule
\end{tabular}
\label{tab:iCVI_opt_RW}
\end{table}

Table~\ref{tab:iCVI_opt_RW} shows that, for the real world data sets, iNI-ARTMAP no longer performed better than the other methods for the USPS and MNIST-test data sets, while its performance for the other three data sets remained the same. On the other hand, iCH-ARTMAP yielded a slightly poorer performance for the Olivetti faces and USPS and maintained the same level of performance for the other three data sets. Specifically, this variant still outperformed kmeans when clustering the Olivetti faces and performed similarly for the remaining data sets; the comparative analysis with VAT+FA and VAT+DVFA remain the same. Therefore, the iCVI-ARTMAP model presented here is still viable for applications in which the practitioner can only rely on iCVI techniques.  

\medskip

\bibliography{references}

\end{document}